# Learning Representations of Entities and Relations

*Ivana Balažević*

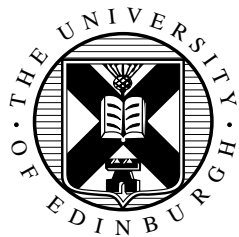

Doctor of Philosophy
School of Informatics
University of Edinburgh

November 2021

# Abstract

Learning to represent factual knowledge about the world in a succinct and accessible manner is a fundamental machine learning problem. Encoding facts as representations of entities and binary relationships between them, as learned by *knowledge graph representation models*, is useful for various tasks, including predicting new facts (i.e. link prediction), question answering, fact checking and information retrieval. The focus of this thesis is on (i) improving knowledge graph representation with the aim of tackling the link prediction task; and (ii) devising a theory on how semantics can be captured in the geometry of relation representations.

Most knowledge graphs are very incomplete and manually adding new information is costly, which drives the development of methods which can automatically infer missing facts. This thesis introduces three knowledge graph representation methods, each applied to the link prediction task. The first contribution is *HypER*, a convolutional model which simplifies and improves upon the link prediction performance of the existing convolutional state-of-the-art model ConvE and can be mathematically explained in terms of constrained tensor factorisation. Drawing inspiration from the tensor factorisation view of HypER, the second contribution is *TuckER*, a relatively straightforward linear knowledge graph representation model, which, at the time of its introduction, obtained state-of-the-art link prediction performance across standard datasets. With a specific focus on representing hierarchical knowledge graph relations, the third contribution is *MuRP*, first multi-relational graph representation model embedded in hyperbolic space. MuRP outperforms all existing models and its Euclidean counterpart MuRE in link prediction on hierarchical knowledge graph relations whilst requiring far fewer dimensions. Since their publication, all above mentioned models have influenced a range of subsequent developments in the knowledge graph representation field.

Despite the development of a large number of knowledge graph representation models with gradually increasing predictive performance, relatively little is known of the latent structure they learn. We generalise recent theoretical understanding of how semantic relations of similarity, paraphrase and analogy are encoded in the geometric interactions of word embeddings to how more general relations, as found in knowledge graphs, can be encoded in their representations. This increased theoretical understanding can be used to aid future knowledge graph representation model design, as well as to improve models which incorporate logical rules between relations into their representations or those that jointly learn from multiple data sources (e.g. knowledge graphs and text).

# Lay Summary

Vast amounts of factual knowledge about the real world can be stored in a succinct manner in the shape of a graph, forming typed, directed graph data structures called knowledge graphs. In knowledge graphs, facts are stored as relations between entities, where entities are nodes in the graph and relations are edges between them. For example, the fact that London is the capital of the UK can be summarised as *(London, capital_of, UK)* and stored in a knowledge graph as a directed edge of type *capital_of* pointing from the node denoting the subject entity *London* towards the node denoting the object entity *UK*. Knowledge graphs are typically created manually by humans or in an automated manner, using handcrafted rules and natural language processing techniques. Some of the most common use cases involving knowledge graphs include question answering, information retrieval, fact checking and inferring new facts.

Knowledge graphs are often very incomplete and manually adding new information is costly, which creates the need for methods which can automatically infer new facts. For example, 71% of people in a well-known Freebase knowledge graph have no known place of birth, 94% have no known parents, and 99% have no known ethnicity. To address this, Chapters 3, 4 and 5 of this thesis introduce a range of methods which can automatically infer new facts based on the ones already present in a knowledge graph.

Despite their impressive accuracy when inferring new knowledge graph facts, most existing methods have been developed heuristically and there is little theoretical rationale for why and how they work. In Chapter 6 of this thesis, we develop a theory on how such methods can capture the meaning of entities and relations between them in their geometric components, which may be used to guide future method design.

# Acknowledgements

I would like to thank my primary supervisor Timothy Hospedales for his continued support, guidance and encouragement throughout the course of my PhD. Special thanks goes to my secondary supervisor Ivan Titov for his keen interest in my work and numerous productive discussions.

During my PhD, I was fortunate enough to do an internship at Facebook AI Research under the supervision of Sebastian Riedel. Working with Sebastian has had an immense impact on my thinking and my professional development, for which I am extremely grateful. I would also like to thank other members of my Facebook AI Research team, especially Fabio Petroni and Patrick Lewis for many helpful discussions.

I am very grateful to my PhD and internship collaborators Carl Allen, Sebastian Riedel, Sameer Singh, Fabio Petroni, Robert L. Logan IV, and Eric Wallace for their contribution and everything I learned from them.

I would also like to thank all my friends from the School of Informatics, the CDT in Data Science, and Edinburgh in general for making my PhD experience enjoyable and memorable through many interesting research discussions, numerous pub nights throughout the Edinburgh winters, sanity preserving climbing sessions and hikes, and long coffee breaks: Carl Allen, Nelly Papalampidi, Maria Gorinova, Magdalena Navarro, Nestor Sanchez, Artur Bekasov, Ed Fincham, Anna Page, Ribanna Dittrich, Jonathan Mallinson, Benedek Rózemberczki, Nick Hoernle, and Theresa Schöpp. Special thanks to all my friends in Croatia and around the world who remained close despite the distance: Ana Roksandić, Brian McWilliams, Ivana Kajić, Izabela Bratović, Adriana Biuk, Tomislav Zoričić, Ana Marasović, Daniil Sorokin, and Jelena Cvitanović.

A big thank you goes to my parents for their unwavering love and support, for comforting me when things were rough and for celebrating all the successes in their own way, loudly and proudly. Finally, the most special thank you goes to Carl, for making my life infinitely better and always being there, in research and in life. It is not an exaggeration to say that none of this would have been possible without you.

This work was supported in part by the EPSRC Centre for Doctoral Training in Data Science, funded by the UK Engineering and Physical Sciences Research Council (grant EP/L016427/1) and the University of Edinburgh.

# Declaration

I declare that this thesis was composed by myself, that the work contained herein is my own except where explicitly stated otherwise in the text, and that this work has not been submitted for any other degree or professional qualification except as specified.

(*Ivana Balažević*)

# Table of Contents

# Chapter 1

# Introduction

The ability to reason about the world in a structured manner in terms of entities and relationships between them is an important aspect of human and machine cognition (Spelke and Kinzler, 2007), deemed crucial for achieving *combinatorial generalisation* (Battaglia et al., 2018) — constructing new inferences, predictions, and behaviors from known building blocks. Recently, a growing number of machine learning methods have been focusing on *extracting structure* from unstructured data (e.g. entity and relation extraction from text (Riedel et al., 2013; Luan et al., 2018; Wadden et al., 2019), object extraction from images and videos (Greff et al., 2019; Engelcke et al., 2020; Kipf et al., 2020)). Complementary to such efforts, many works assume data is already available in structured form and instead focus on *representing and reasoning over such structured data* (e.g. knowledge graph representation (Nickel et al., 2011; Bordes et al., 2013; Yang et al., 2015; Trouillon et al., 2016), modelling object interactions in structured world models (Battaglia et al., 2016; Kipf et al., 2018), state transitions of an agent in reinforcement learning (Kipf et al., 2020), or citation networks (Kipf and Welling, 2017; Veličković et al., 2018)), with a wide range of potential applications to downstream tasks (e.g. inferring new facts, drug discovery, recommender systems).

In line with recent work on representation learning from structured data, the focus of this thesis is on development as well as theoretical analysis of methods which *encode factual knowledge* about the world in a succinct and accessible manner by capturing statistical patterns in structured data. Specifically, we look at embedding-based approaches to *knowledge graph representation* — learning representations of entities and relations between them from *knowledge graphs*, structured collections of known facts. Knowledge graphs store facts in terms of binary relationships between entities in the form of *(subject_entity, relation, object_entity)* triples, e.g. *(London, capital_of, UK)*. Embeddings learned by knowledge graph representation models can be used for various downstream tasks such as question answering, fact checking, information retrieval and inferring new facts.

In contrast to large language models, where knowledge is stored in an unstructured manner in their parameters (Petroni et al., 2019) and thus it is unclear how new facts can be added or existing ones modified, knowledge graphs are important information resources which store facts in an accessible and interpretable manner. However, due to the ever increasing amount of data and the difficulty in manually capturing all available factual information, knowledge graphs are often very incomplete, e.g. 71% of people in Freebase (Bollacker et al., 2008) have no known place of birth, 94% have no known parents, and 99% have no known ethnicity (West et al., 2014). Since manually adding new facts is time consuming and costly, numerous methods which automatically infer missing facts have been developed in recent years — a task known as *link prediction* or *knowledge graph completion*. Building on a large body of prior work in knowledge graph representation, Chapters 3, 4 and 5 of the thesis introduce novel knowledge graph representation methods, each evaluated on the link prediction task.

Despite the consistent increase in predictive performance of knowledge graph representation models, there is little theoretical rationale on how and why they work, which is addressed in Chapter 6. Following from recent theoretical understanding of how various semantic relations (similarity, paraphrase and analogy) are encoded in the geometric interactions of word embeddings (Allen and Hospedales, 2019; Allen et al., 2019), we expand this understanding to more general semantic relations which can be found in knowledge graphs. By providing a theoretical link between semantics and geometry of relation representations, the presented analysis may assist with future knowledge graph representation model design, as well as with developing models which incorporate logical reasoning or jointly learn from multiple data sources.

## 1.1   List of Contributions

The main contributions of this thesis are three different knowledge graph representation methods, one of which is particularly focused on modelling hierarchical knowledge graph relations, and a theoretical model for how semantic relations between words can be encoded in the geometric structure of their representations.

- **HypER** is a convolutional knowledge graph representation model which simplifies and improves upon the link prediction performance of the at the time state-of-the-art convolutional model ConvE (Dettmers et al., 2018). The black-box convolutional model can be alternatively framed as a constrained tensor factori-sation, positioning HypER close to a well established family of factorisation-based knowledge graph representation models.

- **TuckER**, which was inspired by previous tensor factorisation approaches (e.g. Nickel et al., 2011; Trouillon et al., 2016) as well as the tensor factorisation view of HypER, uses Tucker decomposition to obtain state-of-the-art results in link prediction across standard datasets. Despite being a relatively straightforward linear model, TuckER outperforms not only existing tensor factorisation models (which can all be seen as a special case of TuckER), such as RESCAL (Nickel et al., 2011), DistMult (Yang et al., 2015) and ComplEx (Trouillon et al., 2016), but also more complex deep neural network and reinforcement learning architectures (e.g. Schlichtkrull et al., 2018; Das et al., 2018). This demonstrates that simple linear models should serve as a baseline before moving onto more elaborate models. The improved performance of TuckER is achieved by leveraging shared latent structure between entities and relations through multi-task learning.

- **MuRP** was inspired by a large body of work on embedding hierarchical data in hyperbolic instead of Euclidean space (e.g. Nickel and Kiela, 2017, 2018; Ganea et al., 2018b). Hyperbolic space, which can be thought of as a continuous analogue to discrete trees, is particularly suitable for modelling hierarchical data structures. MuRP is the first multi-relational graph representation model to embed entities and relations in hyperbolic space which outperforms existing Euclidean models on the link prediction task, particularly at lower embedding dimensionality.

- **Interpreting Knowledge Graph Relation Representation from Word Embeddings** draws a link between context-free word embeddings (Mikolov et al., 2013a; Pennington et al., 2014) and knowledge graph representation, suggesting commonality in their latent structures despite significant differences in their training data and methodology. This work builds upon recent theoretical understanding (Allen and Hospedales, 2019; Allen et al., 2019) of how particular semantic relations, e.g. similarity and analogy, are captured in the interactions of word embeddings and devises a theoretical model for how more general knowledge graph relations can be captured in their corresponding relation representations.

## 1.2 Thesis Structure

This thesis consists of four chapters of main content, each of which is centered around a different published paper. The four chapters are listed below.

Chapter 3 introduces HypER and it is based on the following paper:

I. Balažević, C. Allen, and T. Hospedales.  Hypernetwork Knowledge Graph Embeddings. *International Conference on Artificial Neural Networks*, 2019

Chapter 4 introduces TuckER and it is based on the following paper:

I. Balažević, C. Allen, and T. Hospedales.  TuckER: Tensor Factorization for Knowledge Graph Completion. *Empirical Methods in Natural Language Processing*, 2019

Chapter 5 introduces MuRP and it is based on the following paper:

I. Balažević, C. Allen, and T. Hospedales.  Multi-relational Poincaré Graph Embeddings. *Advances in Neural Information Processing Systems*, 2019

Chapter 6 introduces the theory of how semantics can be captured in the geometry of relation representations and it is based on the following paper (* indicates equal contribution):

C. Allen*, I. Balažević*, and T. Hospedales.  Interpreting Knowledge Graph Relation Representation from Word Embeddings. *International Conference on Learning Representations*, 2021

# Chapter 2

# Background and Related Work

This chapter contains the required background and an overview of prior work on knowledge graph representation and word embeddings. Section 2.1 introduces knowledge graphs and the link prediction task. Section 2.2 presents notable knowledge graph representation models, followed by the description of the training and evaluation methods used across all models presented in this thesis. Section 2.3 introduces static/context-free word embedding models, as well as a recent theory on how semantics is captured in the geometry of word embedding interactions. Section 2.4 contains the background on hyperbolic geometry of the Poincaré ball, required for understanding Chapter 5.

## 2.1   Knowledge Graphs

*Knowledge graphs* are large repositories of known facts, typically stored as a collection of binary[1] relationships between entities in *(subject_entity, relation, object_entity)* triple form. A collection of triples $(e_s, r, e_o) \in \mathcal{E} \times \mathcal{R} \times \mathcal{E}$, where $\mathcal{E}$ is the set of entities and $\mathcal{R}$ is the set of relations, can alternatively be represented as a *directed, multi-relational graph*, with entities as the *nodes* of the graph and relations as typed, directed *edges* between them. The direction of an edge indicates whether a particular entity occurs as a subject or as an object under each relation, with an edge pointing from the subject to the object. The existence of a particular edge in a knowledge graph denotes a known fact, e.g. *(London, capital_of, UK)*, whereas the absence of an edge, due to knowledge graph incompleteness, denotes the unknown — either a missing or an incorrect fact. A knowledge graph can also be represented as a *binary adjacency 3D tensor* $\mathcal{G} \in \{0, 1\}^{n_e \times n_r \times n_e}$, where $n_e$ denotes the number of entities and $n_r$ the number of relations, with 1 indicating that a relation exists between a pair of entities. From a probabilistic perspective, each entry of $\mathcal{G}$ can be seen as a realisation of a Bernoulli-distributed random variable.

---

[1] Higher-arity relations (e.g. to capture time or location) are not considered in this work.

Knowledge graphs are typically constructed manually or in an automated manner, using a combination of handcrafted rules and natural language processing techniques. Since it is practically impossible for a knowledge graph construction process to cover all existing facts, knowledge graphs are known to be incomplete (Nickel et al., 2015)). This creates the need for methods which can automatically infer other facts based on the ones already present in a knowledge graph — a task known as *link prediction* or *knowledge graph completion* — on which the methods developed in this thesis are evaluated and compared. Other tasks of interest relating to knowledge graphs, which are outside the scope of this thesis, include information retrieval, question answering and fact checking.

## 2.2   Knowledge Graph Representation

The ability to infer new facts in a computationally efficient manner and without the need for handcrafted logical rules or human intervention requires *representation*. The following two sections introduce the most common approaches to knowledge graph representation. Section 2.2.1 presents *embedding models*, which learn entity and relation representations, defined exclusively through a score function, by optimising their parameters to improve the prediction accuracy. Section 2.2.2 briefly introduces *neural network models*, where representations are learned by using convolutional neural networks (CNNs) or graph neural networks (GNNs) in conjunction with a score function. Sections 2.2.3 and 2.2.4 describe the training and evaluation procedures common to all knowledge graph representation methods presented in this thesis.

### 2.2.1   Embedding Models

Despite their relative simplicity in comparison to neural network models, embedding-based approaches hold the current state-of-the-art in link prediction. Most embedding models in knowledge graph representation literature (e.g. Nickel et al., 2011; Bordes et al., 2013; Yang et al., 2015; Trouillon et al., 2016; Dettmers et al., 2018) represent entities $e_s, e_o \in \mathcal{E}$ as vectors/*embeddings* in $d_e$-dimensional space, where $d_e << n_e$, with $\mathbf{e}_s, \mathbf{e}_o \in \mathbb{R}^{d_e}$ denoting subject and object entity embeddings respectively. Relations $r \in \mathcal{R}$ on the other hand are represented as *transformations* applied to $\mathbf{e}_s$ and $\mathbf{e}_o$. A *proximity measure* is used to measure the "closeness" between relation-transformed subject and object entity embeddings. Entity representations, relation representations and the proximity measure are all defined through a *score function* $\phi : \mathcal{E} \times \mathcal{R} \times \mathcal{E} \to \mathbb{R}$, which assigns a score $\phi(e_s, r, e_o)$ to each triple in order to predict whether the triple is true or false. Score functions of existing embedding models differ primarily by 1) relation representation type (e.g. additive, multiplicative); and 2) choice of proximity measure (e.g. Euclidean distance, dot product).

**Additive Models**  The relation-specific transformation of additive models is a *translation vector* $\mathbf{r} \in \mathbb{R}^{d_e}$, applied to a subject entity embedding (or alternatively, an object entity embedding). The most common proximity measure of additive models is (squared) Euclidean distance. The score function of TransE, (Bordes et al., 2013), one of the early knowledge graph representation models and the most prominent additive model, is defined as $\phi(e_s, r, e_o) = -\|\mathbf{e}_s + \mathbf{r} - \mathbf{e}_o\|_2^2$.

**Multiplicative Models**  Across multiplicative embedding models, relations are represented via *matrix multiplication* by $\mathbf{R} \in \mathbb{R}^{d_e \times d_e}$, with dot product as the most common proximity measure between the relation-transformed entity embeddings. The generic score function of multiplicative models is defined as $\phi(e_s, r, e_o) = \mathbf{e}_s^\top \mathbf{R} \mathbf{e}_o$. Various proposed models differ in terms of the constraints they impose on the relation-specific matrix $\mathbf{R}$. RESCAL (Nickel et al., 2011) is an early multiplicative model where $\mathbf{R}$ is full rank, which makes RESCAL prone to overfitting due to its large number of parameters, which increases quadratically with the number of relations. DistMult (Yang et al., 2015) is a special case of RESCAL with $\mathbf{R}$ constrained to be a diagonal matrix, which limits the linear transformation to a stretch, so DistMult cannot model asymmetric relations. ComplEx (Trouillon et al., 2016) extends DistMult to the complex domain. Subject and object entity embeddings for the same entity are complex conjugates, which introduces asymmetry into the tensor decomposition and thus enables ComplEx to model asymmetric relations. Kazemi and Poole (2018) and Balažević et al. (2019b) show that the score function of ComplEx is equivalent to a constrained Tucker decomposition (Tucker, 1964) of the binary adjacency tensor $\mathcal{G} \in \{0, 1\}^{n_e \times n_r \times n_e}$ in the real domain. In TuckER (Balažević et al., 2019b), each relation-specific matrix $\mathbf{R} = \mathcal{W} \times_3 \mathbf{w}_r$ is a linear combination of $d_r$ "prototype" relation matrices, i.e. slices from the core tensor $\mathcal{W} \in \mathbb{R}^{d_e \times d_r \times d_e}$ of the Tucker decomposition of $\mathcal{G}$. All multiplicative models introduced above are a special case of TuckER, as shown in Section 4.2.

**Additive and Multiplicative Models**  Some models represent relations by combining both multiplicative ($\mathbf{R}$) and additive ($\mathbf{r}$) components. The generic score function is defined as $\phi(e_s, r, e_o) = -\|\mathbf{R}_s \mathbf{e}_s + \mathbf{r} - \mathbf{R}_o \mathbf{e}_o\|_2^2 + b_s + b_o$. An example of an early additive and multiplicative model is STransE (Nguyen et al., 2016), where the biases $b_s, b_o \in \mathbb{R}$ are equal to 0. The score function of MuRE (Balažević et al., 2019a) corresponds to the generic score function for both additive and multiplicative models with $\mathbf{R}_o = \mathbf{I}$.

### 2.2.2   Neural Network Models

In contrast to embedding models which represent factual information simply through relation-specific additive and multiplicative transformations applied to entity embeddings, convolutional neural network (CNN) and graph neural network (GNN) models aim at increasing expressiveness by introducing several nonlinear layers.

**Convolutional Neural Networks**   Several works use CNNs to model entity and relation interactions. ConvE (Dettmers et al., 2018) is the first convolutional knowledge graph representation model, where a global 2D convolution operation is performed on the subject entity embedding and relation embedding after reshaping them to a matrix and concatenating them. HypER (Balažević et al., 2019) simplifies ConvE by removing the reshaping and concatenation of embeddings and replacing the global 2D convolution with a relation-specific 1D convolution. Further, HypER can be recast in terms of constrained tensor factorisation and is thus adjacent to the family of multiplicative models described in Section 2.2.1. The link prediction performance improvements of HypER in comparison to ConvE suggest that the role of convolution is as a convenient computational means of introducing sparsity and parameter tying, rather than in capturing 2D structure within entity embeddings.

**Graph Neural Networks**   GNNs are used for learning the topological structure of node neighborhoods under an encoder–decoder framework, where the encoder is typically a graph convolutional network (GCN) architecture adapted for multi-relational graphs and the decoder is a score function of one of the embedding models described in Section 2.2.1. R-GCN (Schlichtkrull et al., 2018) is the first such model which introduces relation-specific weight matrices to model a GCN layer. This leads to overparameterisation and overfitting issues, which is alleviated by using basis and block-diagonal decomposition of the weight matrices. R-GCN uses the DistMult score function to train the model for the link prediction task. Recently, Vashishth et al. (2020) introduce CompGCN which embeds relations as vectors rather than matrices, which makes CompGCN more parameter efficient than R-GCN. TransE, DistMult and ConvE score functions are used for link prediction. However, despite their increased expressiveness, the link prediction performance of recent GNN-based knowledge graph representation models matches the performance of the simpler embedding approaches.

### 2.2.3   Training Knowledge Graph Representation Models

Knowledge graph representation is an example of a *positive and unlabelled learning (PU learning)* problem. Knowledge graphs store a subset of all true triples, i.e. only a subset of all positive entries — the realisations of the Bernoulli-distributed random variables that are equal to 1 — are observed in $\mathcal{G}$. More precisely, only a subset $(e_s, r, e_o) \in \mathcal{S} \subseteq \mathcal{F}$ of all positive triples $\mathcal{F}$ are observed and all other triples $(e_s, r, e_o) \in (\mathcal{F} \setminus \mathcal{S}) \bigcup \mathcal{F}^-$ are unlabelled, where $\mathcal{F} \setminus \mathcal{S}$ indicates the set of all unobserved true triples and $\mathcal{F}^-$ false triples. To enable learning, existing knowledge graph representation models generate negative triples either by negative sampling or 1-N scoring (Dettmers et al., 2018).

**Negative Sampling**  Negative sampling relies upon the very sparse nature of knowledge graphs where unobserved true facts constitute a very small subset of all triples (i.e. $|\mathcal{F} \setminus \mathcal{S}| << |\mathcal{F}^-|$), so sampling from $(\mathcal{F} \setminus \mathcal{S}) \bigcup \mathcal{F}^-$ will likely result in selecting truly negative triples $(e_s, r, e_o) \in \mathcal{F}^-$. For each observed positive triple $(e_s, r, e_o) \in \mathcal{S}$, $k$ negative samples (typically with $1 \leq k << n_e$) are generated by corrupting either the object $(e_s, r, e_o') \in (\mathcal{F} \setminus \mathcal{S}) \bigcup \mathcal{F}^-$ or the subject $(e_s', r, e_o) \in (\mathcal{F} \setminus \mathcal{S}) \bigcup \mathcal{F}^-$ entity by sampling a random entity from the set of all entities $e_s', e_o' \in \mathcal{E}$.

**1-N Scoring**  To speed up training, Dettmers et al. (2018) introduce 1-N scoring, where entity-relation pairs $(e_s, r)$ and $(r, e_o)$ are simultaneously scored with all entities $e_o \in \mathcal{E}$ and $e_s \in \mathcal{E}$ respectively. All triples $(e_s, r, e_o) \in \mathcal{S}$ which complete the pairs $(e_s, r)$ and $(r, e_o)$ are considered as positive samples and all other triples $(e_s, r, e_o) \in (\mathcal{F} \setminus \mathcal{S}) \bigcup \mathcal{F}^-$ as negative samples. The number of negative samples generated in this manner is typically larger than with classic negative sampling which has been shown to benefit performance (Kadlec et al., 2017), despite increasing the likelihood of incorrectly choosing triples from $(e_s, r, e_o) \in (\mathcal{F} \setminus \mathcal{S})$ as negative samples.

Having generated negative triples, the score function is used to compute the score for both positive and negative triples and logistic sigmoid is applied to measure the probability $p \in [0, 1]$ of a triple being true, e.g. $p = \sigma(\phi(e_s, r, e_o))$. Gradient-based methods are then used to train the model using the Bernoulli negative log-likelihood loss function, where representation parameters are optimised so that positive samples are assigned a high score (i.e. $p \approx 1$) and negative samples a low score (i.e. $p \approx 0$).

### 2.2.4 Evaluating Knowledge Graph Representation Models

Knowledge graph representation models are typically evaluated on the link prediction task using *ranking metrics*. For each triple from the test set, $n_e$ evaluation triples are generated by combining the test entity-relation pair with all possible entities $\mathcal{E}$. The scores for each evaluation triple are obtained by applying the score function and then ranked after removing all true triples from the evaluation triples apart from the current test triple, i.e. the commonly used *filtered setting* (Bordes et al., 2013). The evaluation metrics standard across the link prediction literature, mean rank (MR), mean reciprocal rank (MRR) and hits@$k$, $k \in \{1, 3, 10\}$, are used to measure how the score of the true triple compares to the scores of other evaluation triples. Mean rank is the average rank assigned to the true triple, over all $n_e$ evaluation triples. Mean reciprocal rank is the average of the inverse of a mean rank assigned to the true triple over all evaluation triples. Hits@$k$ measures the percentage of times the true triple appears in the top $k$ ranked evaluation triples.

Despite their widespread use, ranking metrics have several drawbacks compared to simple binary classification metrics. Most importantly, the evaluation is flawed if an entity is related to $k' > k$ others, which would cause a correct prediction falling within the top $k'$ but not the top $k$ to appear incorrect under the metric. Further, ranking evaluation is computationally costly since each true triple is compared to all other evaluation triples, whereas classification metrics evaluate standalone predictions. Several recent works note the importance of standalone predictions and enhance existing datasets with negative test samples for classification-based evaluation (Speranskaya et al., 2020; Pezeshkpour et al., 2020).

## 2.3   Word Embedding

A classic approach to word meaning is *distributional semantics*, which is based on the idea that words have similar meanings if they have similar patterns of usage and co-occurrence with other words (Harris, 1954; Firth, 1957). Static word embedding algorithms, such as Word2Vec (Mikolov et al., 2013a) and GloVe (Pennington et al., 2014), are amongst the most well-known models of distributional semantics and have been shown to generate low-dimensional word embeddings which perform well on downstream tasks (Baroni et al., 2014). Despite the very strong performance achieved by recent language models (e.g. Devlin et al., 2018; Radford et al., 2018, 2019) which use contextualised word embeddings, in this thesis we focus on the simpler, context-free embeddings since knowledge graph entities have no obvious context and, importantly, they offer insight into embedding interpretability. Section 2.3.1 describes Word2Vec and GloVe, while Section 2.3.2 covers recent work (Allen and Hospedales, 2019; Allen et al., 2019) on how semantics is captured in the geometry of word embedding interactions.

### 2.3.1   Word Embedding Models

**Word2Vec** (Mikolov et al., 2013a) takes as input $D$ word pairs $(w_i, c_j)$ extracted from a large text corpus, where *target words* $w_i \in \mathcal{V}$ range over the corpus, *context words* $c_j \in \mathcal{V}$ are selected from a window of size $s$ symmetric about $w_i$, and $\mathcal{V}$ is the word vocabulary with $n = |\mathcal{V}|$ indicating the vocabulary size, i.e. the number of distinct words in the corpus. For each observed word pair, $k$ negative sample pairs are generated at random by sampling each word in the pair from the unigram distribution. Word2Vec's simple two-layer neural network architecture consists of the product of two weight matrices $\mathbf{W}, \mathbf{C} \in \mathbb{R}^{d \times n}$ subject to the logistic sigmoid function, where $d$ is the embedding dimension (where typically $d << n$). Columns of $\mathbf{W}$ and $\mathbf{C}$ are the *word embeddings*: $\mathbf{w}_i \in \mathbb{R}^d$, the $i^{\text{th}}$ column of $\mathbf{W}$, represents the target word $w_i$; and $\mathbf{c}_j \in \mathbb{R}^d$, the $j^{\text{th}}$ column of $\mathbf{C}$, represents the context word $c_j$. Levy and Goldberg (2014) show that the loss function of Word2Vec is given by

$$\ell_{\text{W2V}} = -\sum_{i=1}^{n}\sum_{j=1}^{n} \#(w_i, c_j) \log \sigma(\mathbf{w}_i^\top \mathbf{c}_j) + \frac{k}{D}\#(w_i)\#(c_j)\log(\sigma(-\mathbf{w}_i^\top \mathbf{c}_j)), \quad (2.1)$$

which is minimised when

$$\mathbf{w}_i^\top \mathbf{c}_j = \underbrace{\log \frac{p(w_i, c_j)}{p(w_i)p(c_j)}}_{\mathbf{P}_{i,j}} - \log k = \mathbf{S}_{i,j}, \quad (2.2)$$

where $\mathbf{P}_{i,j} = \log \frac{p(w_i, c_j)}{p(w_i)p(c_j)} = \log \frac{p(c_j|w_i)}{p(c_j)}$ is *pointwise mutual information* (PMI), $\mathbf{S}_{i,j}$ denotes *shifted* PMI and the $-\log k$ shift term is an artefact of the negative sampling of the Word2Vec algorithm. In matrix form, this equates to $\mathbf{W}$ and $\mathbf{C}$ factorising $\mathbf{S} \in \mathbb{R}^{n \times n}$, i.e. $\mathbf{W}^\top \mathbf{C} = \mathbf{S}$. PMI is an information-theoretic measure of association between a pair of discrete outcomes (target and context words $w_i$ and $c_j$ respectively in Word2Vec). A PMI value of zero signifies independence of a pair of words, a positive PMI value indicates that one word appearing in text makes another word more likely to appear, and a negative PMI value indicates that one word makes another less likely.

**GloVe** (Pennington et al., 2014) is a context-free word embedding model with the same architecture as Word2Vec, but a different loss function, minimised when

$$\mathbf{w}_i^\top \mathbf{c}_j = \log p(w_i, c_j) - b_i - b_j + \log Z, \quad (2.3)$$

for biases $b_i$, $b_j$ and normalising constant $Z$. The biases in GloVe provide added flexibility compared to Word2Vec, broadening the family of statistical relationships that GloVe embeddings can learn.

### 2.3.2 From Word Embedding Geometry to Semantics

It has been empirically observed that Word2Vec word embeddings exhibit certain semantic relationships: semantically similar words have similar embeddings and the embeddings of words that constitute an *analogy* often form a parallelogram structure. For example, the embeddings of the canonical analogy "*man* is to *king* as *woman* is to *queen*" satisfy the *common vector offset* $\mathbf{r} \approx \mathbf{w}_{\text{king}} - \mathbf{w}_{\text{man}} \approx \mathbf{w}_{\text{queen}} - \mathbf{w}_{\text{woman}}$ relationship (Mikolov et al., 2013b). Drawing a parallel between word embeddings and knowledge graph entity embeddings, Bordes et al. (2013) utilise the empirically observed vector offset $\mathbf{r}$ of analogies as a basis for modelling the relation representation of the TransE model. However, until recently, a rigorous mathematical explanation for the emergence of semantics in word embedding interactions has been lacking.

Word meaning gives rise to particular word usage patterns in text, which is reflected in word co-occurrence statistics and thus in the corresponding conditional probabilities $p(w_i|c_j)$. Building upon the work of Levy and Goldberg (2014), Allen et al. (2019) recently show how the semantic relationships of *similarity*, *paraphrase* and *analogy* are encoded in the geometric interactions of the PMI-based word embeddings $\mathbf{w}_i \in \mathbb{R}^d$. Central to their theory are the so-called *PMI vectors* $\mathbf{p}_i \in \mathbb{R}^n$ — the high dimensional rows of the PMI matrix $\mathbf{P} \in \mathbb{R}^{n \times n}$. Allen et al. (2019) first show how semantic relationships correspond to linear geometric interactions between PMI vectors — *multiplicative* interactions between word co-occurrence probabilities correspond to *additive* interactions between logarithmic PMI statistics, which results in additive interactions between PMI vectors (subject to defined error terms). They further show that word embeddings are order-preserving, low-rank projections of PMI vectors, thus explaining why semantic relationships manifest as linear geometric relationships between word embeddings. This is the reason why semantically similar words, i.e. those that have similar co-occurrence distributions defined over the vocabulary $\mathcal{V}$, have similar embeddings. Further, where multiple word pairs share a common difference in their co-occurrence distributions, they form an analogy whose embeddings satisfy the common vector offset relationship, explaining the above mentioned empirical findings.

Note that although the relative geometry reflecting semantic relationships is preserved in word embeddings under the low-rank projection, the direct interpretability of dimensions (as seen in PMI vectors where each element $p_{ij}$ of a PMI vector $\mathbf{p}_i$ indicates strength of word association between $w_i$ and $c_j$) is lost since the embedding matrices $\mathbf{W}$ and $\mathbf{C}$ can be arbitrarily scaled/rotated if the other is inversely transformed.

## 2.4   Hyperbolic Geometry of the Poincaré Ball

The Poincaré ball model is one of five *isometric* models of hyperbolic geometry (Cannon et al., 1997), each offering different perspectives for performing mathematical operations in hyperbolic space. The isometry means there exists a one-to-one distance-preserving mapping from the *metric space* of one model $(\mathcal{X}, d)$ onto that of another $(\mathcal{X}', d')$, where $\mathcal{X}, \mathcal{X}'$ are sets and $d, d'$ distance functions, or *metrics*, providing a notion of equivalence between the models.

The Poincaré ball $(\mathbb{B}_c^d, g^{\mathbb{B}})$ of radius $1/\sqrt{c}$ ($-c$ indicates the negative *curvature*, $c > 0$) is a $d$-dimensional manifold $\mathbb{B}_c^d = \{\mathbf{x} \in \mathbb{R}^d : c\|\mathbf{x}\|^2 < 1\}$ equipped with the Riemannian metric $g^{\mathbb{B}}$ which is *conformal* to the Euclidean metric $g^{\mathbb{E}} = \mathbf{I}_d$ (i.e. angle-preserving with respect to the Euclidean space) with the *conformal factor* $\lambda_\mathbf{x}^c = 2/(1 - c\|\mathbf{x}\|^2)$, i.e. $g^{\mathbb{B}} = (\lambda_\mathbf{x}^c)^2 g^{\mathbb{E}}$. The distance between two points $\mathbf{x}, \mathbf{y} \in \mathbb{B}_c^d$ is measured along a *geodesic* (i.e. shortest path between the points) and is given by:

$$d_{\mathbb{B}}(\mathbf{x}, \mathbf{y}) = \frac{2}{\sqrt{c}} \tanh^{-1}(\sqrt{c}\| - \mathbf{x} \oplus_c \mathbf{y}\|), \tag{2.4}$$

where $\|\cdot\|$ denotes the Euclidean norm and $\oplus_c$ *Möbius addition* (Ungar, 2001):

$$\mathbf{x} \oplus_c \mathbf{y} = \frac{(1 + 2c\langle \mathbf{x}, \mathbf{y} \rangle + c\|\mathbf{y}\|^2)\mathbf{x} + (1 - c\|\mathbf{x}\|^2)\mathbf{y}}{1 + 2c\langle \mathbf{x}, \mathbf{y} \rangle + c^2\|\mathbf{x}\|^2\|\mathbf{y}\|^2}, \tag{2.5}$$

with $\langle \cdot, \cdot \rangle$ being the Euclidean inner product.

Each point $\mathbf{x} \in \mathbb{B}_c^d$ has a *tangent space* $T_\mathbf{x}\mathbb{B}_c^d$, a $d$-dimensional vector space, that is a local first-order approximation of the manifold $\mathbb{B}_c^d$ around $\mathbf{x}$, which for the Poincaré ball $\mathbb{B}_c^d$ is a $d$-dimensional Euclidean space, i.e. $T_\mathbf{x}\mathbb{B}_c^d = \mathbb{R}^d$. The *exponential map* $\exp_\mathbf{x}^c : T_\mathbf{x}\mathbb{B}_c^d \to \mathbb{B}_c^d$ allows one to move on the manifold from $\mathbf{x}$ in the direction of a vector $\mathbf{v} \in T_\mathbf{x}\mathbb{B}_c^d$, tangential to $\mathbb{B}_c^d$ at $\mathbf{x}$. The inverse is the *logarithmic map* $\log_\mathbf{x}^c : \mathbb{B}_c^d \to T_\mathbf{x}\mathbb{B}_c^d$. For the Poincaré ball, these are defined (Ganea et al., 2018a) as:

$$\exp_\mathbf{x}^c(\mathbf{v}) = \mathbf{x} \oplus_c \left( \tanh\left( \sqrt{c}\frac{\lambda_\mathbf{x}^c\|\mathbf{v}\|}{2} \right) \frac{\mathbf{v}}{\sqrt{c}\|\mathbf{v}\|} \right), \tag{2.6}$$

$$\log_\mathbf{x}^c(\mathbf{y}) = \frac{2}{\sqrt{c}\lambda_\mathbf{x}^c} \tanh^{-1}(\sqrt{c}\| - \mathbf{x} \oplus_c \mathbf{y}\|) \frac{-\mathbf{x} \oplus_c \mathbf{y}}{\| - \mathbf{x} \oplus_c \mathbf{y}\|}, \tag{2.7}$$

Ganea et al. (2018a) show that matrix-vector multiplication in hyperbolic space (*Möbius matrix-vector multiplication*) can be obtained by projecting a point $\mathbf{x} \in \mathbb{B}_c^d$ onto the tangent space at $\mathbf{0} \in \mathbb{B}_c^d$ with $\log_\mathbf{0}^c(\mathbf{x})$, performing matrix multiplication by $\mathbf{M} \in \mathbb{R}^{d \times k}$ in the Euclidean tangent space, and projecting back to $\mathbb{B}_c^d$ via the exponential map at $\mathbf{0}$:

$$\mathbf{M} \otimes_c \mathbf{x} = \exp_\mathbf{0}^c(\mathbf{M}\log_\mathbf{0}^c(\mathbf{x})). \tag{2.8}$$

# Chapter 3

# Hypernetwork Knowledge Graph Embeddings

The main contribution of this chapter is the paper *Hypernetwork Knowledge Graph Embeddings* which was published at the *International Conference on Artificial Neural Networks* in September 2019. We first outline the motivation for this work (Section 3.1), followed by the paper itself (Section 3.2), the impact that it has had so far (Section 3.3), and discussion (Section 3.4).

## 3.1 Introduction

The work presented in this section was largely inspired by the first convolutional knowledge graph representation method ConvE (Dettmers et al., 2018), which at the time of its introduction, outperformed all the simple embedding methods to achieve new state-of-the-art on the link prediction task. These results indicated, somewhat surprisingly, that convolution plays an important role in knowledge graph representation. ConvE performs a 2D convolution operation on the subject entity embedding $\mathbf{e}_s \in \mathbb{R}^d$ and the relation embedding $\mathbf{r} \in \mathbb{R}^d$, after they are each reshaped to a matrix and lengthwise concatenated. However, it is highly unintuitive why convolution — particularly 2D convolution — should be effective for extracting information from 1D entity and relation embeddings. Convolution is widely used in computer vision, where feature maps reflect patterns in the image such as lines or edges. Its role in the knowledge graph representation domain is harder to interpret, since little is known of the meaning of a single dimension in an entity embedding. Thus, we decided to study ConvE in an attempt to determine the role of convolution in knowledge graph representation and examine the potential benefits of imposing 2D structure on entity and relation embeddings, which resulted in the paper presented in Section 3.2.

The following section introduces *HypER*, a simplified convolutional knowledge graph representation method, which outperforms ConvE to achieve state-of-the-art results across all standard link prediction datasets. HypER removes the arbitrary reshaping and concatenation of entity and relation embeddings in ConvE and replaces 2D with 1D convolution, thus avoiding implying any inherent 2D structure in the embeddings. We further show that 1D convolution is equivalent to multiplication by a highly sparse tensor with tied weights. The multiplication of this "convolutional tensor" and other weights gives an *implicit relation matrix*, relating HypER to tensor factorisation approaches of RESCAL (Nickel et al., 2011), DistMult (Yang et al., 2015) and ComplEx (Trouillon et al., 2016). This suggests that convolution provides a good trade-off between expressiveness and parameter number compared to a dense network, dispelling the suggestion that 2D structure exists in entity embeddings implied by ConvE.

## 3.2   The Paper

### Author Contributions

The paper is co-authored by myself, Carl Allen and Timothy Hospedales. Timothy Hospedales seeded the idea of a hypernetwork model inspired by ConvE, supervised the project, offered suggestions, and helped revise the final version. As the lead author, I developed HypER, performed the experiments, and wrote the paper. Carl Allen helped with the interpretation of HypER in terms of tensor operations, participated in various discussions and proofread the paper.

### Differences in Notation

The paper defines subject and object entity embeddings as $\mathbf{e}_1$ and $\mathbf{e}_2$, whereas they are denoted as $\mathbf{e}_s$ and $\mathbf{e}_o$ respectively in the remainder of this thesis.

# Hypernetwork Knowledge Graph Embeddings

Ivana Balažević[1*], Carl Allen[1], and Timothy M. Hospedales[1,2]

[1] School of Informatics, University of Edinburgh, UK
[2] Samsung AI Centre, Cambridge, UK
{ivana.balazevic, carl.allen, t.hospedales}@ed.ac.uk

**Abstract.** Knowledge graphs are graphical representations of large databases of facts, which typically suffer from incompleteness. Inferring missing relations (links) between entities (nodes) is the task of link prediction. A recent state-of-the-art approach to link prediction, ConvE, implements a convolutional neural network to extract features from concatenated subject and relation vectors. Whilst results are impressive, the method is unintuitive and poorly understood. We propose a hypernetwork architecture that generates simplified relation-specific convolutional filters that (i) outperforms ConvE and all previous approaches across standard datasets; and (ii) can be framed as tensor factorization and thus set within a well established family of factorization models for link prediction. We thus demonstrate that convolution simply offers a convenient computational means of introducing sparsity and parameter tying to find an effective trade-off between non-linear expressiveness and the number of parameters to learn.

## 1   Introduction

Knowledge graphs, such as WordNet, Freebase, and Google Knowledge Graph, are large graph-structured databases of facts, containing information in the form of triples $(e_1, r, e_2)$, with $e_1$ and $e_2$ representing subject and object entities and $r$ a relation between them. They are considered important information resources, used for a wide variety of tasks ranging from question answering to information retrieval and text summarization. One of the main challenges with existing knowledge graphs is their incompleteness: many of the links between entities in the graph are missing. This has inspired substantial work in the field of *link prediction*, i.e. the task of inferring missing links in knowledge graphs.

Until recently, many approaches to link prediction have been based on different factorizations of a 3-moded binary tensor representation of the training triples [12, 17, 23, 22]. Such approaches are shallow and linear, with limited expressiveness. However, attempts to increase expressiveness with additional fully connected layers and non-linearities often lead to overfitting [12, 17]. For this reason, Dettmers et al. introduce ConvE, a model that uses 2D convolutions over reshaped and concatenated entity and relation embeddings [3]. They motivate the use of convolutions by being parameter efficient and fast to compute on a GPU, as well as having various robust methods from computer vision to prevent

2      Balažević et al.

overfitting. Even though results achieved by ConvE are impressive, it is highly unintuitive that convolution – particularly 2D convolution – should be effective for extracting information from 1D entity and relation embeddings.

In this paper, we introduce HypER, a model that uses a *hypernetwork* [5] to generate convolutional filter weights for each relation. A hypernetwork is an approach by which one network generates weights for another network, that can be used to enable weight-sharing across layers and to dynamically synthesize weights given an input. In our context, we generate relation-specific filter weights to process input entities, and also achieve *multi-task knowledge sharing* across relations in the knowledge graph. Our proposed HypER model uses a hypernetwork to generate a set of 1D relation-specific filters to process the subject entity embeddings. This simplifies the interaction between subject entity and relation embeddings compared to ConvE, in which a global set of *2D* filters are convolved over *reshaped and concatenated* subject entity and relation embeddings, which is unintuitive as it suggests the presence of 2D structure in word embeddings. Moreover, interaction between subject and relation in ConvE depends on an arbitrary choice about how they are reshaped and concatenated. In contrast, HypER's hypernetwork generates relation-specific filters, and thus extracts *relation-specific features* from the subject entity embedding. This necessitates no 2D reshaping, and allows entity and relation to interact more completely, rather than only around the concatenation boundary. We show that this simplified approach, in addition to improving link prediction performance, can be understood in terms of tensor factorization, thus placing HypER within a well established family of factorization models. The apparent obscurity of using convolution within word embeddings is thereby explained as simply a convenient computational means of introducing *sparsity* and *parameter tying*.

We evaluate HypER against several previously proposed link prediction models using standard datasets (FB15k-237, WN18RR, FB15k, WN18, YAGO3-10), across which it consistently achieves state-of-the-art performance. In summary, our key contributions are:

–  proposing a new model for link prediction (HypER) which achieves state-of-the-art performance across all standard datasets;
–  showing that the benefit of using convolutional instead of fully connected layers is due to restricting the number of dimensions that interact (i.e. explicit regularization), rather than finding higher dimensional structure in the embeddings (as implied by ConvE); and
–  showing that HypER in fact falls within a broad class of tensor factorization models despite the use of convolution, which serves to provide a good trade-off between expressiveness and number of parameters to learn.

## 2   Related Work

Numerous matrix factorization approaches to link prediction have been proposed. An early model, RESCAL [12], tackles the link prediction task by optimizing a scoring function containing a bilinear product between vectors for each of the subject and object entities and a full rank matrix for each relation. Dist-Mult [23] can be viewed as a special case of RESCAL with a diagonal matrix per relation type, which limits the linear transformation performed on entity vectors to a stretch. ComplEx [22] extends DistMult to the complex domain. TransE [1] is an affine model that represents a relation as a translation operation between subject and object entity vectors.

A somewhat separate line of link prediction research introduces Relational Graph Convolutional Networks (R-GCNs) [15]. R-GCNs use a convolution operator to capture locality information in graphs. The model closest to our own and which we draw inspiration from, is ConvE [3], where a convolution operation is performed on the subject entity vector and the relation vector, after they are each reshaped to a matrix and lengthwise concatenated. The obtained feature maps are flattened, put through a fully connected layer, and the inner product is taken with all object entity vectors to generate a score for each triple. Advantages of ConvE over previous approaches include its expressiveness, achieved by using multiple layers of non-linear features, its scalability to large knowledge graphs, and its robustness to overfitting. However, it is not intuitive why convolving across concatenated and reshaped subject entity and relation vectors should be effective.

The proposed HypER model does no such reshaping or concatenation and thus avoids both implying any inherent 2D structure in the embeddings and restricting interaction to the concatenation boundary. Instead, HypER convolves *every* dimension of the subject entity embedding with *relation-specific* convolutional filters generated by the hypernetwork. This way, entity and relation embeddings are combined in a non-linear (quadratic) manner, unlike the linear combination (weighted sum) in ConvE. This gives HypER more expressive power, while also reducing parameters.

Interestingly, we find that the differences in moving from ConvE to HypER in fact bring the factorization and convolutional approaches together, since the 1D convolution process is equivalent to multiplication by a highly sparse tensor with tied weights (see Figure 2). The multiplication of this "convolutional tensor" (defined by the relation embedding and hypernetwork) and other weights gives an implicit relation matrix, corresponding to those in e.g. RESCAL, DistMult and ComplEx. Other than the method of deriving these relation matrices, the key difference to existing factorization approaches is the ReLU non-linearity applied prior to interaction with the object embedding.

## 3  Link Prediction

In link prediction, the aim is to learn a scoring function $\phi$ that assigns a score $s = \phi(e_1, r, e_2) \in \mathbb{R}$ to each input triple $(e_1, r, e_2)$, where $e_1, e_2 \in \mathcal{E}$ are subject and object entities and $r \in \mathcal{R}$ a relation. The score indicates the strength of prediction that the given triple corresponds to a true fact, with positive scores meaning true and negative scores, false. Link prediction models typically map entity pair $e_1, e_2$ to their corresponding distributed embedding representations $\mathbf{e}_1, \mathbf{e}_2 \in \mathbb{R}^{d_e}$ and a score is assigned using a *relation-specific* function,

4      Balažević et al.

**Table 1.** Scoring functions of state-of-the-art link prediction models, the dimensionality of their relation parameters, and their space complexity. $d_e$ and $d_r$ are the dimensions of entity and relation embeddings respectively, $\bar{\mathbf{e}}_2 \in \mathbb{C}^{d_e}$ denotes the complex conjugate of $\mathbf{e}_2$, and $\underline{\mathbf{e}}_1, \underline{\mathbf{w}}_r \in \mathbb{R}^{d_w \times d_h}$ denote a 2D reshaping of $\mathbf{e}_1$ and $\mathbf{w}_r$ respectively. $*$ is the convolution operator, $\mathbf{F}_r = \mathrm{vec}^{-1}(\mathbf{w}_r\mathbf{H})$ the matrix of relation specific convolutional filters, vec is a vectorization of a matrix and $\mathrm{vec}^{-1}$ its inverse, $f$ is a non-linear function, and $n_e$ and $n_r$ respectively denote the number of entities and relations.

| Model | Scoring Function | Relation Parameters | Space Complexity |
|-------|------------------|---------------------|------------------|
| RESCAL [12] | $\mathbf{e}_1^\top \mathbf{W}_r \mathbf{e}_2$ | $\mathbf{W}_r \in \mathbb{R}^{d_e{}^2}$ | $\mathcal{O}(n_e d_e + n_r d_e^2)$ |
| TransE [1] | $\|\mathbf{e}_1 + \mathbf{w}_r - \mathbf{e}_2\|$ | $\mathbf{w}_r \in \mathbb{R}^{d_e}$ | $\mathcal{O}(n_e d_e + n_r d_e)$ |
| NTN [17] | $\mathbf{u}_r^\top f(\mathbf{e}_1\mathbf{W}_r^{[1..k]}\mathbf{e}_2 + \mathbf{V}_r \begin{bmatrix} \mathbf{e}_1 \\ \mathbf{e}_2 \end{bmatrix} + \mathbf{b}_r)$ | $\mathbf{W}_r \in \mathbb{R}^{d_e{}^2 k}, \mathbf{V}_r \in \mathbb{R}^{2d_e k},$ $\mathbf{u}_r \in \mathbb{R}^k, \mathbf{b}_r \in \mathbb{R}^k$ | $\mathcal{O}(n_e d_e + n_r d_e^2 k)$ |
| DistMult [23] | $\langle \mathbf{e}_1, \mathbf{w}_r, \mathbf{e}_2 \rangle$ | $\mathbf{w}_r \in \mathbb{R}^{d_e}$ | $\mathcal{O}(n_e d_e + n_r d_e)$ |
| ComplEx [22] | $\mathrm{Re}(\langle \mathbf{e}_1, \mathbf{w}_r, \bar{\mathbf{e}}_2 \rangle)$ | $\mathbf{w}_r \in \mathbb{C}^{d_e}$ | $\mathcal{O}(n_e d_e + n_r d_e)$ |
| ConvE [3] | $f(\mathrm{vec}(f(\underline{\mathbf{e}}_1 * \underline{\mathbf{w}}_r) * w))\mathbf{W})\mathbf{e}_2$ | $\mathbf{w}_r \in \mathbb{R}^{d_r}$ | $\mathcal{O}(n_e d_e + n_r d_r)$ |
| HypER (ours) | $f(\mathrm{vec}(\mathbf{e}_1 * \mathrm{vec}^{-1}(\mathbf{w}_r\mathbf{H}))\mathbf{W})\mathbf{e}_2$ | $\mathbf{w}_r \in \mathbb{R}^{d_r}$ | $\mathcal{O}(n_e d_e + n_r d_r)$ |

$s = \phi_r(\mathbf{e}_1, \mathbf{e}_2)$. The majority of link prediction models apply the logistic sigmoid function $\sigma(\cdot)$ to the score to give a probabilistically interpretable prediction $p = \sigma(s) \in [0,1]$ as to whether the queried fact is true. The scoring functions for models from across the literature and HypER are summarized in Table 1, together with the dimensionality of their relation parameters and the significant terms of their space complexity.

## 4    Hypernetwork Knowledge Graph Embeddings

In this work, we propose a novel hypernetwork model for link prediction in knowledge graphs. In summary, the hypernetwork projects a vector embedding of each relation via a fully connected layer, the result of which is reshaped to give a set of convolutional filter weight vectors for each relation. We explain this process in more detail below. The idea of using convolutions on entity and relation embeddings stems from computer vision, where feature maps reflect patterns in the image such as lines or edges. Their role in the text domain is harder to interpret, since little is known of the meaning of a single dimension in a word embedding. We believe convolutional filters have a regularizing effect when applied to word embeddings (compared to the corresponding full tensor), as the filter size restricts which dimensions of embeddings can interact. This allows nonlinear expressiveness while limiting overfitting by using few parameters. A visualization of HypER is given in Figure 1.

### 4.1    Scoring Function and Model Architecture

The relation-specific scoring function for the HypER model is:

$$\begin{aligned} \phi_r(\mathbf{e}_1, \mathbf{e}_2) &= f(\mathrm{vec}(\mathbf{e}_1 * \mathbf{F}_r)\mathbf{W})\mathbf{e}_2 \\ &= f(\mathrm{vec}(\mathbf{e}_1 * \mathrm{vec}^{-1}(\mathbf{w}_r\mathbf{H}))\mathbf{W})\mathbf{e}_2, \end{aligned} \quad (1)$$

where the $\mathrm{vec}^{-1}$ operator reshapes a vector to a matrix, and non-linearity $f$ is chosen to be a rectified linear unit (ReLU).

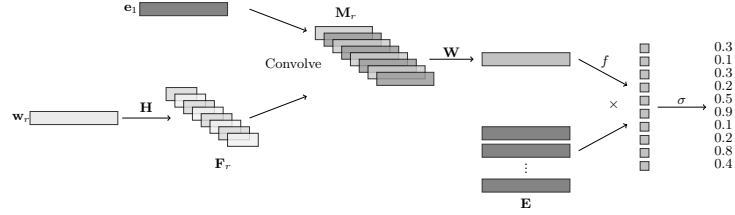

**Fig. 1.** Visualization of the HypER model architecture. Subject entity embedding $\mathbf{e}_1$ is convolved with filters $\mathbf{F}_r$, created by the hypernetwork $\mathbf{H}$ from relation embedding $\mathbf{w}_r$. The obtained feature maps $\mathbf{M}_r$ are mapped to $d_e$-dimensional space via $\mathbf{W}$ and the non-linearity $f$ applied before being combined with all object vectors $\mathbf{e}_2 \in \mathbf{E}$ through an inner product to give a score for each triple. Predictions are obtained by applying the logistic sigmoid function to each score.

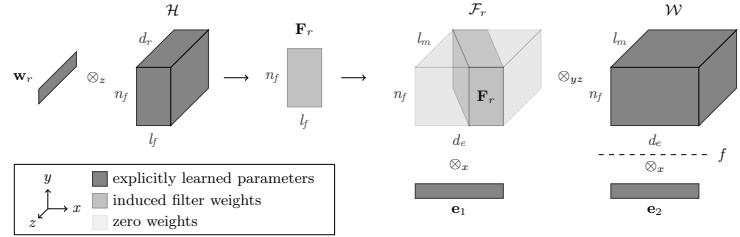

**Fig. 2.** Interpretation of the HypER model in terms of tensor operations. Each relation embedding $\mathbf{w}_r$ generates a set of filters $\mathbf{F}_r$ via the hypernetwork $\mathcal{H}$. The act of convolving $\mathbf{F}_r$ over $\mathbf{e}_1$ is equivalent to multiplication of $\mathbf{e}_1$ by a tensor $\mathcal{F}_r$ (in which $\mathbf{F}_r$ is diagonally duplicated and zero elsewhere). The tensor product $\mathcal{F}_r \otimes_{yz} \mathcal{W}$ gives a $d_e \times d_e$ matrix specific to each relation. Axes labels indicate the modes of tensor interaction (via inner product).

In the feed-forward pass, the model obtains embeddings for the input triple from the entity and relation embedding matrices $\mathbf{E} \in \mathbb{R}^{n_e \times d_e}$ and $\mathbf{R} \in \mathbb{R}^{n_r \times d_r}$. The hypernetwork is a fully connected layer $\mathbf{H} \in \mathbb{R}^{d_r \times l_f n_f}$ ($l_f$ denotes filter length and $n_f$ the number of filters per relation, i.e. *output channels* of the convolution) that is applied to the relation embedding $\mathbf{w}_r \in \mathbb{R}^{d_r}$. The result is reshaped to generate a matrix of convolutional filters $\mathbf{F}_r = \text{vec}^{-1}(\mathbf{w}_r \mathbf{H}) \in \mathbb{R}^{l_f \times n_f}$. Whilst the overall dimensionality of the filter set is $l_f n_f$, the rank is restricted to $d_r$ to encourage parameter sharing between relations.

The subject entity embedding $\mathbf{e}_1$ is convolved with the set of relation-specific filters $\mathbf{F}_r$ to give a 2D feature map $\mathbf{M}_r \in \mathbb{R}^{l_m \times n_f}$, where $l_m = d_e - l_f + 1$ is the feature map length. The feature map is vectorized to $\text{vec}(\mathbf{M}_r) \in \mathbb{R}^{l_m n_f}$, and projected to $d_e$-dimensional space by the weight matrix $\mathbf{W} \in \mathbb{R}^{l_m n_f \times d_e}$. After applying a ReLU activation function, the result is combined by way of inner product with each and every object entity embedding $\mathbf{e}_2^{(i)}$, where $i$ varies over all entities in the dataset (of size $n_e$), to give a vector of scores. The logistic sigmoid is applied element-wise to the score vector to obtain the predicted probability of each prospective triple being true $\mathbf{p}_i = \sigma(\phi_r(\mathbf{e}_1, \mathbf{e}_2^{(i)}))$.

### 4.2   Understanding HypER as Tensor Factorization

Having described the HypER architecture, we can view it as a series of tensor operations by considering the hypernetwork $\mathbf{H}$ and weight matrix $\mathbf{W}$ as tensors $\mathcal{H} \in \mathbb{R}^{d_r \times l_f \times n_f}$ and $\mathcal{W} \in \mathbb{R}^{l_m \times n_f \times d_e}$ respectively. The act of convolving $\mathbf{F}_r = \mathbf{w}_r \otimes \mathcal{H}$ over the subject entity embedding $\mathbf{e}_1$ is equivalent to the multiplication of $\mathbf{e}_1$ by a sparse tensor $\mathcal{F}_r$ within which $\mathbf{F}_r$ is diagonally duplicated with zeros elsewhere (see Figure 2). The result is multiplied by $\mathcal{W}$ to give a vector, which is subject to ReLU before the final dot product with $\mathbf{e}_2$. Linearity allows the product $\mathcal{F}_r \otimes \mathcal{W}$ to be considered separately as generating a $d_e \times d_e$ matrix for each relation. Further, rather than duplicating entries of $\mathbf{F}_r$ within $\mathcal{F}_r$, we can generalize $\mathcal{F}_r$ to a relation-agnostic sparse 4 moded tensor $\mathcal{F} \in \mathbb{R}^{d_r \times d_e \times n_f \times l_m}$ by replacing entries with $d_r$-dimensional strands of $\mathcal{H}$. Thus, the HypER model can be described explicitly as tensor multiplication of $\mathbf{e}_1, \mathbf{e}_2$ and $\mathbf{w}_r$ with a core tensor $\mathcal{F} \otimes \mathcal{W} \in \mathbb{R}^{d_e \times d_e \times d_r}$, where $\mathcal{F}$ is heavily constrained in terms of its number of free variables. This insight allows HypER to be viewed in a very similar light to the family of factorization approaches to link prediction, such as RESCAL, DistMult and ComplEx.

### 4.3   Training Procedure

Following the training procedure introduced by [3], we use *1-N scoring* with the Adam optimizer [8] to minimize the binary cross-entropy loss:

$$\mathcal{L}(\mathbf{p}, \mathbf{y}) = -\frac{1}{n_e} \sum_i (\mathbf{y}_i \log(\mathbf{p}_i) + (1 - \mathbf{y}_i) \log(1 - \mathbf{p}_i)), \tag{2}$$

where $\mathbf{y} \in \mathbb{R}^{n_e}$ is the label vector containing ones for true triples and zeros otherwise, subject to *label smoothing*. **Label smoothing** is a widely used technique shown to improve generalization [20, 14]. Label smoothing changes the ground-truth label distribution by adding a uniform prior to encourage the model to be less confident, achieving a regularizing effect. **1-N scoring** refers to simultaneously scoring $(e_1, r, \mathcal{E})$, i.e. for all entities $e_2 \in \mathcal{E}$, in contrast to *1-1 scoring*, the practice of training individual triples $(e_1, r, e_2)$ one at a time. As shown by [3], 1-N scoring offers a significant speedup (3x on train and 300x on test time) and improved accuracy compared to 1-1 scoring. A potential extension of the HypER model described above would be to apply convolutional filters to *both* subject and object entity embeddings. However, since this is not trivially implementable with 1-N scoring and wanting to keep its benefits, we leave this to future work.

### 4.4   Number of Parameters

Table 2 compares the number of parameters of ConvE and HypER (for the FB15k-237 dataset, which determines $n_e$ and $n_r$). It can be seen that, overall, HypER has fewer parameters (4.3M) than ConvE (5.1M) due to the way HypER directly transforms relations to convolutional filters.

**Table 2.** Comparison of number of parameters for ConvE and HypER on FB15k-237. $h_m$ and $w_m$ are height and width of the ConvE feature maps respectively.

| Model | **E** | **R** | Filters | **W** |
|-------|-------|-------|---------|-------|
| ConvE | $n_e \times d_e$ | $n_r \times d_r$ | $l_f n_f$ | $h_m w_m n_f \times d_e$ |
|       | $2.9M$ | $0.1M$ | $0.0M$ | $2.1M$ |
| HypER | $n_e \times d_e$ | $n_r \times d_r$ | $d_r \times l_f n_f$ | $l_m n_f \times d_e$ |
|       | $2.9M$ | $0.1M$ | $0.1M$ | $1.2M$ |

## 5   Experiments

### 5.1   Datasets

We evaluate our HypER model on the standard link prediction task using the following datasets (see Table 3):

**FB15k**   [1] a subset of Freebase, a large database of facts about the real world.

**WN18**   [1] a subset of WordNet, containing lexical relations between words.

**FB15k-237**   created by [21], noting that the validation and test sets of FB15k and WN18 contain the inverse of many relations present in the training set, making it easy for simple models to do well. FB15k-237 is a subset of FB15k with the inverse relations removed.

**WN18RR**   [3] a subset of WN18, created by removing the inverse relations.

**YAGO3-10**   [3] a subset of YAGO3 [10], containing entities which have a minimum of 10 relations each.

**Table 3.** Summary of dataset statistics.

| Dataset | Entities ($n_e$) | Relations ($n_r$) |
|---------|-----------------|------------------|
| FB15k | 14,951 | 1,345 |
| WN18 | 40,943 | 18 |
| FB15k-237 | 14,541 | 237 |
| WN18RR | 40,943 | 11 |
| YAGO3-10 | 123,182 | 37 |

### 5.2   Experimental Setup

We implement HypER in PyTorch [13] and make our code publicly available.[1]

**Implementation Details**   We train our model with 200 dimension entity and relation embeddings ($d_e = d_r = 200$) and 1-N scoring. Whilst the relation embedding dimension does not have to equal the entity embedding dimension, we set $d_r = 200$ to match ConvE for fairness of comparison.

To accelerate training and prevent overfitting, we use batch normalization [6] and dropout [18] on the input embeddings, feature maps and the hidden layer. We perform a hyperparameter search and select the best performing model by mean reciprocal rank (MRR) on the validation set. Having tested the values $\{0., 0.1, 0.2, 0.3\}$, we find that the following combination of parameters works well across all datasets: input dropout 0.2, feature map dropout 0.2, and hidden dropout 0.3, apart from FB15k-237, where we set input dropout to 0.3. We select the learning rate from $\{0.01, 0.005, 0.003, 0.001, 0.0005, 0.0001\}$ and exponential learning rate decay from $\{1., 0.99, 0.995\}$ for each dataset and find the best performing learning rate and learning rate decay to be dataset-specific. We set the convolution stride to 1, number of feature maps to 32 with the filter size $3 \times 3$ for ConvE and $1 \times 9$ for HypER, after testing different numbers of feature maps $n_f \in \{16, 32, 64\}$ and filter sizes $l_f \in \{1 \times 1, 1 \times 2, 1 \times 3, 1 \times 6, 1 \times 9, 1 \times 12\}$ (see Table 9). We train all models using the Adam optimizer with batch size 128. One epoch on FB15k-237 takes approximately 12 seconds on a single GPU compared to 1 minute for e.g. RESCAL, largely due to 1-N scoring.

---

[1] `https://github.com/ibalazevic/HypER`

**Evaluation**    Results are obtained by iterating over all triples in the test set. A particular triple is evaluated by replacing the object entity $e_2$ with all entities $\mathcal{E}$ while keeping the subject entity $e_1$ fixed and vice versa, obtaining scores for each combination. These scores are then ranked using the "filtered" setting only, i.e. we remove all true cases other than the current test triple [1].

We evaluate HypER on five different metrics found throughout the link prediction literature: mean rank (MR), mean reciprocal rank (MRR), hits@10, hits@3, and hits@1. Mean rank is the average rank assigned to the true triple, over all test triples. Mean reciprocal rank takes the average of the reciprocal rank assigned to the true triple. Hits@k measures the percentage of cases in which the true triple appears in the top k ranked triples. Overall, the aim is to achieve high mean reciprocal rank and hits@k and low mean rank. For a more extensive description of how each of these metrics is calculated, we refer to [3].

### 5.3   Results

Link prediction results for all models across the five datasets are shown in Tables 4, 5 and 6. Our key findings are:

- whilst having fewer parameters than the closest comparator ConvE, HypER consistently outperforms all other models across all datasets, thereby achieving state-of-the-art results on the link prediction task; and
- our filter dimension study suggests that no benefit is gained by convolving over reshaped 2D entity embeddings in comparison with 1D entity embedding vectors and that most information can be extracted with very small convolutional filters (Table 9).

Overall, HypER outperforms all other models on all metrics apart from mean reciprocal rank on WN18 and mean rank on WN18RR, FB15k-237, WN18, and YAGO3-10. Given that mean rank is known to be highly sensitive to outliers [11], this suggests that HypER correctly ranks many true triples in the top 10, but makes larger ranking errors elsewhere.

Given that most models in the literature, with the exception of ConvE, were trained with 100 dimension embeddings and 1-1 scoring, we reimplement previous models (DistMult, ComplEx and ConvE) with 200 dimension embeddings and 1-N scoring for fair comparison and report the obtained results on WN18RR in Table 7. We perform the same hyperparameter search for every model and present the mean and standard deviation of each result across five runs (different random seeds). This improves most previously published results, except for ConvE where we fail to replicate some values. Notwithstanding, HypER remains the best performing model overall despite better tuning of the competitors.

**Table 4.** Link prediction results on WN18RR and FB15k-237. The RotatE [19] results are reported without their self-adversarial negative sampling (see Appendix H in the original paper) for fair comparison, given that it is not specific to that model only.

| | WN18RR | | | | | FB15k-237 | | | | |
|---|---|---|---|---|---|---|---|---|---|---|
| | MR | MRR | H@10 | H@3 | H@1 | MR | MRR | H@10 | H@3 | H@1 |
| DistMult [23] | 5110 | .430 | .490 | .440 | .390 | 254 | .241 | .419 | .263 | .155 |
| ComplEx [22] | 5261 | .440 | .510 | .460 | .410 | 339 | .247 | .428 | .275 | .158 |
| Neural LP [24] | – | – | – | – | – | .250 | .408 | – | – | – |
| R-GCN [15] | – | – | – | – | – | – | .248 | .417 | .264 | .151 |
| MINERVA [2] | – | – | – | – | – | – | .456 | – | – | – |
| ConvE [3] | **4187** | .430 | .520 | .440 | .400 | 244 | .325 | .501 | .356 | .237 |
| M-Walk [16] | – | .437 | – | .445 | .414 | – | – | – | – | – |
| RotatE [19] | – | – | – | – | – | **185** | .297 | .480 | .328 | .205 |
| HypER (ours) | 5798 | **.465** | **.522** | **.477** | **.436** | 250 | **.341** | **.520** | **.376** | **.252** |

**Table 5.** Link prediction results on WN18 and FB15k.

| | WN18 | | | | | FB15k | | | | |
|---|---|---|---|---|---|---|---|---|---|---|
| | MR | MRR | H@10 | H@3 | H@1 | MR | MRR | H@10 | H@3 | H@1 |
| TransE [1] | **251** | – | .892 | – | – | 125 | – | .471 | – | – |
| DistMult [23] | 902 | .822 | .936 | .914 | .728 | 97 | .654 | .824 | .733 | .546 |
| ComplEx [22] | – | .941 | .947 | .936 | .936 | – | .692 | .840 | .759 | .599 |
| ANALOGY [9] | – | .942 | .947 | .944 | .939 | – | .725 | .854 | .785 | .646 |
| Neural LP [24] | – | .940 | .945 | – | – | – | .760 | .837 | – | – |
| R-GCN [15] | – | .819 | **.964** | .929 | .697 | – | .696 | .842 | .760 | .601 |
| TorusE [4] | – | .947 | .954 | .950 | .943 | – | .733 | .832 | .771 | .674 |
| ConvE [3] | 374 | .943 | .956 | .946 | .935 | 51 | .657 | .831 | .723 | .558 |
| SimplE [7] | – | .942 | .947 | .944 | .939 | – | .727 | .838 | .773 | .660 |
| HypER (ours) | 431 | **.951** | .958 | **.955** | **.947** | 44 | **.790** | **.885** | **.829** | **.734** |

**Table 6.** Link prediction results on YAGO3-10.

| | YAGO3-10 | | | | |
|---|---|---|---|---|---|
| | MR | MRR | H@10 | H@3 | H@1 |
| DistMult [23] | 5926 | .340 | .540 | .380 | .240 |
| ComplEx [22] | 6351 | .360 | .550 | .400 | .260 |
| ConvE [3] | **1676** | .440 | .620 | .490 | .350 |
| HypER (ours) | 2529 | **.533** | **.678** | **.580** | **.455** |

To ensure that the difference between reported results for HypER and ConvE is not simply due to HypER having a reduced number of parameters (implicit regularization), we trained ConvE reducing the number of feature maps to 16 instead of 32 to have a comparable number of parameters to HypER (explicit

10      Balaževič et al.

**Table 7.** Link prediction results on WN18RR; all models trained with 200 dimension embeddings and 1-N scoring.

| | WN18RR | | | |
|---|---|---|---|---|
| | MR | MRR | H@10 | H@3 | H@1 |
| DistMult [23] | **4911** ± **109** | .434 ± .002 | .508 ± .002 | .447 ± .001 | .399 ± .002 |
| ComplEx [22] | 5930 ± 125 | .446 ± .001 | **.523** ± **.002** | .462 ± .001 | .409 ± .001 |
| ConvE [3] | **4997** ± **99** | .431 ± .001 | .504 ± .002 | .443 ± .002 | .396 ± .001 |
| HypER (ours) | 5798 ± 124 | **.465** ± **.002** | .522 ± .003 | **.477** ± **.002** | **.436** ± **.003** |

regularization). This showed no improvement in ConvE results, indicating HypER's architecture does more than merely reducing the number of parameters.

**Table 8.** Results with and without hypernetwork on WN18RR and FB15k-237.

| | WN18RR | | FB15k-237 | |
|---|---|---|---|---|
| | MRR | H@10 | MRR | H@10 |
| HypER | **.465** ± **.002** | **.522** ± **.003** | **.341** ± **.001** | **.520** ± **.002** |
| HypER (no **H**) | .459 ± .002 | .511 ± .002 | .338 ± .001 | .515 ± .001 |

**Hypernetwork Influence**    To test the influence of the hypernetwork and, thereby, knowledge sharing between relations, we compare HypER results on WN18RR and FB15k-237 with the hypernetwork component removed, i.e. without the first fully connected layer and with the relation embeddings directly corresponding to a set of convolutional filters. Results presented in Table 8 show that the hypernetwork component improves performance, demonstrating the value of multi-task learning across different relations.

**Filter Dimension Study**    Table 9 shows results of our study investigating the influence of different convolutional filter sizes on the performance of HypER. The lower part of the table shows results for 2D filters convolved over reshaped ($10 \times 20$) 2D subject entity embeddings. It can be seen that reshaping the embeddings is of no benefit, especially on WN18RR. These results indicate that the purpose of convolution on word embeddings is not to find patterns in a 2D embedding (as with images), but perhaps to limit the number of dimensions that can interact with each other, thereby avoiding overfitting. In the upper part of the table, we vary the length of 1D filters, showing that comparable results can be achieved with filter sizes $1 \times 6$ and $1 \times 9$, with diminishing results for smaller (e.g. $1 \times 1$) and larger (e.g. $1 \times 12$) filters.

**Label Smoothing**    Contrary to the ablation study of [3], showing the influence of hyperparameters on mean reciprocal rank for FB15k-237, from which they deem label smoothing unimportant, we find label smoothing to give a significant improvement in prediction scores for WN18RR. However, we find it does have a negative influence on the FB15k scores and as such, exclude label smoothing from our experiments on that dataset. We therefore recommend evaluating the influence of label smoothing on a per dataset basis and leave to future work analysis of the utility of label smoothing in the general case.

**Table 9.** Influence of different filter dimension choices on prediction results.

| Filter Size | WN18RR | | FB15k-237 | |
|---|---|---|---|---|
| | MRR | H@1 | MRR | H@1 |
| $1 \times 1$ | .455 | .422 | .337 | .248 |
| $1 \times 2$ | .458 | .428 | .337 | .248 |
| $1 \times 3$ | .457 | .427 | .339 | .250 |
| $1 \times 6$ | .459 | .429 | .340 | .251 |
| $1 \times 9$ | **.465** | **.436** | **.341** | **.252** |
| $1 \times 12$ | .457 | .428 | **.341** | **.252** |
| $2 \times 2$ | .456 | .429 | .340 | .250 |
| $3 \times 3$ | .458 | .430 | .339 | .250 |
| $5 \times 5$ | .452 | .423 | .340 | **.252** |

## 6    Conclusion

In this work, we introduce HypER, a hypernetwork model for link prediction on knowledge graphs. HypER generates relation-specific convolutional filters and applies them to subject entity embeddings. The hypernetwork component allows information to be shared between relation vectors, enabling multi-task learning across relations. To our knowledge, HypER is the first link prediction model that creates non-linear interaction between entity and relation embeddings by convolving relation-specific filters over the entity embeddings.

We show that no benefit is gained from 2D convolutional filters over 1D, dispelling the suggestion that 2D structure exists in entity embeddings implied by ConvE. We also recast HypER in terms of tensor operations showing that, despite the convolution operation, it is closely related to the established family of tensor factorization models. Our results suggest that convolution provides a good trade-off between expressiveness and parameter number compared to a dense network. HypER is fast, robust to overfitting, has relatively few parameters, and achieves state-of-the-art results across almost all metrics on multiple link prediction datasets.

Future work might include expanding the current architecture by applying convolutional filters to both subject and object entity embeddings. We may also analyze the influence of label smoothing and explore the interpretability of convolutional feature maps to gain insight and potentially improve the model.

### Acknowledgements

We thank Ivan Titov for helpful discussions on this work. Ivana Balažević and Carl Allen were supported by the Centre for Doctoral Training in Data Science, funded by EPSRC (grant EP/L016427/1) and the University of Edinburgh.

## References

1. Bordes, A., Usunier, N., Garcia-Duran, A., Weston, J., Yakhnenko, O.: Translating Embeddings for Modeling Multi-relational Data. In: Advances in Neural Information Processing Systems (2013)

12      Balažević et al.

2. Das, R., Dhuliawala, S., Zaheer, M., Vilnis, L., Durugkar, I., Krishnamurthy, A., Smola, A., McCallum, A.: Go for a Walk and Arrive at the Answer: Reasoning over Paths in Knowledge Bases Using Reinforcement Learning. In: International Conference on Learning Representations (2018)

3. Dettmers, T., Minervini, P., Stenetorp, P., Riedel, S.: Convolutional 2D Knowledge Graph Embeddings. In: Association for the Advancement of Artificial Intelligence (2018)

4. Ebisu, T., Ichise, R.: TorusE: Knowledge Graph Embedding on a Lie Group. In: Association for Artificial Intelligence (2018)

5. Ha, D., Dai, A., Le, Q.V.: Hypernetworks. In: International Conference on Learning Representations (2017)

6. Ioffe, S., Szegedy, C.: Batch Normalization: Accelerating Deep Network Training by Reducing Internal Covariate Shift. In: International Conference on Machine Learning (2015)

7. Kazemi, S.M., Poole, D.: SimplE Embedding for Link Prediction in Knowledge Graphs. In: Advances in Neural Information Processing Systems (2018)

8. Kingma, D.P., Ba, J.: Adam: A Method for Stochastic Optimization. In: International Conference on Learning Representations (2015)

9. Liu, H., Wu, Y., Yang, Y.: Analogical Inference for Multi-relational Embeddings. In: International Conference on Machine Learning (2017)

10. Mahdisoltani, F., Biega, J., Suchanek, F.M.: Yago3: A Knowledge Base from Multilingual Wikipedias. In: Conference on Innovative Data Systems Research (2013)

11. Nickel, M., Rosasco, L., Poggio, T.A.: Holographic Embeddings of Knowledge Graphs. In: Association for the Advancement of Artificial Intelligence (2016)

12. Nickel, M., Tresp, V., Kriegel, H.P.: A Three-Way Model for Collective Learning on Multi-Relational Data. In: International Conference on Machine Learning (2011)

13. Paszke, A., Gross, S., Chintala, S., Chanan, G., Yang, E., DeVito, Z., Lin, Z., Desmaison, A., Antiga, L., Lerer, A.: Automatic Differentiation in PyTorch. In: NIPS-W (2017)

14. Pereyra, G., Tucker, G., Chorowski, J., Kaiser, L., Hinton, G.: Regularizing neural networks by penalizing confident output distributions. arXiv preprint arXiv:1701.06548 (2017)

15. Schlichtkrull, M., Kipf, T.N., Bloem, P., van den Berg, R., Titov, I., Welling, M.: Modeling Relational Data with Graph Convolutional Networks. In: European Semantic Web Conference (2018)

16. Shen, Y., Chen, J., Huang, P.S., Guo, Y., Gao, J.: M-Walk: Learning to Walk over Graphs using Monte Carlo Tree Search. In: Advances in Neural Information Processing Systems (2018)

17. Socher, R., Chen, D., Manning, C.D., Ng, A.: Reasoning with Neural Tensor Networks for Knowledge Base Completion. In: Advances in Neural Information Processing Systems (2013)

18. Srivastava, N., Hinton, G., Krizhevsky, A., Sutskever, I., Salakhutdinov, R.: Dropout: A Simple Way to Prevent Neural Networks from Overfitting. Journal of Machine Learning Research **15**(1), 1929–1958 (2014)

19. Sun, Z., Deng, Z.H., Nie, J.Y., Tang, J.: RotatE: Knowledge Graph Embedding by Relational Rotation in Complex Space. In: International Conference on Learning Representations (2019)

20. Szegedy, C., Vanhoucke, V., Ioffe, S., Shlens, J., Wojna, Z.: Rethinking the Inception Architecture for Computer Vision. In: Computer Vision and Pattern Recognition (2016)

21. Toutanova, K., Chen, D., Pantel, P., Poon, H., Choudhury, P., Gamon, M.: Representing Text for Joint Embedding of Text and Knowledge Bases. In: Empirical Methods in Natural Language Processing (2015)
22. Trouillon, T., Welbl, J., Riedel, S., Gaussier, É., Bouchard, G.: Complex Embeddings for Simple Link Prediction. In: International Conference on Machine Learning (2016)
23. Yang, B., Yih, W.t., He, X., Gao, J., Deng, L.: Embedding Entities and Relations for Learning and Inference in Knowledge Bases. In: International Conference on Learning Representations (2015)
24. Yang, F., Yang, Z., Cohen, W.W.: Differentiable Learning of Logical Rules for Knowledge Base Reasoning. In: Advances in Neural Information Processing Systems (2017)

## 3.3   Impact

According to Google Scholar, the paper has received 80 citations as of November 2021. It was featured in numerous surveys (e.g. Ji et al., 2021; Nguyen, 2020; Hogan et al., 2020; Kazemi et al., 2020) and in part inspired the TuckER (Balažević et al., 2019b) model presented in the following chapter.

## 3.4   Discussion

Apart from simplifying ConvE and improving upon its link prediction performance, the more general takeaways of the paper are twofold: (i) showing lack of advantage in using 2D convolutional filters over 1D, thus casting doubt on the existence of 2D structure in entity embeddings as implied by ConvE; and (ii) demonstrating the benefit of constrained tensor factorisation for the link prediction task.

In knowledge graph representation, entity and relation representations are typically randomly initialised, so it is perhaps unsurprising that reshaping them and convolving 2D filters over them does not bring any improvement in performance. However, it would be interesting to explore whether this finding would still hold if word embeddings or alternatively entity and relation feature vectors (where available) were used to initialise entity and relation representations, thus potentially providing insight into interpretability of learned convolutional feature maps.

The demonstrated benefit of constrained tensor factorisation of HypER raised the question if imposing hard constraints on the core tensor weights is necessary or whether soft, regularisation constraints are preferable, which we investigate in the following chapter.

# Chapter 4

# TuckER: Tensor Factorisation for Knowledge Graph Completion

The main contribution of this chapter is the paper *TuckER: Tensor Factorization for Knowledge Graph Completion* which was published at the *Empirical Methods in Natural Language Processing* conference in November 2019. We first outline the motivation for this work (Section 4.1), followed by the paper itself (Section 4.2), the impact that it has had so far (Section 4.3), and discussion (Section 4.4).

## 4.1 Introduction

The paper in Section 4.2 was motivated by various existing multiplicative/tensor factorisation approaches to knowledge graph representation (e.g. Nickel et al., 2011; Yang et al., 2015; Trouillon et al., 2016) and the promising results achieved by HypER (Balažević et al., 2019) which can be recast as constrained tensor factorisation.

The following section introduces *TuckER*, a simple knowledge graph representation model based on the Tucker decomposition (Tucker, 1964) of the binary adjacency tensor of knowledge graph triples $\mathcal{G} \in \{0,1\}^{n_e \times n_r \times n_e}$, which outperforms all previous models on the link prediction task to achieve state-of-the-art results across standard datasets. TuckER is trained using gradient-based methods as explicit analytical factorisation cannot be computed, since the logistic sigmoid is applied to the score function of every triple to approximate $\mathcal{G}$ and thus the implicit underlying tensor $\mathcal{H} = \sigma^{-1}(\mathcal{G}) \in \{-\infty, \infty\}^{n_e \times n_r \times n_e}$ being factorised is comprised of $-\infty$ and $\infty$. The implicit tensor $\mathcal{H}$ is decomposed into three matrices — two equivalent ones for subject and object entity embeddings $\mathbf{E} \in \mathbb{R}^{n_e \times d_e}$, one for relation embeddings $\mathbf{R} \in \mathbb{R}^{n_r \times d_r}$ — and a much smaller core tensor $\mathcal{W} \in \mathbb{R}^{d_e \times d_r \times d_e}$.

We show that TuckER subsumes several previously proposed tensor factorisation approaches, i.e. RESCAL (Nickel et al., 2011), DistMult (Yang et al., 2015), ComplEx (Trouillon et al., 2016) and SimplE (Kazemi and Poole, 2018) can all be seen as special case of TuckER with hard constraints imposed on the core tensor $\mathcal{W}$. In contrast, TuckER does not constrain the parameters of $\mathcal{W}$ and allows them to be learned from the data. Thus, TuckER does not encode all the learned knowledge into the embeddings — some is stored in the entity- and relation-agnostic core tensor and shared between all entities and relations through *multi-task learning.* Rather than learning distinct relation-specific matrices, the core tensor $\mathcal{W}$ of TuckER can be viewed as containing a shared pool of "prototype" relation matrices, which are linearly combined according to the parameters in each relation embedding $\mathbf{w}_r \in \mathbf{R}^{d_r}$ (row of the $\mathbf{R}$ matrix), thus allowing knowledge sharing between entities and relations which may particularly benefit relations with not much training data available.

## 4.2   The Paper

### Author Contributions

The paper is co-authored by myself, Carl Allen and Timothy Hospedales. Timothy Hospedales seeded the idea of a multi-task approach to knowledge graph representation, supervised the project, offered suggestions, and helped revise the final version. As the lead author, I conceived the use of Tucker decomposition to achieve multi-task learning, developed TuckER, performed the experiments, and wrote the paper. Carl Allen participated in many discussions, helped relate TuckER to existing tensor factorisation approaches and proofread the paper.

# TuckER: Tensor Factorization for Knowledge Graph Completion

**Ivana Balažević**[1]        **Carl Allen**[1]        **Timothy M. Hospedales**[1,2]

[1] School of Informatics, University of Edinburgh, UK

[2] Samsung AI Centre, Cambridge, UK

{ivana.balazevic, carl.allen, t.hospedales}@ed.ac.uk

## Abstract

Knowledge graphs are structured representations of real world facts. However, they typically contain only a small subset of all possible facts. Link prediction is a task of inferring missing facts based on existing ones. We propose TuckER, a relatively straightforward but powerful linear model based on Tucker decomposition of the binary tensor representation of knowledge graph triples. TuckER outperforms previous state-of-the-art models across standard link prediction datasets, acting as a strong baseline for more elaborate models. We show that TuckER is a fully expressive model, derive sufficient bounds on its embedding dimensionalities and demonstrate that several previously introduced linear models can be viewed as special cases of TuckER.

## 1 Introduction

Vast amounts of information available in the world can be represented succinctly as *entities* and *relations* between them. *Knowledge graphs* are large, graph-structured databases which store facts in triple form $(e_s, r, e_o)$, with $e_s$ and $e_o$ representing subject and object entities and $r$ a relation. However, far from all available information is currently stored in existing knowledge graphs and manually adding new information is costly, which creates the need for algorithms that are able to automatically infer missing facts.

Knowledge graphs can be represented as a third-order binary tensor, where each element corresponds to a triple, 1 indicating a true fact and 0 indicating the unknown (either a false or a missing fact). The task of *link prediction* is to predict whether two entities are related, based on known facts already present in a knowledge graph, i.e. to infer which of the 0 entries in the tensor are indeed false, and which are missing but actually true.

A large number of approaches to link prediction so far have been linear, based on various methods of factorizing the third-order binary tensor (Nickel et al., 2011; Yang et al., 2015; Trouillon et al., 2016; Kazemi and Poole, 2018). Recently, state-of-the-art results have been achieved using non-linear convolutional models (Dettmers et al., 2018; Balažević et al., 2019). Despite achieving very good performance, the fundamental problem with deep, non-linear models is that they are non-transparent and poorly understood, as opposed to more mathematically principled and widely studied tensor decomposition models.

In this paper, we introduce TuckER (E stands for entities, R for relations), a straightforward linear model for link prediction on knowledge graphs, based on *Tucker decomposition* (Tucker, 1966) of the binary tensor of triples, acting as a strong baseline for more elaborate models. Tucker decomposition, used widely in machine learning (Schein et al., 2016; Ben-Younes et al., 2017; Yang and Hospedales, 2017), factorizes a tensor into a core tensor multiplied by a matrix along each mode. It can be thought of as a form of higher-order SVD in the special case where matrices are orthogonal and the core tensor is "all-orthogonal" (Kroonenberg and De Leeuw, 1980). In our case, rows of the matrices contain entity and relation embeddings, while entries of the core tensor determine the level of interaction between them. Subject and object entity embedding matrices are assumed equivalent, i.e. we make no distinction between the embeddings of an entity depending on whether it appears as a subject or as an object in a particular triple. Due to the low rank of the core tensor, TuckER benefits from *multi-task learning* by parameter sharing across relations.

A link prediction model should have enough expressive power to represent all relation types (e.g. symmetric, asymmetric, transitive). We thus show

5185

that TuckER is *fully expressive*, i.e. given any ground truth over the triples, there exists an assignment of values to the entity and relation embeddings that accurately separates the true triples from false ones. We also derive a dimensionality bound which guarantees full expressiveness.

Finally, we show that several previous state-of-the-art linear models, RESCAL (Nickel et al., 2011), DistMult (Yang et al., 2015), ComplEx (Trouillon et al., 2016) and SimplE (Kazemi and Poole, 2018), are special cases of TuckER.

In summary, key contributions of this paper are:

- proposing TuckER, a new *linear* model for link prediction on knowledge graphs, that is simple, expressive and achieves *state-of-the-art results* across all standard datasets;

- proving that TuckER is *fully expressive* and deriving a bound on the embedding dimensionality for full expressiveness; and

- showing how TuckER subsumes several previously proposed tensor factorization approaches to link prediction.

## 2   Related Work

Several *linear* models for link prediction have previously been proposed:

**RESCAL** (Nickel et al., 2011) optimizes a scoring function containing a bilinear product between subject and object entity vectors and a full rank relation matrix. Although a very expressive and powerful model, RESCAL is prone to overfitting due to its large number of parameters, which increases quadratically in the embedding dimension with the number of relations in a knowledge graph. **DistMult** (Yang et al., 2015) is a special case of RESCAL with a diagonal matrix per relation, which reduces overfitting. However, the linear transformation performed on entity embedding vectors in DistMult is limited to a stretch. The binary tensor learned by DistMult is symmetric in the subject and object entity mode and thus DistMult cannot model asymmetric relations. **ComplEx** (Trouillon et al., 2016) extends DistMult to the complex domain. Subject and object entity embeddings for the same entity are complex conjugates, which introduces asymmetry into the tensor decomposition and thus enables ComplEx to model asymmetric relations. **SimplE** (Kazemi and Poole, 2018) is based on Canonical Polyadic (CP) decomposition (Hitchcock, 1927), in which subject and object entity embeddings for the same entity are independent (note that DistMult is a special case of CP). SimplE's scoring function alters CP to make subject and object entity embedding vectors dependent on each other by computing the average of two terms, first of which is a bilinear product of the subject entity head embedding, relation embedding and object entity tail embedding and the second is a bilinear product of the object entity head embedding, inverse relation embedding and subject entity tail embedding.

Recently, state-of-the-art results have been achieved with *non-linear* models:

**ConvE** (Dettmers et al., 2018) performs a global 2D convolution operation on the subject entity and relation embedding vectors, after they are reshaped to matrices and concatenated. The obtained feature maps are flattened, transformed through a linear layer, and the inner product is taken with all object entity vectors to generate a score for each triple. Whilst results achieved by ConvE are impressive, its reshaping and concatenating of vectors as well as using 2D convolution on word embeddings is unintuitive.

**HypER** (Balažević et al., 2019) is a simplified convolutional model, that uses a hypernetwork to generate 1D convolutional filters for each relation, extracting relation-specific features from subject entity embeddings. The authors show that convolution is a way of introducing sparsity and parameter tying and that HypER can be understood in terms of tensor factorization up to a non-linearity, thus placing HypER closer to the well established family of factorization models. The drawback of HypER is that it sets most elements of the core weight tensor to 0, which amounts to hard regularization, rather than letting the model learn which parameters to use via soft regularization.

Scoring functions of all models described above and TuckER are summarized in Table 1.

## 3   Background

Let $\mathcal{E}$ denote the set of all entities and $\mathcal{R}$ the set of all relations present in a knowledge graph. A triple is represented as $(e_s, r, e_o)$, with $e_s, e_o \in \mathcal{E}$ denoting subject and object entities respectively and $r \in \mathcal{R}$ the relation between them.

### 3.1   Link Prediction

In link prediction, we are given a subset of all true triples and the aim is to learn a *scoring function* $\phi$

| Model | Scoring Function | Relation Parameters | Space Complexity |
|---|---|---|---|
| RESCAL (Nickel et al., 2011) | $\mathbf{e}_s^\top \mathbf{W}_r \mathbf{e}_o$ | $\mathbf{W}_r \in \mathbb{R}^{d_e^2}$ | $\mathcal{O}(n_e d_e + n_r d_r^2)$ |
| DistMult (Yang et al., 2015) | $\langle \mathbf{e}_s, \mathbf{w}_r, \mathbf{e}_o \rangle$ | $\mathbf{w}_r \in \mathbb{R}^{d_e}$ | $\mathcal{O}(n_e d_e + n_r d_e)$ |
| ComplEx (Trouillon et al., 2016) | $\mathrm{Re}(\langle \mathbf{e}_s, \mathbf{w}_r, \overline{\mathbf{e}}_o \rangle)$ | $\mathbf{w}_r \in \mathbb{C}^{d_e}$ | $\mathcal{O}(n_e d_e + n_r d_e)$ |
| ConvE (Dettmers et al., 2018) | $f(\mathrm{vec}(f([\underline{\mathbf{e}}_s; \underline{\mathbf{w}}_r] * w))\mathbf{W})\mathbf{e}_o$ | $\mathbf{w}_r \in \mathbb{R}^{d_r}$ | $\mathcal{O}(n_e d_e + n_r d_r)$ |
| SimplE (Kazemi and Poole, 2018) | $\frac{1}{2}(\langle \mathbf{h}_{e_s}, \mathbf{w}_r, \mathbf{t}_{e_o} \rangle + \langle \mathbf{h}_{e_o}, \mathbf{w}_{r^{-1}}, \mathbf{t}_{e_s} \rangle)$ | $\mathbf{w}_r \in \mathbb{R}^{d_e}$ | $\mathcal{O}(n_e d_e + n_r d_e)$ |
| HypER (Balažević et al., 2019) | $f(\mathrm{vec}(\mathbf{e}_s * \mathrm{vec}^{-1}(\mathbf{w}_r \mathbf{H}))\mathbf{W})\mathbf{e}_o$ | $\mathbf{w}_r \in \mathbb{R}^{d_r}$ | $\mathcal{O}(n_e d_e + n_r d_r)$ |
| TuckER (ours) | $\mathcal{W} \times_1 \mathbf{e}_s \times_2 \mathbf{w}_r \times_3 \mathbf{e}_o$ | $\mathbf{w}_r \in \mathbb{R}^{d_r}$ | $\mathcal{O}(n_e d_e + n_r d_r)$ |

Table 1: Scoring functions of state-of-the-art link prediction models, the dimensionality of their relation parameters, and significant terms of their space complexity. $d_e$ and $d_r$ are the dimensionalities of entity and relation embeddings, while $n_e$ and $n_r$ denote the number of entities and relations respectively. $\overline{\mathbf{e}}_o \in \mathbb{C}^{d_e}$ is the complex conjugate of $\mathbf{e}_o$, $\underline{\mathbf{e}}_s, \underline{\mathbf{w}}_r \in \mathbb{R}^{d_w \times d_h}$ denote a 2D reshaping of $\mathbf{e}_s$ and $\mathbf{w}_r$ respectively, $\mathbf{h}_{e_s}, \mathbf{t}_{e_s} \in \mathbb{R}^{d_e}$ are the head and tail entity embedding of entity $e_s$, and $\mathbf{w}_{r^{-1}} \in \mathbb{R}^{d_r}$ is the embedding of relation $r^{-1}$ (which is the inverse of relation $r$). $*$ is the convolution operator, $\langle \cdot \rangle$ denotes the dot product and $\times_n$ denotes the tensor product along the $n$-th mode, $f$ is a non-linear function, and $\mathcal{W} \in \mathbb{R}^{d_e \times d_e \times d_r}$ is the core tensor of a Tucker decomposition.

that assigns a score $s = \phi(e_s, r, e_o) \in \mathbb{R}$ which indicates whether a triple is true, with the ultimate goal of being able to correctly score all missing triples. The scoring function is either a specific form of tensor factorization in the case of linear models or a more complex (deep) neural network architecture for non-linear models. Typically, a positive score for a particular triple indicates a true fact predicted by the model, while a negative score indicates a false one. With most recent models, a non-linearity such as the logistic sigmoid function is typically applied to the score to give a corresponding probability prediction $p = \sigma(s) \in [0, 1]$ as to whether a certain fact is true.

### 3.2 Tucker Decomposition

Tucker decomposition, named after Ledyard R. Tucker (Tucker, 1964), decomposes a tensor into a set of matrices and a smaller core tensor. In a three-mode case, given the original tensor $\mathcal{X} \in \mathbb{R}^{I \times J \times K}$, Tucker decomposition outputs a tensor $\mathcal{Z} \in \mathbb{R}^{P \times Q \times R}$ and three matrices $\mathbf{A} \in \mathbb{R}^{I \times P}$, $\mathbf{B} \in \mathbb{R}^{J \times Q}$, $\mathbf{C} \in \mathbb{R}^{K \times R}$:

$$\mathcal{X} \approx \mathcal{Z} \times_1 \mathbf{A} \times_2 \mathbf{B} \times_3 \mathbf{C}, \quad (1)$$

with $\times_n$ indicating the tensor product along the n-th mode. *Factor matrices* $\mathbf{A}$, $\mathbf{B}$ and $\mathbf{C}$, when orthogonal, can be thought of as the principal components in each mode. Elements of the *core tensor* $\mathcal{Z}$ show the level of interaction between the different components. Typically, $P$, $Q$, $R$ are smaller than $I$, $J$, $K$ respectively, so $\mathcal{Z}$ can be thought of as a compressed version of $\mathcal{X}$. Tucker decomposition is not unique, i.e. we can transform $\mathcal{Z}$ without affecting the fit if we apply the inverse transformation to $\mathbf{A}$, $\mathbf{B}$ and $\mathbf{C}$ (Kolda and Bader, 2009).

## 4 Tucker Decomposition for Link Prediction

We propose a model that uses Tucker decomposition for link prediction on the binary tensor representation of a knowledge graph, with entity embedding matrix $\mathbf{E}$ that is equivalent for subject and object entities, i.e. $\mathbf{E} = \mathbf{A} = \mathbf{C} \in \mathbb{R}^{n_e \times d_e}$ and relation embedding matrix $\mathbf{R} = \mathbf{B} \in \mathbb{R}^{n_r \times d_r}$, where $n_e$ and $n_r$ represent the number of entities and relations and $d_e$ and $d_r$ the dimensionality of entity and relation embedding vectors.

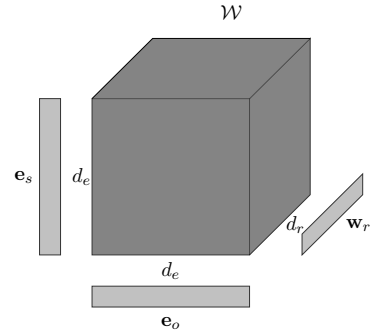

Figure 1: Visualization of the TuckER architecture.

We define the scoring function for TuckER as:

$$\phi(e_s, r, e_o) = \mathcal{W} \times_1 \mathbf{e}_s \times_2 \mathbf{w}_r \times_3 \mathbf{e}_o, \quad (2)$$

where $\mathbf{e}_s, \mathbf{e}_o \in \mathbb{R}^{d_e}$ are the rows of $\mathbf{E}$ representing the subject and object entity embedding vectors, $\mathbf{w}_r \in \mathbb{R}^{d_r}$ the rows of $\mathbf{R}$ representing the relation embedding vector and $\mathcal{W} \in \mathbb{R}^{d_e \times d_r \times d_e}$ is the core tensor. We apply logistic sigmoid to each score $\phi(e_s, r, e_o)$ to obtain the predicted probability $p$ of a triple being true. Visualization of the TuckER architecture can be seen in Figure 1. As proven in Section 5.1, TuckER is *fully expressive*. Further, its number of parameters increases *linearly* with respect to entity and relation embedding dimensionality $d_e$ and $d_r$, as the number of entities and relations increases, since the number of parameters of $\mathcal{W}$ depends only on the entity and relation embedding dimensionality and not on the number of entities or relations. By having the core tensor $\mathcal{W}$, unlike simpler models such as DistMult, ComplEx and SimplE, TuckER does not encode all the learned knowledge into the embeddings; some is stored in the core tensor and shared between all entities and relations through *multi-task learning*. Rather than learning distinct relation-specific matrices, the core tensor of TuckER can be viewed as containing a shared pool of "prototype" relation matrices, which are linearly combined according to the parameters in each relation embedding.

## 4.1   Training

Since the logistic sigmoid is applied to the scoring function to approximate the true binary tensor, the implicit underlying tensor is comprised of $-\infty$ and $\infty$. Given this prevents an explicit analytical factorization, we use numerical methods to train TuckER. We use the standard data augmentation technique, first used by Dettmers et al. (2018) and formally described by Lacroix et al. (2018), of adding reciprocal relations for every triple in the dataset, i.e. we add $(e_o, r^{-1}, e_s)$ for every $(e_s, r, e_o)$. Following the training procedure introduced by Dettmers et al. (2018) to speed up training, we use *1-N scoring*, i.e. we simultaneously score entity-relation pairs $(e_s, r)$ and $(e_o, r^{-1})$ with all entities $e_o \in \mathcal{E}$ and $e_s \in \mathcal{E}$ respectively, in contrast to *1-1 scoring*, where individual triples $(e_s, r, e_o)$ and $(e_o, r^{-1}, e_s)$ are trained one at a time. The model is trained to minimize the Bernoulli negative log-likelihood loss function. A component of the loss for one entity-relation pair with all others entities is defined as:

$$L = -\frac{1}{n_e} \sum_{i=1}^{n_e} (\mathbf{y}^{(i)} \log(\mathbf{p}^{(i)}) + (1 - \mathbf{y}^{(i)}) \log(1 - \mathbf{p}^{(i)})), \quad (3)$$

where $\mathbf{p} \in \mathbb{R}^{n_e}$ is the vector of predicted probabilities and $\mathbf{y} \in \mathbb{R}^{n_e}$ is the binary label vector.

## 5   Theoretical Analysis

### 5.1   Full Expressiveness and Embedding Dimensionality

A tensor factorization model is fully expressive if for any ground truth over all entities and relations, there exist entity and relation embeddings that accurately separate true triples from the false. As shown in (Trouillon et al., 2017), ComplEx is fully expressive with the embedding dimensionality bound $d_e = d_r = n_e \cdot n_r$. Similarly to ComplEx, Kazemi and Poole (2018) show that SimplE is fully expressive with entity and relation embeddings of size $d_e = d_r = min(n_e \cdot n_r, \gamma + 1)$, where $\gamma$ represents the number of true facts. They further prove other models are not fully expressive: DistMult, because it cannot model asymmetric relations; and transitive models such as TransE (Bordes et al., 2013) and its variants FTransE (Feng et al., 2016) and STransE (Nguyen et al., 2016), because of certain contradictions that they impose between different relation types. By Theorem 1, we establish the bound on entity and relation embedding dimensionality (i.e. decomposition rank) that guarantees full expressiveness of TuckER.

**Theorem 1.** *Given any ground truth over a set of entities $\mathcal{E}$ and relations $\mathcal{R}$, there exists a TuckER model with entity embeddings of dimensionality $d_e = n_e$ and relation embeddings of dimensionality $d_r = n_r$, where $n_e = |\mathcal{E}|$ is the number of entities and $n_r = |\mathcal{R}|$ the number of relations, that accurately represents that ground truth.*

*Proof.* Let $\mathbf{e}_s$ and $\mathbf{e}_o$ be the $n_e$-dimensional one-hot binary vector representations of subject and object entities $e_s$ and $e_o$ respectively and $\mathbf{w}_r$ the $n_r$-dimensional one-hot binary vector representation of relation $r$. For each subject entity $e_s^{(i)}$, relation $r^{(j)}$ and object entity $e_o^{(k)}$, we let the $i$-th, $j$-th and $k$-th element respectively of the corresponding vectors $\mathbf{e}_s$, $\mathbf{w}_r$ and $\mathbf{e}_o$ be 1 and all other elements 0. Further, we set the $ijk$ element of the tensor $\mathcal{W} \in \mathbb{R}^{n_e \times n_r \times n_e}$ to 1 if the fact $(e_s, r, e_o)$ holds and -1 otherwise. Thus the product of the entity embeddings and the relation embedding with the core tensor, after applying the logistic sigmoid, accurately represents the original tensor. $\quad\square$

The purpose of Theorem 1 is to prove that TuckER is capable of potentially capturing all information (and noise) in the data. In practice however, we expect the embedding dimensionalities needed for full reconstruction of the underlying binary tensor to be much smaller than the bound stated above, since the assignment of values to the tensor is not random but follows a certain structure, otherwise nothing unknown could be predicted. Even more so, low decomposition rank is actually a desired property of any bilin-ear link prediction model, forcing it to learn that structure and generalize to new data, rather than simply memorizing the input. In general, we expect TuckER to perform better than ComplEx and SimplE with embeddings of lower dimensionality due to parameter sharing in the core tensor (shown empirically in Section 6.4), which could be of importance for efficiency in downstream tasks.

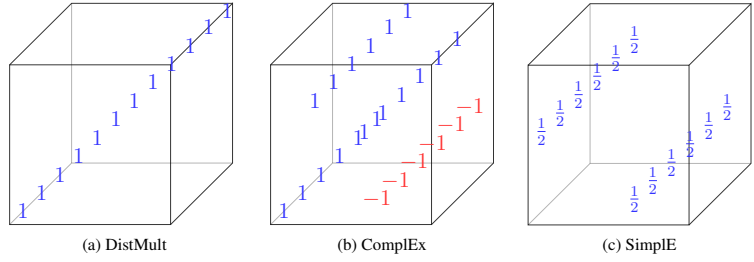

Figure 2: Constraints imposed on the values of core tensor $\mathcal{Z} \in \mathbb{R}^{d_e \times d_e \times d_e}$ for DistMult and $\mathcal{Z} \in \mathbb{R}^{2d_e \times 2d_e \times 2d_e}$ for ComplEx and SimplE. Elements that are set to 0 are represented in white.

## 5.2   Relation to Previous Linear Models

Several previous tensor factorization models can be viewed as a special case of TuckER:

**RESCAL (Nickel et al., 2011)**   Following the notation introduced in Section 3.2, the RESCAL scoring function (see Table 1) has the form:

$$\mathcal{X} \approx \mathcal{Z} \times_1 \mathbf{A} \times_3 \mathbf{C}. \tag{4}$$

This corresponds to Equation 1 with $I = K = n_e$, $P = R = d_e$, $Q = J = n_r$ and $\mathbf{B} = \mathbf{I}_J$ the $J \times J$ identity matrix. This is also known as Tucker2 decomposition (Kolda and Bader, 2009). As is the case with TuckER, the entity embedding matrix of RESCAL is shared between subject and object entities, i.e. $\mathbf{E} = \mathbf{A} = \mathbf{C} \in \mathbb{R}^{n_e \times d_e}$ and the relation matrices $\mathbf{W}_r \in \mathbb{R}^{d_e \times d_e}$ are the $d_e \times d_e$ slices of the core tensor $\mathcal{Z}$. As mentioned in Section 2, the drawback of RESCAL compared to TuckER is that its number of parameters grows *quadratically* in the entity embedding dimension $d_e$ as the number of relations increases.

**DistMult (Yang et al., 2015)**   The scoring function of DistMult (see Table 1) can be viewed as equivalent to that of TuckER (see Equation 1) with a core tensor $\mathcal{Z} \in \mathbb{R}^{P \times Q \times R}$, $P = Q = R = d_e$, which is *superdiagonal* with 1s on the superdiagonal, i.e. all elements $z_{pqr}$ with $p = q = r$ are 1 and all the other elements are 0 (as shown in Figure 2a). Rows of $\mathbf{E} = \mathbf{A} = \mathbf{C} \in \mathbb{R}^{n_e \times d_e}$ contain subject and object entity embedding vectors $\mathbf{e}_s, \mathbf{e}_o \in \mathbb{R}^{d_e}$ and rows of $\mathbf{R} = \mathbf{B} \in \mathbb{R}^{n_r \times d_e}$ contain relation embedding vectors $\mathbf{w}_r \in \mathbb{R}^{d_e}$. It is interesting to note that the TuckER interpretation of the DistMult scoring function, given that matrices $\mathbf{A}$ and $\mathbf{C}$ are identical, can alternatively be interpreted as a special case of CP decomposition (Hitchcock, 1927), since Tucker decomposition with a superdiagonal core tensor is equivalent to CP decomposition. Due to enforced symmetry in subject and object entity mode, DistMult cannot learn to represent asymmetric relations.

**ComplEx (Trouillon et al., 2016)**   *Bilinear models* represent subject and object entity embeddings as vectors $\mathbf{e}_s, \mathbf{e}_o \in \mathbb{R}^{d_e}$, relation as a matrix $\mathbf{W}_r \in \mathbb{R}^{d_e \times d_e}$ and the scoring function as a bilinear product $\phi(e_s, r, e_o) = \mathbf{e}_s \mathbf{W}_r \mathbf{e}_o$. It is trivial to show that both RESCAL and DistMult belong to the family of bilinear models. As explained by Kazemi and Poole (2018), ComplEx can be considered a bilinear model with the real and imaginary part of an embedding for each entity concatenated in a single vector, $[\text{Re}(\mathbf{e}_s); \text{Im}(\mathbf{e}_s)] \in \mathbb{R}^{2d_e}$ for subject, $[\text{Re}(\mathbf{e}_o); \text{Im}(\mathbf{e}_o)] \in \mathbb{R}^{2d_e}$ for object, and a relation matrix $\mathbf{W}_r \in \mathbb{R}^{2d_e \times 2d_e}$, constrained so that its leading diagonal contains duplicated elements of $\text{Re}(\mathbf{w}_r)$, its $d_e$-diagonal elements of $\text{Im}(\mathbf{w}_r)$ and its $-d_e$-diagonal elements of $-\text{Im}(\mathbf{w}_r)$, with all other elements set to 0, where $d_e$ and $-d_e$ represent offsets from the leading diagonal.

Similarly to DistMult, we can regard the scoring function of ComplEx (see Table 1) as equivalent to the scoring function of TuckER (see Equation 1), with core tensor $\mathcal{Z} \in \mathbb{R}^{P \times Q \times R}$, $P = Q = R = 2d_e$, where $3d_e$ elements on different tensor diagonals are set to 1, $d_e$ elements on one tensor diagonal are set to -1 and all other elements are set to 0 (see Figure 2b). This shows that the scoring function of ComplEx, which computes a bilinear product with *complex* entity and relation embeddings and disregards the imaginary part of the obtained result, is equivalent to a hard regularization of the core tensor of TuckER in the *real* domain.

**SimplE (Kazemi and Poole, 2018)**    The authors show that SimplE belongs to the family of bilinear models by concatenating embeddings for head and tail entities for both subject and object into vectors $[\mathbf{h}_{e_s}; \mathbf{t}_{e_s}] \in \mathbb{R}^{2d_e}$ and $[\mathbf{h}_{e_o}; \mathbf{t}_{e_o}] \in \mathbb{R}^{2d_e}$ and constraining the relation matrix $\mathbf{W}_r \in \mathbb{R}^{2d_e \times 2d_e}$ so that it contains the relation embedding vector $\frac{1}{2}\mathbf{w}_r$ on its $d_e$-diagonal and the inverse relation embedding vector $\frac{1}{2}\mathbf{w}_{r^{-1}}$ on its $-d_e$-diagonal and 0s elsewhere. The SimplE scoring function (see Table 1) is therefore equivalent to that of TuckER (see Equation 1), with core tensor $\mathcal{Z} \in \mathbb{R}^{P \times Q \times R}$, $P = Q = R = 2d_e$, where $2d_e$ elements on two tensor diagonals are set to $\frac{1}{2}$ and all other elements are set to 0 (see Figure 2c).

## 5.3   Representing Asymmetric Relations

Each relation in a knowledge graph can be characterized by a certain set of properties, such as symmetry, reflexivity, transitivity. So far, there have been two possible ways in which linear link prediction models introduce *asymmetry* into factorization of the binary tensor of triples:

- distinct (although possibly related) embeddings for subject and object entities and a diagonal matrix (or equivalently a vector) for each relation, as is the case with models such as ComplEx and SimplE; or

- equivalent subject and object entity embeddings and each relation represented by a full rank matrix, which is the case with RESCAL.

The latter approach appears more intuitive, since asymmetry is a property of the relation, rather than the entities. However, the drawback of the latter approach is quadratic growth of parameter number with the number of relations, which often leads to overfitting, especially for relations with a small number of training triples. TuckER overcomes this by representing relations as vectors $\mathbf{w}_r$, which makes the parameter number grow linearly with the number of relations, while still keeping the desirable property of allowing relations to be asymmetric by having an asymmetric *relation-agnostic* core tensor $\mathcal{W}$, rather than encoding the relation-specific information in the entity embeddings. Multiplying $\mathcal{W} \in \mathbb{R}^{d_e \times d_r \times d_e}$ with $\mathbf{w}_r \in \mathbb{R}^{d_r}$ along the second mode, we obtain a full rank relation-specific matrix $\mathbf{W}_r \in \mathbb{R}^{d_e \times d_e}$, which can perform all possible linear transformations on the entity embeddings, i.e. rotation, reflection or stretch, and is thus also capable of modeling asymmetry. Regardless of what kind of transformation is needed for modeling a particular relation, TuckER can learn it from the data. To demonstrate this, we show sample heatmaps of learned relation matrices $\mathbf{W}_r$ for a WordNet symmetric relation "derivationally_related_form" and an asymmetric relation "hypernym" in Figure 3, where one can see that TuckER learns to model the symmetric relation with the relation matrix that is approximately symmetric about the main diagonal, whereas the matrix belonging to the asymmetric relation exhibits no obvious structure.

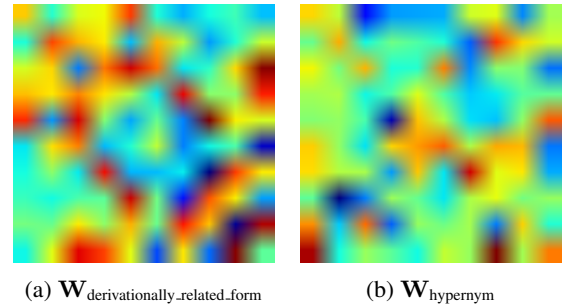

(a) $\mathbf{W}_{\text{derivationally\_related\_form}}$    (b) $\mathbf{W}_{\text{hypernym}}$

Figure 3: Learned relation matrices for a symmetric (derivationally_related_form) and an asymmetric (hypernym) WN18RR relation. $\mathbf{W}_{\text{derivationally\_related\_form}}$ is approximately symmetric about the leading diagonal.

## 6   Experiments and Results

### 6.1   Datasets

We evaluate TuckER using four standard link prediction datasets (see Table 2):

**FB15k (Bordes et al., 2013)** is a subset of Freebase, a large database of real world facts.

**FB15k-237 (Toutanova et al., 2015)** was created from FB15k by removing the inverse of many relations that are present in the training set from validation and test sets, making it more difficult for simple models to do well.

**WN18 (Bordes et al., 2013)** is a subset of WordNet, a hierarchical database containing lexical relations between words.

**WN18RR (Dettmers et al., 2018)** is a subset of WN18, created by removing the inverse relations from validation and test sets.

### 6.2   Implementation and Experiments

We implement TuckER in PyTorch (Paszke et al., 2017) and make our code available on GitHub.[1]

We choose all hyper-parameters by random search based on validation set performance. For

---

FB15k and FB15k-237, we set entity and relation embedding dimensionality to $d_e = d_r = 200$. For WN18 and WN18RR, which both contain a significantly smaller number of relations relative to the number of entities as well as a small number of relations compared to FB15k and FB15k-237, we set $d_e = 200$ and $d_r = 30$. We use batch normalization (Ioffe and Szegedy, 2015) and dropout (Srivastava et al., 2014) to speed up training. We find that lower dropout values $(0.1, 0.2)$ are required for datasets with a higher number of training triples per relation and thus less risk of overfitting (WN18 and WN18RR), whereas higher dropout values $(0.3, 0.4, 0.5)$ are required for FB15k and FB15k-237. We choose the learning rate from $\{0.01, 0.005, 0.003, 0.001, 0.0005\}$ and learning rate decay from $\{1, 0.995, 0.99\}$. We find the following combinations of learning rate and learning rate decay to give the best results: $(0.003, 0.99)$ for FB15k, $(0.0005, 1.0)$ for FB15k-237, $(0.005, 0.995)$ for WN18 and $(0.01, 1.0)$ for WN18RR (see Table 5 in the Appendix A for a complete list of hyper-parameter values on each dataset). We train the model using Adam (Kingma and Ba, 2015) with the batch size 128.

At evaluation time, for each test triple we generate $n_e$ candidate triples by combining the test entity-relation pair with all possible entities $\mathcal{E}$, ranking the scores obtained. We use the *filtered* setting (Bordes et al., 2013), i.e. all known true triples are removed from the candidate set except for the current test triple. We use evaluation metrics standard across the link prediction literature: mean reciprocal rank (MRR) and hits@$k$, $k \in \{1, 3, 10\}$. Mean reciprocal rank is the average of the inverse of the mean rank assigned to the true triple over all candidate triples. Hits@$k$ measures the percentage of times a true triple is ranked within the top $k$ candidate triples.

| Dataset | # Entities ($n_e$) | # Relations ($n_r$) |
|---|---|---|
| FB15k | 14,951 | 1,345 |
| FB15k-237 | 14,541 | 237 |
| WN18 | 40,943 | 18 |
| WN18RR | 40,943 | 11 |

Table 2: Dataset statistics.

## 6.3 Link Prediction Results

Link prediction results on all datasets are shown in Tables 3 and 4. Overall, TuckER outperforms previous state-of-the-art models on all metrics across all datasets (apart from hits@10 on WN18 where a non-linear model, R-GCN, does better). Results achieved by TuckER are not only better than those of other linear models, such as DistMult, ComplEx and SimplE, but also better than the results of many more complex deep neural network and reinforcement learning architectures, e.g. R-GCN, MINERVA, ConvE and HypER, demonstrating the expressive power of linear models and supporting our claim that simple linear models should serve as a baseline before moving onto more elaborate models.

Even with fewer parameters than ComplEx and SimplE at $d_e = 200$ and $d_r = 30$ on WN18RR ($\sim$9.4 vs $\sim$16.4 million), TuckER consistently obtains better results than any of those models. We believe this is because TuckER exploits knowledge sharing between relations through the core tensor, i.e. multi-task learning. This is supported by the fact that the margin by which TuckER outperforms other linear models is notably increased on datasets with a large number of relations. For example, improvement on FB15k is $+14\%$ over ComplEx and $+8\%$ over SimplE on the toughest hits@1 metric. To our knowledge, ComplEx-N3 (Lacroix et al., 2018) is the only other linear link prediction model that benefits from multi-task learning. There, rank regularization of the embedding matrices is used to encourage a low-rank factorization, thus forcing parameter sharing between relations. We do not include their published results in Tables 3 and 4, since they use the highly non-standard $d_e = d_r = 2000$ and thus a far larger parameter number (18x more parameters than TuckER on WN18RR; 5.5x on FB15k-237), making their results incomparable to those typically reported, including our own. However, running their model with equivalent parameter number to TuckER shows comparable performance, supporting our belief that the two models both attain the benefits of multi-task learning, although by different means.

## 6.4 Influence of Parameter Sharing

The ability of knowledge sharing through the core tensor suggests that TuckER should need a lower number of parameters for obtaining good results than ComplEx or SimplE. To test this, we re-implement ComplEx and SimplE with reciprocal relations, 1-N scoring, batch normalization and dropout for fair comparison, perform random search to choose best hyper-parameters

| | | WN18RR | | | | FB15k-237 | | | |
|---|---|---|---|---|---|---|---|---|---|
| | Linear | MRR | Hits@10 | Hits@3 | Hits@1 | MRR | Hits@10 | Hits@3 | Hits@1 |
| DistMult (Yang et al., 2015) | yes | .430 | .490 | .440 | .390 | .241 | .419 | .263 | .155 |
| ComplEx (Trouillon et al., 2016) | yes | .440 | .510 | .460 | .410 | .247 | .428 | .275 | .158 |
| Neural LP (Yang et al., 2017) | no | – | – | – | – | .250 | .408 | – | – |
| R-GCN (Schlichtkrull et al., 2018) | no | – | – | – | – | .248 | .417 | .264 | .151 |
| MINERVA (Das et al., 2018) | no | – | – | – | – | – | .456 | – | – |
| ConvE (Dettmers et al., 2018) | no | .430 | .520 | .440 | .400 | .325 | .501 | .356 | .237 |
| HypER (Balažević et al., 2019) | no | .465 | .522 | .477 | .436 | .341 | .520 | .376 | .252 |
| M-Walk (Shen et al., 2018) | no | .437 | – | .445 | .414 | – | – | – | – |
| RotatE (Sun et al., 2019) | no | – | – | – | – | .297 | .480 | .328 | .205 |
| TuckER (ours) | yes | **.470** | **.526** | **.482** | **.443** | **.358** | **.544** | **.394** | **.266** |

Table 3: Link prediction results on WN18RR and FB15k-237. The RotatE (Sun et al., 2019) results are reported without their self-adversarial negative sampling (see Appendix H in the original paper) for fair comparison.

| | | WN18 | | | | FB15k | | | |
|---|---|---|---|---|---|---|---|---|---|
| | Linear | MRR | Hits@10 | Hits@3 | Hits@1 | MRR | Hits@10 | Hits@3 | Hits@1 |
| TransE (Bordes et al., 2013) | no | – | .892 | – | – | – | .471 | – | – |
| DistMult (Yang et al., 2015) | yes | .822 | .936 | .914 | .728 | .654 | .824 | .733 | .546 |
| ComplEx (Trouillon et al., 2016) | yes | .941 | .947 | .936 | .936 | .692 | .840 | .759 | .599 |
| ANALOGY (Liu et al., 2017) | yes | .942 | .947 | .944 | .939 | .725 | .854 | .785 | .646 |
| Neural LP (Yang et al., 2017) | no | .940 | .945 | – | – | .760 | .837 | – | – |
| R-GCN (Schlichtkrull et al., 2018) | no | .819 | **.964** | .929 | .697 | .696 | .842 | .760 | .601 |
| TorusE (Ebisu and Ichise, 2018) | no | .947 | .954 | .950 | .943 | .733 | .832 | .771 | .674 |
| ConvE (Dettmers et al., 2018) | no | .943 | .956 | .946 | .935 | .657 | .831 | .723 | .558 |
| HypER (Balažević et al., 2019) | no | .951 | 958 | **.955** | .947 | .790 | .885 | .829 | .734 |
| SimplE (Kazemi and Poole, 2018) | yes | .942 | .947 | .944 | .939 | .727 | .838 | .773 | .660 |
| TuckER (ours) | yes | **.953** | .958 | **.955** | **.949** | **.795** | **.892** | **.833** | **.741** |

Table 4: Link prediction results on WN18 and FB15k.

(see Table 6 in the Appendix A for exact hyper-parameter values used) and train all three models on FB15k-237 with embedding sizes $d_e = d_r \in \{20, 50, 100, 200\}$. Figure 4 shows the obtained MRR on the test set for each model. It is important to note that at embedding dimensionalities 20, 50 and 100, TuckER has fewer parameters than ComplEx and SimplE (e.g. ComplEx and SimplE have $\sim$3 million and TuckER has $\sim$2.5 million parameters for embedding dimensionality 100).

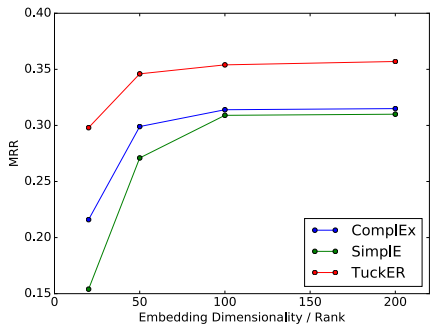

Figure 4: MRR for ComplEx, SimplE and TuckER for different embeddings sizes on FB15k-237.

We can see that the difference between the MRRs of ComplEx, SimplE and TuckER is approximately constant for embedding sizes 100 and 200. However, for lower embedding sizes, the dif-ference between MRRs increases by 0.7% for embedding size 50 and by 4.2% for embedding size 20 for ComplEx and by 3% for embedding size 50 and by 9.9% for embedding size 20 for SimplE. At embedding size 20 ($\sim$300k parameters), the performance of TuckER is almost as good as the performance of ComplEx and SimplE at embedding size 200 ($\sim$6 million parameters), which supports our initial assumption.

## 7   Conclusion

In this work, we introduce TuckER, a relatively straightforward linear model for link prediction on knowledge graphs, based on the Tucker decomposition of a binary tensor of known facts. TuckER achieves state-of-the-art results on standard link prediction datasets, in part due to its ability to perform multi-task learning across relations. Whilst being fully expressive, TuckER's number of parameters grows linearly with respect to the number of entities or relations in the knowledge graph. We further show that previous linear state-of-the-art models, RESCAL, DistMult, ComplEx and SimplE, can be interpreted as special cases of our model. Future work might include exploring how to incorporate background knowledge on individual relation properties into the existing model.

**Acknowledgements**

Ivana Balažević and Carl Allen were supported by the Centre for Doctoral Training in Data Science, funded by EPSRC (grant EP/L016427/1) and the University of Edinburgh.

# References

Ivana Balažević, Carl Allen, and Timothy M Hospedales. 2019. Hypernetwork Knowledge Graph Embeddings. In *International Conference on Artificial Neural Networks*.

Hedi Ben-Younes, Rémi Cadene, Matthieu Cord, and Nicolas Thome. 2017. MUTAN: Multimodal Tucker Fusion for Visual Question Answering. In *International Conference on Computer Vision*.

Antoine Bordes, Nicolas Usunier, Alberto Garcia-Duran, Jason Weston, and Oksana Yakhnenko. 2013. Translating Embeddings for Modeling Multi-relational Data. In *Advances in Neural Information Processing Systems*.

Rajarshi Das, Shehzaad Dhuliawala, Manzil Zaheer, Luke Vilnis, Ishan Durugkar, Akshay Krishnamurthy, Alex Smola, and Andrew McCallum. 2018. Go for a Walk and Arrive at the Answer: Reasoning over Paths in Knowledge Bases Using Reinforcement Learning. In *International Conference on Learning Representations*.

Tim Dettmers, Pasquale Minervini, Pontus Stenetorp, and Sebastian Riedel. 2018. Convolutional 2D Knowledge Graph Embeddings. In *Association for the Advancement of Artificial Intelligence*.

Takuma Ebisu and Ryutaro Ichise. 2018. TorusE: Knowledge Graph Embedding on a Lie Group. In *Association for the Advancement of Artificial Intelligence*.

Jun Feng, Minlie Huang, Mingdong Wang, Mantong Zhou, Yu Hao, and Xiaoyan Zhu. 2016. Knowledge Graph Embedding by Flexible Translation. In *Principles of Knowledge Representation and Reasoning*.

Frank L Hitchcock. 1927. The Expression of a Tensor or a Polyadic as a Sum of Products. *Journal of Mathematics and Physics*, 6(1-4):164–189.

Sergey Ioffe and Christian Szegedy. 2015. Batch Normalization: Accelerating Deep Network Training by Reducing Internal Covariate Shift. In *International Conference on Machine Learning*.

Seyed Mehran Kazemi and David Poole. 2018. SimplE Embedding for Link Prediction in Knowledge Graphs. In *Advances in Neural Information Processing Systems*.

Diederik P Kingma and Jimmy Ba. 2015. Adam: A Method for Stochastic Optimization. In *International Conference on Learning Representations*.

Tamara G Kolda and Brett W Bader. 2009. Tensor Decompositions and Applications. *SIAM review*, 51(3):455–500.

Pieter M Kroonenberg and Jan De Leeuw. 1980. Principal Component Analysis of Three-Mode Data by Means of Alternating Least Squares Algorithms. *Psychometrika*, 45(1):69–97.

Timothée Lacroix, Nicolas Usunier, and Guillaume Obozinski. 2018. Canonical Tensor Decomposition for Knowledge Base Completion. In *International Conference on Machine Learning*.

Hanxiao Liu, Yuexin Wu, and Yiming Yang. 2017. Analogical Inference for Multi-relational Embeddings. In *International Conference on Machine Learning*.

Dat Quoc Nguyen, Kairit Sirts, Lizhen Qu, and Mark Johnson. 2016. STransE: a Novel Embedding Model of Entities and Relationships in Knowledge Bases. In *North American Chapter of the Association for Computational Linguistics: Human Language Technologies*.

Maximilian Nickel, Volker Tresp, and Hans-Peter Kriegel. 2011. A Three-Way Model for Collective Learning on Multi-Relational Data. In *International Conference on Machine Learning*.

Adam Paszke, Sam Gross, Soumith Chintala, Gregory Chanan, Edward Yang, Zachary DeVito, Zeming Lin, Alban Desmaison, Luca Antiga, and Adam Lerer. 2017. Automatic Differentiation in PyTorch. In *NIPS-W*.

Aaron Schein, Mingyuan Zhou, David Blei, and Hanna Wallach. 2016. Bayesian Poisson Tucker Decomposition for Learning the Structure of International Relations. In *International Conference on Machine Learning*.

Michael Schlichtkrull, Thomas N Kipf, Peter Bloem, Rianne van den Berg, Ivan Titov, and Max Welling. 2018. Modeling Relational Data with Graph Convolutional Networks. In *European Semantic Web Conference*.

Yelong Shen, Jianshu Chen, Po-Sen Huang, Yuqing Guo, and Jianfeng Gao. 2018. M-Walk: Learning to Walk over Graphs using Monte Carlo Tree Search. In *Advances in Neural Information Processing Systems*.

Nitish Srivastava, Geoffrey Hinton, Alex Krizhevsky, Ilya Sutskever, and Ruslan Salakhutdinov. 2014. Dropout: A Simple Way to Prevent Neural Networks from Overfitting. *Journal of Machine Learning Research*, 15(1):1929–1958.

Zhiqing Sun, Zhi-Hong Deng, Jian-Yun Nie, and Jian Tang. 2019. RotatE: Knowledge Graph Embedding by Relational Rotation in Complex Space. In *International Conference on Learning Representations*.

Kristina Toutanova, Danqi Chen, Patrick Pantel, Hoifung Poon, Pallavi Choudhury, and Michael Gamon. 2015. Representing Text for Joint Embedding of Text and Knowledge Bases. In *Empirical Methods in Natural Language Processing*.

Théo Trouillon, Christopher R Dance, Éric Gaussier, Johannes Welbl, Sebastian Riedel, and Guillaume Bouchard. 2017. Knowledge Graph Completion via Complex Tensor Factorization. *Journal of Machine Learning Research*, 18(1):4735–4772.

Théo Trouillon, Johannes Welbl, Sebastian Riedel, Éric Gaussier, and Guillaume Bouchard. 2016. Complex Embeddings for Simple Link Prediction. In *International Conference on Machine Learning*.

Ledyard R Tucker. 1964. The Extension of Factor Analysis to Three-Dimensional Matrices. *Contributions to Mathematical Psychology*, 110119.

Ledyard R Tucker. 1966. Some Mathematical Notes on Three-Mode Factor Analysis. *Psychometrika*, 31(3):279–311.

Bishan Yang, Wen-tau Yih, Xiaodong He, Jianfeng Gao, and Li Deng. 2015. Embedding Entities and Relations for Learning and Inference in Knowledge Bases. In *International Conference on Learning Representations*.

Fan Yang, Zhilin Yang, and William W Cohen. 2017. Differentiable Learning of Logical Rules for Knowledge Base Reasoning. In *Advances in Neural Information Processing Systems*.

Yongxin Yang and Timothy Hospedales. 2017. Deep Multi-task Representation Learning: A Tensor Factorisation Approach. In *International Conference on Learning Representations*.

## A Hyper-parameters

Table 5 shows best performing hyper-parameter values for TuckER across all datasets, where lr denotes learning rate, dr decay rate, ls label smoothing, and d#$k$, $k \in \{1, 2, 3\}$ dropout values applied on the subject entity embedding, relation matrix and subject entity embedding after it has been transformed by the relation matrix respectively.

| Dataset | lr | dr | $d_e$ | $d_r$ | d#1 | d#2 | d#3 | ls |
|---------|------|------|-----|-----|-----|-----|-----|-----|
| FB15k | 0.003 | 0.99 | 200 | 200 | 0.2 | 0.2 | 0.3 | 0. |
| FB15k-237 | 0.0005 | 1.0 | 200 | 200 | 0.3 | 0.4 | 0.5 | 0.1 |
| WN18 | 0.005 | 0.995 | 200 | 30 | 0.2 | 0.1 | 0.2 | 0.1 |
| WN18RR | 0.01 | 1.0 | 200 | 30 | 0.2 | 0.2 | 0.3 | 0.1 |

Table 5: Best performing hyper-parameter values for TuckER across all datasets.

Table 6 shows best performing hyper-parameter values for ComplEx and SimplE on FB15k-237, used to produce the result in Figure 4.

| Model | lr | dr | $d_e$ | $d_r$ | d#1 | d#2 | d#3 | ls |
|-------|------|------|-----|-----|-----|-----|-----|-----|
| ComplEx | 0.0001 | 0.99 | 200 | 200 | 0.2 | 0. | 0. | 0.1 |
| SimplE | 0.0001 | 0.995 | 200 | 200 | 0.2 | 0. | 0. | 0.1 |

Table 6: Best performing hyper-parameter values for ComplEx and SimplE on FB15k-237.

## 4.3   Impact

According to Google Scholar, the paper has received 234 citations as of November 2021 and it was featured in numerous surveys (e.g. Ji et al., 2021; Nguyen, 2020; Hogan et al., 2020; Kazemi et al., 2020).

TuckER has directly influenced several followup works. Ma et al. (2019) generalise TuckER to temporal knowledge graphs. Liu et al. (2020) extend TuckER to tackle link prediction in n-ary relational knowledge graphs by introducing n-TuckER. Since the higher-order core tensor in n-TuckER grows exponentially with the relation arity, they decompose the core tensor into a sequence of 3D tensors using tensor ring decomposition. Amin et al. (2020) introduce LowFER, a knowledge graph representation model which decomposes the full-rank 3D core tensor $\mathcal{W}$ of TuckER into a sum of $k$ low-rank 3D tensors, thus reducing the overall parameter number of TuckER.

TuckER embeddings have also been used to incorporate external knowledge into existing systems. Peters et al. (2019) integrate TuckER entity embeddings with word embeddings to enhance BERT with external knowledge. Sarkar et al. (2020) use TuckER entity embeddings to improve conversational recommender systems.

Several recent works perform extensive experimental studies comparing popular knowledge graph representation model architectures and training strategies across a wide range of hyperparameter settings. Both Ruffinelli et al. (2019) and Sun et al. (2020) corroborate the state-of-the-art results achieved by TuckER, while finding evaluation issues with some of the more recent deep neural network architectures.

## 4.4   Discussion

Despite its simplicity, TuckER outperformed much more complex deep neural network and reinforcement learning models (e.g. R-GCN (Schlichtkrull et al., 2018), MINERVA (Das et al., 2018), ConvE (Dettmers et al., 2018), HypER (Balažević et al., 2019)) on the link prediction task. Thus, the main takeaway of the paper is that simple models should be explored before moving onto more elaborate ones. Further, both TuckER (by knowledge sharing in the core tensor) and ComplEx-N3 (Lacroix et al., 2018) (through rank regularisation of the embedding matrices which encourages a low rank factorisation and parameter sharing between relations) demonstrate the benefit of multi-task learning for link prediction. Investigating which relations are helped most by multi-task learning and why would be an interesting direction for future work.

One potential limitation of TuckER is the exponential growth in parameter number of the core tensor $\mathcal{W}$ with the embedding dimension and arity. This results in a cubic growth of parameter number of TuckER if the embedding dimension is increased, as well as prevents a direct extension of TuckER to temporal or n-ary relational knowledge graphs. However, Liu et al. (2020) and Amin et al. (2020) alleviate this issue by proposing different low-rank factorisations of the core tensor.

Another downside of TuckER is that it is a purely multiplicative model, whereas in Chapter 6 we argue that both multiplicative and additive relation-specific transformations are necessary for modelling all semantic relation types. We leave an extension of TuckER to include additive components to future work.

# Chapter 5

# Multi-relational Poincaré Graph Embeddings

The main contribution of this chapter is the paper *Multi-relational Poincaré Graph Embeddings* which was published at the *Neural Information Processing Systems* conference in December 2019. We first outline the motivation for this work (Section 5.1), followed by the paper itself (Section 5.2), the impact that it has had so far (Section 5.3), and discussion (Section 5.4).

## 5.1 Introduction

Knowledge graph relations exhibit multiple properties, such as symmetry, asymmetry, and transitivity. Certain relations, such as *hypernym* and *has_part*, induce a hierarchical ordering over entities. Hyperbolic space — which can be thought of as a continuous equivalent of discrete trees (Sarkar, 2011; De Sa et al., 2018) — has been shown to be particularly suited for modelling hierarchical data, as embedding entities in hyperbolic rather than Euclidean space has resulted in improved representations (e.g. Sarkar, 2011; Nickel and Kiela, 2017, 2018; Ganea et al., 2018b; Tifrea et al., 2019). However, different knowledge graph relations can induce different hierarchies over the same set of entities. For example, the noun *chair* is a parent node to different chair types (e.g. *folding_chair*, *armchair*) under the relation *hypernym* and both *chair* and its types are parent nodes to parts of a typical chair (e.g. *backrest*, *leg*) under the relation *has_part*. With the aim of capturing multiple simultaneous hierarchies, the paper in the following section introduces the Multi-Relational Poincaré (*MuRP*) model, first method to embed *hierarchical multi-relational data* in the Poincaré ball model of hyperbolic geometry.

This work was largely inspired by the work of Nickel and Kiela (2017), which were the first to use hyperbolic space in modern representation learning. The authors embed the transitive closure of the WordNet noun hierarchy in the Poincaré ball, showing that low-dimensional hyperbolic embeddings can significantly outperform higher-dimensional Euclidean embeddings in terms of both representation capacity and generalisation ability. However, their proposed model is unable to handle multiple hierarchies as they appear in knowledge graphs, which is addressed in our work. MuRP learns relation-specific parameters to transform entity embeddings by Möbius matrix-vector multiplication and Möbius addition (Ungar, 2001). We show that MuRP outperforms its Euclidean counterpart *MuRE* and existing Euclidean models on the link prediction task on the hierarchical WN18RR (Dettmers et al., 2018) knowledge graph as well as hierarchical subsets of NELL-995 (Xiong et al., 2017), and requires far lower dimensionality to achieve comparable performance to its Euclidean analogue.

## 5.2 The Paper

**Author Contributions**

The paper is co-authored by myself, Carl Allen and Timothy Hospedales. As the lead author, I conceived and developed MuRP, performed the experiments, and wrote the paper. Carl Allen participated in many discussions, developed the geometric intuition of the model score function as spheres of influence, linked the model to the theory of word embeddings and proofread the paper. Timothy Hospedales offered suggestions and helped revise the final version.

# Multi-relational Poincaré Graph Embeddings

**Ivana Balažević**[1]        **Carl Allen**[1]        **Timothy Hospedales**[1,2]

[1] School of Informatics, University of Edinburgh, UK

[2] Samsung AI Centre, Cambridge, UK

{ivana.balazevic, carl.allen, t.hospedales}@ed.ac.uk

## Abstract

Hyperbolic embeddings have recently gained attention in machine learning due to their ability to represent hierarchical data more accurately and succinctly than their Euclidean analogues. However, multi-relational knowledge graphs often exhibit multiple simultaneous hierarchies, which current hyperbolic models do not capture. To address this, we propose a model that embeds multi-relational graph data in the Poincaré ball model of hyperbolic space. Our Multi-Relational Poincaré model (MuRP) learns relation-specific parameters to transform entity embeddings by Möbius matrix-vector multiplication and Möbius addition. Experiments on the hierarchical WN18RR knowledge graph show that our Poincaré embeddings outperform their Euclidean counterpart and existing embedding methods on the link prediction task, particularly at lower dimensionality.

## 1  Introduction

Hyperbolic space can be thought of as a continuous analogue of discrete trees, making it suitable for modelling *hierarchical* data [28, 10]. Various types of hierarchical data have recently been embedded in hyperbolic space [25, 26, 16, 32], requiring relatively few dimensions and achieving promising results on downstream tasks. This demonstrates the advantage of modelling tree-like structures in spaces with constant negative curvature (hyperbolic) over zero-curvature spaces (Euclidean).

Certain data structures, such as knowledge graphs, often exhibit multiple hierarchies simultaneously. For example, *lion* is near the top of the animal food chain but near the bottom in a tree of taxonomic mammal types [22]. Despite the widespread use of hyperbolic geometry in representation learning, the only existing approach to embedding *hierarchical multi-relational graph data* in hyperbolic space [31] does not outperform Euclidean models. The difficulty with representing multi-relational data in hyperbolic space lies in finding a way to represent entities (nodes), shared across relations, such that they form a different hierarchy under different relations, e.g. nodes near the root of the tree under one relation may be leaf nodes under another. Further, many state-of-the-art approaches to modelling multi-relational data, such as DistMult [37], ComplEx [34], and TuckER [2] (i.e. *bilinear models*), rely on *inner product* as a similarity measure and there is no clear correspondence to the Euclidean inner product in hyperbolic space [32] by which these models can be converted. Existing translational approaches that use *Euclidean distance* to measure similarity, such as TransE [6] and STransE [23], can be converted to the hyperbolic domain, but do not currently compete with the bilinear models in terms of predictive performance. However, it has recently been shown in the closely related field of word embeddings [1] that the difference (i.e. relation) between word pairs that form analogies manifests as a *vector offset*, suggesting a translational approach to modelling relations.

In this paper, we propose MuRP, a theoretically inspired method to embed hierarchical multi-relational data in the *Poincaré ball* model of hyperbolic space. By considering the surface area of a hypersphere of increasing radius centered at a particular point, Euclidean space can be seen to "grow" polynomially, whereas in hyperbolic space the equivalent growth is exponential [10]. Therefore, moving outwards from the root of a tree, there is more "room" to separate leaf nodes in hyperbolic space than in Euclidean. MuRP learns relation-specific parameters that transform entity embeddings by *Möbius matrix-vector multiplication* and *Möbius addition* [35]. The model outperforms not only its Euclidean counterpart, but also current state-of-the-art models on the link prediction task on the *hierarchical* WN18RR dataset. We also show that our Poincaré embeddings require far fewer dimensions than Euclidean embeddings to achieve comparable performance. We visualize the learned embeddings and analyze the properties of the Poincaré model compared to its Euclidean analogue, such as convergence rate, performance per relation, and influence of embedding dimensionality.

## 2   Background and preliminaries

**Multi-relational link prediction**  A *knowledge graph* is a multi-relational graph representation of a collection $\mathcal{F}$ of *facts* in triple form $(e_s, r, e_o) \in \mathcal{E} \times \mathcal{R} \times \mathcal{E}$, where $\mathcal{E}$ is the set of entities (nodes) and $\mathcal{R}$ is the set of binary relations (typed directed edges) between them. If $(e_s, r, e_o) \in \mathcal{F}$, then subject entity $e_s$ is related to object entity $e_o$ by relation $r$. Knowledge graphs are often incomplete, so the aim of *link prediction* is to infer other true facts. Typically, a *score function* $\phi : \mathcal{E} \times \mathcal{R} \times \mathcal{E} \rightarrow \mathbb{R}$ is learned, that assigns a score $s = \phi(e_s, r, e_o)$ to each triple, indicating the strength of prediction that a particular triple corresponds to a true fact. A non-linearity, such as the logistic sigmoid function, is often used to convert the score to a predicted probability $p = \sigma(s) \in [0, 1]$ of the triple being true.

Knowledge graph relations exhibit multiple properties, such as symmetry, asymmetry, and transitivity. Certain knowledge graph relations, such as *hypernym* and *has_part*, induce a *hierarchical* structure over entities, suggesting that embedding them in *hyperbolic* rather than Euclidean space may lead to improved representations [28, 25, 26, 14, 32]. Based on this intuition, we focus on embedding multi-relational knowledge graph data in hyperbolic space.

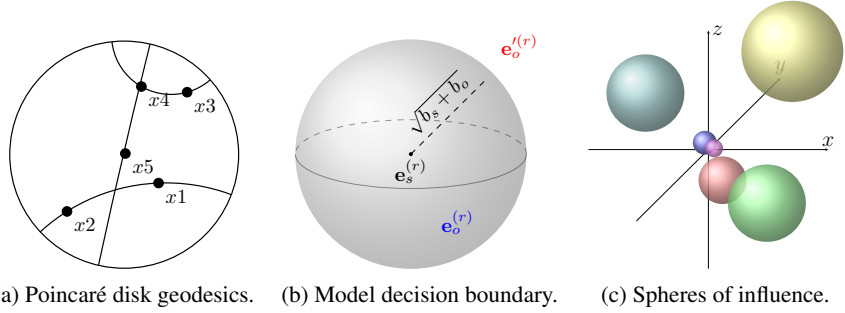

(a) Poincaré disk geodesics.    (b) Model decision boundary.    (c) Spheres of influence.

Figure 1: (a) Geodesics in the Poincaré disk, indicating the shortest paths between pairs of points. (b) The model predicts the triple $(e_s, r, e_o)$ as true and $(e_s, r, e_o')$ as false. (c) Each entity embedding has a *sphere of influence*, whose radius is determined by the entity-specific bias.

**Hyperbolic geometry of the Poincaré ball**  The Poincaré ball $(\mathbb{B}_c^d, g^{\mathbb{B}})$ of radius $1/\sqrt{c}, c > 0$ is a $d$-dimensional manifold $\mathbb{B}_c^d = \{\mathbf{x} \in \mathbb{R}^d : c\|\mathbf{x}\|^2 < 1\}$ equipped with the Riemannian metric $g^{\mathbb{B}}$ which is *conformal* to the Euclidean metric $g^{\mathbb{E}} = \mathbf{I}_d$ with the *conformal factor* $\lambda_{\mathbf{x}}^c = 2/(1 - c\|\mathbf{x}\|^2)$, i.e. $g^{\mathbb{B}} = (\lambda_{\mathbf{x}}^c)^2 g^{\mathbb{E}}$. The distance between two points $\mathbf{x}, \mathbf{y} \in \mathbb{B}_c^d$ is measured along a *geodesic* (i.e. shortest path between the points, see Figure 1a) and is given by:

$$d_{\mathbb{B}}(\mathbf{x}, \mathbf{y}) = \frac{2}{\sqrt{c}} \tanh^{-1}(\sqrt{c}\| - \mathbf{x} \oplus_c \mathbf{y}\|), \tag{1}$$

where $\| \cdot \|$ denotes the Euclidean norm and $\oplus_c$ represents *Möbius addition* [35]:

$$\mathbf{x} \oplus_c \mathbf{y} = \frac{(1 + 2c\langle \mathbf{x}, \mathbf{y} \rangle + c\|\mathbf{y}\|^2)\mathbf{x} + (1 - c\|\mathbf{x}\|^2)\mathbf{y}}{1 + 2c\langle \mathbf{x}, \mathbf{y} \rangle + c^2\|\mathbf{x}\|^2\|\mathbf{y}\|^2}, \tag{2}$$

with $\langle \cdot, \cdot \rangle$ being the Euclidean inner product. Ganea et al. [13] show that *Möbius matrix-vector multiplication* can be obtained by projecting a point $\mathbf{x} \in \mathbb{B}_c^d$ onto the tangent space at $\mathbf{0} \in \mathbb{B}_c^d$ with the *logarithmic map* $\log_{\mathbf{0}}^c(\mathbf{x})$, performing matrix multiplication by $\mathbf{M} \in \mathbb{R}^{d \times k}$ in the Euclidean tangent space, and projecting back to $\mathbb{B}_c^d$ via the *exponential map* at $\mathbf{0}$, i.e.:

$$\mathbf{M} \otimes_c \mathbf{x} = \exp_{\mathbf{0}}^c(\mathbf{M}\log_{\mathbf{0}}^c(\mathbf{x})). \tag{3}$$

For the definitions of exponential and logarithmic maps, see Appendix A.

# 3 Related work

## 3.1 Hyperbolic geometry

Embedding hierarchical data in hyperbolic space has recently gained popularity in representation learning. Nickel and Kiela [25] first embedded the *transitive closure*[1] of the WordNet noun hierarchy, in the Poincaré ball, showing that low-dimensional hyperbolic embeddings can significantly outperform higher-dimensional Euclidean embeddings in terms of both representation capacity and generalization ability. The same authors subsequently embedded hierarchical data in the Lorentz model of hyperbolic geometry [26].

Ganea et al. [13] introduced Hyperbolic Neural Networks, connecting hyperbolic geometry with deep learning. They build on the definitions for Möbius addition, Möbius scalar multiplication, exponential and logarithmic maps of Ungar [35] to derive expressions for linear layers, bias translation and application of non-linearity in the Poincaré ball. Hyperbolic analogues of several other algorithms have been developed since, such as Poincaré GloVe [32] and Hyperbolic Attention Networks [16]. More recently, Gu et al. [15] note that data can be non-uniformly hierarchical and learn embeddings on a *product manifold* with components of different curvature: spherical, hyperbolic and Euclidean. To our knowledge, only Riemannian TransE [31] seeks to embed *multi-relational* data in hyperbolic space, but the Riemannian translation method fails to outperform Euclidean baselines.

## 3.2 Link prediction for knowledge graphs

**Bilinear models** typically represent relations as linear transformations acting on entity vectors. An early model, RESCAL [24], optimizes a score function $\phi(e_s, r, e_o) = \mathbf{e}_s^\top \mathbf{M}_r \mathbf{e}_o$, containing the bilinear product between the subject entity embedding $\mathbf{e}_s$, a full rank relation matrix $\mathbf{M}_r$ and the object entity embedding $\mathbf{e}_o$. RESCAL is prone to overfitting due to the number of parameters per relation being quadratic relative to the number per entity. DistMult [37] is a special case of RESCAL with diagonal relation matrices, reducing parameters per relation and controlling overfitting. However, due to its symmetry, DistMult cannot model asymmetric relations. ComplEx [34] extends DistMult to the complex domain, enabling asymmetry to be modelled. TuckER [2] performs a Tucker decomposition of the tensor of triples, which enables multi-task learning between different relations via the core tensor. The authors show each of the linear models above to be a special case of TuckER.

**Translational models** regard a relation as a translation (or vector offset) from the subject to the object entity embeddings. These models include TransE [6] and its many successors, e.g. FTransE [12], STransE [23]. The score function for translational models typically considers Euclidean distance between the translated subject entity embedding and the object entity embedding.

# 4 Multi-relational Poincaré embeddings

A set of entities can form different hierarchies under different relations. In the WordNet knowledge graph [22], the *hypernym*, *has_part* and *member_meronym* relations each induce different hierarchies over the same set of entities. For example, the noun *chair* is a parent node to different chair types (e.g. *folding_chair*, *armchair*) under the relation *hypernym* and both *chair* and its types are parent nodes to parts of a typical chair (e.g. *backrest*, *leg*) under the relation *has_part*. An ideal embedding model should capture all hierarchies simultaneously.

**Score function** As mentioned above, bilinear models measure similarity between the subject entity embedding (after relation-specific transformation) and an object entity embedding using the Euclidean inner product [24, 37, 34, 2]. However, a clear correspondence to the Euclidean inner product does not exist in hyperbolic space [32]. The Euclidean inner product can be expressed as a function of Euclidean distance and norms, i.e. $\langle \mathbf{x}, \mathbf{y} \rangle = \frac{1}{2}(-d_\mathbb{E}(\mathbf{x}, \mathbf{y})^2 + \|\mathbf{x}\|^2 + \|\mathbf{y}\|^2)$, $d_\mathbb{E}(\mathbf{x}, \mathbf{y}) = \|\mathbf{x} - \mathbf{y}\|$. Noting this, in Poincaré GloVe, Tifrea et al. [32] absorb squared norms into biases $b_\mathbf{x}, b_\mathbf{y}$ and replace the Euclidean with the Poincaré distance $d_\mathbb{B}(\mathbf{x}, \mathbf{y})$ to obtain the hyperbolic version of GloVe [27].

Separately, it has recently been shown in the closely related field of word embeddings that statistics pertaining to *analogies* naturally contain linear structures [1], explaining why similar linear structure appears amongst word embeddings of word2vec [20, 21, 19]. Analogies are word relationships of the form "$w_a$ is to $w_a^*$ as $w_b$ is to $w_b^*$", such as "*man is to woman as king is to queen*", and are in principle not restricted to two pairs (e.g. "...as *brother* is to *sister*"). It can be seen that analogies have much in common with *relations* in multi-relational graphs, as a difference between pairs of words (or entities) common to all pairs, e.g. if $(e_s, r, e_o)$ and $(e_s', r, e_o')$ hold, then we could say "$e_s$ is to $e_o$ as $e_s'$ is to $e_o'$". Of particular relevance is the demonstration that the common difference, i.e. relation, between the word pairs (e.g. (*man, woman*) and (*king, queen*)) manifests as a *common vector offset* [1], justifying the previously heuristic translational approach to modelling relations.

---

[1] Each node in a directed graph is connected not only to its children, but to every descendant, i.e. all nodes to which there exists a directed path from the starting node.

Inspired by these two ideas, we define the basis score function for multi-relational graph embedding:

$$\begin{aligned}
\phi(e_s, r, e_o) &= -d(\mathbf{e}_s^{(r)}, \mathbf{e}_o^{(r)})^2 + b_s + b_o \\
&= -d(\mathbf{R}\mathbf{e}_s, \mathbf{e}_o + \mathbf{r})^2 + b_s + b_o,
\end{aligned} \tag{4}$$

where $d : \mathcal{E} \times \mathcal{R} \times \mathcal{E} \to \mathbb{R}^+$ is a distance function, $\mathbf{e}_s, \mathbf{e}_o \in \mathbb{R}^d$ are the embeddings and $b_s, b_o \in \mathbb{R}$ scalar biases of the subject and object entities $e_s$ and $e_o$ respectively. $\mathbf{R} \in \mathbb{R}^{d \times d}$ is a diagonal relation matrix and $\mathbf{r} \in \mathbb{R}^d$ a translation vector (i.e. vector offset) of relation $r$. $\mathbf{e}_s^{(r)} = \mathbf{R}\mathbf{e}_s$ and $\mathbf{e}_o^{(r)} = \mathbf{e}_o + \mathbf{r}$ represent the subject and object entity embeddings after applying the respective relation-specific transformations, a stretch by $\mathbf{R}$ to $\mathbf{e}_s$ and a translation by $\mathbf{r}$ to $\mathbf{e}_o$.

**Hyperbolic model** Taking the hyperbolic analogue of Equation 4, we define the score function for our ***Multi-Relational Poincaré (MuRP)*** model as:

$$\begin{aligned}
\phi_{\text{MuRP}}(e_s, r, e_o) &= -d_{\mathbb{B}}(\mathbf{h}_s^{(r)}, \mathbf{h}_o^{(r)})^2 + b_s + b_o \\
&= -d_{\mathbb{B}}(\exp_{\mathbf{0}}^c(\mathbf{R}\log_{\mathbf{0}}^c(\mathbf{h}_s)), \mathbf{h}_o \oplus_c \mathbf{r}_h)^2 + b_s + b_o,
\end{aligned} \tag{5}$$

where $\mathbf{h}_s, \mathbf{h}_o \in \mathbb{B}_c^d$ are *hyperbolic embeddings* of the subject and object entities $e_s$ and $e_o$ respectively, and $\mathbf{r}_h \in \mathbb{B}_c^d$ is a *hyperbolic translation vector* of relation $r$. The relation-adjusted subject entity embedding $\mathbf{h}_s^{(r)} \in \mathbb{B}_c^d$ is obtained by *Möbius matrix-vector multiplication*: the original subject entity embedding $\mathbf{h}_s \in \mathbb{B}_c^d$ is projected to the tangent space of the Poincaré ball at $\mathbf{0}$ with $\log_{\mathbf{0}}^c$, transformed by the diagonal relation matrix $\mathbf{R} \in \mathbb{R}^{d \times d}$, and then projected back to the Poincaré ball by $\exp_{\mathbf{0}}^c$. The relation-adjusted object entity embedding $\mathbf{h}_o^{(r)} \in \mathbb{B}_c^d$ is obtained by *Möbius addition* of the relation vector $\mathbf{r}_h \in \mathbb{B}_c^d$ to the object entity embedding $\mathbf{h}_o \in \mathbb{B}_c^d$. Since the relation matrix $\mathbf{R}$ is diagonal, the number of parameters of MuRP increases *linearly* with the number of entities and relations, making it scalable to large knowledge graphs. To obtain the predicted probability of a fact being true, we apply the logistic sigmoid to the score, i.e. $\sigma(\phi_{\text{MuRP}}(e_s, r, e_o))$.

To directly compare the properties of hyperbolic embeddings with the Euclidean, we implement the Euclidean version of Equation 4 with $d(\mathbf{e}_s^{(r)}, \mathbf{e}_o^{(r)}) = d_{\mathbb{E}}(\mathbf{e}_s^{(r)}, \mathbf{e}_o^{(r)})$. We refer to this model as the ***Multi-Relational Euclidean (MuRE)*** model.

**Geometric intuition** We see from Equation 4 that the biases $b_s, b_o$ determine the radius of a *hypersphere decision boundary* centered at $\mathbf{e}_s^{(r)}$. Entities $e_s$ and $e_o$ are predicted to be related by $r$ if relation-adjusted $\mathbf{e}_o^{(r)}$ falls within a hypersphere of radius $\sqrt{b_s + b_o}$ (see Figure 1b). Since biases are subject and object entity-specific, each subject-object pair induces a different decision boundary. The relation-specific parameters $\mathbf{R}$ and $\mathbf{r}$ determine the position of the relation-adjusted embeddings, but the radius of the entity-specific decision boundary is independent of the relation. The score function in Equation 4 resembles the score functions of existing translational models [6, 12, 23], with the main difference being the entity-specific biases, which can be seen to change the geometry of the model. Rather than considering an entity as a point in space, each bias defines an entity-specific *sphere of influence* surrounding the center given by the embedding vector (see Figure 1c). The overlap between spheres measures relatedness between entities. We can thus think of each relation as moving the spheres of influence in space, so that only the spheres of subject and object entities that are connected under that relation overlap.

### 4.1   Training and Riemannian optimization

We use the standard data augmentation technique [11, 18, 2] of adding reciprocal relations for every triple, i.e. we add $(e_o, r^{-1}, e_s)$ for every $(e_s, r, e_o)$. To train both models, we generate $k$ negative samples for each true triple $(e_s, r, e_o)$, where we corrupt either the object $(e_s, r, e_o')$ or the subject $(e_o, r^{-1}, e'_s)$ entity with a randomly chosen entity from the set of all entities $\mathcal{E}$. Both models are trained to minimize the Bernoulli negative log-likelihood loss:

$$\mathcal{L}(y, p) = -\frac{1}{N} \sum_{i=1}^{N} (y^{(i)} \log(p^{(i)}) + (1 - y^{(i)}) \log(1 - p^{(i)})), \tag{6}$$

where $p$ is the predicted probability, $y$ is the binary label indicating whether a sample is positive or negative and $N$ is the number of training samples.

For fairness of comparison, we optimize the Euclidean model using stochastic gradient descent (SGD) and the hyperbolic model using *Riemannian stochastic gradient descent (RSGD)* [5]. We note that the Riemannian equivalent of adaptive optimization methods has recently been developed [3], but leave replacing SGD and RSGD with their adaptive equivalent to future work. To compute the Riemannian gradient $\nabla_R \mathcal{L}$, the Euclidean gradient $\nabla_E \mathcal{L}$ is multiplied by the inverse of the Poincaré metric tensor, i.e. $\nabla_R \mathcal{L} = 1/(\lambda_\theta^c)^2 \nabla_E \mathcal{L}$. Instead of the Euclidean update step $\theta \leftarrow \theta - \eta \nabla_E \mathcal{L}$, a first order approximation of the true Riemannian update, we use $\exp_\theta^c$ to project the gradient $\nabla_R \mathcal{L} \in T_\theta \mathbb{B}_c^d$ onto its corresponding geodesic on the Poincaré ball and compute the Riemannian update $\theta \leftarrow \exp_\theta^c(-\eta \nabla_R \mathcal{L})$, where $\eta$ denotes the learning rate.

# 5 Experiments

To evaluate both Poincaré and Euclidean models, we first test their performance on the knowledge graph link prediction task using standard WN18RR and FB15k-237 datasets:

**FB15k-237** [33] is a subset of Freebase [4], a collection of real world facts, created from FB15k [6] by removing the inverse of many relations from validation and test sets to make the dataset more challenging. FB15k-237 contains 14,541 entities and 237 relations.

**WN18RR** [11] is a subset of WordNet [22], a hierarchical collection of relations between words, created in the same way as FB15k-237 from WN18 [6], containing 40,943 entities and 11 relations.

To demonstrate the usefulness of MuRP on hierarchical datasets (given WN18RR is hierarchical and FB15k-237 is not, see Section 5.3), we also perform experiments on NELL-995 [36], containing 75,492 entities and 200 relations, $\sim 22\%$ of which hierarchical. We create several subsets of the original dataset by varying the proportion of non-hierarchical relations, as described in Appendix B.

We evaluate each triple from the test set by generating $n_e$ (where $n_e$ denotes number of entities in the dataset) *evaluation triples*, which are created by combining the test entity-relation pair with all possible entities $\mathcal{E}$. The scores obtained for each evaluation triple are ranked. All true triples are removed from the evaluation triples apart from the current test triple, i.e. the commonly used *filtered setting* [6]. We evaluate our models using the evaluation metrics standard across the link prediction literature: mean reciprocal rank (MRR) and hits@$k$, $k \in \{1, 3, 10\}$. Mean reciprocal rank is the average of the inverse of a mean rank assigned to the true triple over all $n_e$ evaluation triples. Hits@$k$ measures the percentage of times the true triple appears in the top $k$ ranked evaluation triples.

## 5.1 Implementation details

We implement both models in PyTorch and make our code, as well as all the subsets of the NELL-995 dataset, publicly available.[2] We choose the learning rate from $\{1, 5, 10, 20, 50, 100\}$ by MRR on the validation set and find that the best learning rate is 50 for WN18RR and 10 for FB15k-237 for both models. We initialize all embeddings near the origin where distances are small in hyperbolic space, similar to [25]. We set the batch size to 128 and the number of negative samples to 50. In all experiments, we set the curvature of MuRP to $c = 1$, since preliminary experiments showed that any material change reduced performance.

## 5.2 Link prediction results

Table 1 shows the results obtained for both datasets. As expected, MuRE performs slightly better on the non-hierarchical FB15k-237 dataset, whereas MuRP outperforms on WN18RR which contains hierarchical relations (as shown in Section 5.3). Both MuRE and MuRP outperform previous state-of-the-art models on WN18RR on all metrics apart from hits@1, where MuRP obtains second best overall result. In fact, even at relatively low embedding dimensionality ($d = 40$), this is maintained, demonstrating the ability of hyperbolic models to succinctly represent multiple hierarchies. On FB15k-237, MuRE is outperformed only by TuckER [2] (and similarly ComplEx-N3 [18], since Balažević et al. [2] note that the two models perform comparably), primarily due to *multi-task learning* across relations. This is highly advantageous on FB15k-237 due to a large number of relations compared to WN18RR and thus relatively little data per relation in some cases. As the first model to successfully represent multiple relations in hyperbolic space, MuRP does not also set out to include multi-task learning, but we hope to address this in future work. Further experiments on NELL-995, which substantiate our claim on the advantage of embedding hierarchical multi-relational data in hyperbolic over Euclidean space, are presented in Appendix C.

---

[2]`https://github.com/ibalazevic/multirelational-poincare`

Table 1: Link prediction results on WN18RR and FB15k-237. Best results in bold and underlined, second best in bold. The RotatE [30] results are reported without their self-adversarial negative sampling (see Appendix H in the original paper) for fair comparison.

| | WN18RR | | | | FB15k-237 | | | |
|---|---|---|---|---|---|---|---|---|
| | MRR | Hits@10 | Hits@3 | Hits@1 | MRR | Hits@10 | Hits@3 | Hits@1 |
| TransE [6] | .226 | .501 | — | — | .294 | .465 | — | — |
| DistMult [37] | .430 | .490 | .440 | .390 | .241 | .419 | .263 | .155 |
| ComplEx [34] | .440 | .510 | .460 | .410 | .247 | .428 | .275 | .158 |
| Neural LP [38] | — | — | — | — | .250 | .408 | — | — |
| MINERVA [9] | — | — | — | — | — | .456 | — | — |
| ConvE [11] | .430 | .520 | .440 | .400 | .325 | .501 | .356 | .237 |
| M-Walk [29] | .437 | — | .445 | .414 | — | — | — | — |
| TuckER [2] | .470 | .526 | .482 | **.443** | .358 | .544 | **.394** | **.266** |
| RotatE [30] | — | — | — | — | .297 | .480 | .328 | .205 |
| MuRE $d = 40$ | .459 | .528 | .474 | .429 | .315 | .493 | .346 | .227 |
| MuRE $d = 200$ | .475 | .554 | .487 | .436 | **.336** | **.521** | **.370** | **.245** |
| MuRP $d = 40$ | **.477** | **.555** | **.489** | .438 | .324 | .506 | .356 | .235 |
| MuRP $d = 200$ | **.481** | **.566** | **.495** | **.440** | .335 | .518 | .367 | .243 |

### 5.3   MuRE vs MuRP

**Effect of dimensionality**  We compare the MRR achieved by MuRE and MuRP on WN18RR for embeddings of different dimensionalities $d \in \{5, 10, 15, 20, 40, 100, 200\}$. As expected, the difference is greatest at lower embedding dimensionality (see Figure 2a).

**Convergence rate**  Figure 2b shows the MRR per epoch for MuRE and MuRP on the WN18RR training and validation sets, showing that MuRP also converges faster.

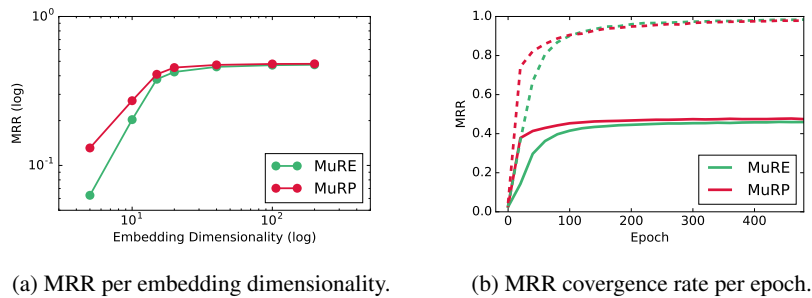

(a) MRR per embedding dimensionality.          (b) MRR covergence rate per epoch.

Figure 2: (a) MRR log-log graph for MuRE and MuRP for different embeddings sizes on WN18RR. (b) Comparison of the MRR convergence rate for MuRE and MuRP on the WN18RR training (dashed line) and validation (solid line) sets with embeddings of size $d = 40$ and learning rate 50.

**Model architecture ablation study**  Table 2 shows an ablation study of relation-specific transformations and bias choices. We note that any change to the current model architecture has a negative effect on performance of both MuRE and MuRP. Replacing biases by the (transformed) entity embedding norms leads to a significant reduction in performance of MuRP, in part because norms are constrained to $[0, 1)$, whereas the biases they replace are unbounded.

Table 2: Ablation study of different model architecture choices on WN18RR: relational transformations (left) and biases (right). Current model (top row) outperforms all others.

(a) Relational transformations.

| Distance function | MuRE | | MuRP | |
|---|---|---|---|---|
| | MRR | H@1 | MRR | H@1 |
| $d(\mathbf{Re}_s, \mathbf{e}_o + \mathbf{r})$ | **.459** | **.429** | **.477** | **.438** |
| $d(\mathbf{e}_s, \mathbf{e}_o + \mathbf{r})$ | .340 | .235 | .307 | .192 |
| $d(\mathbf{Re}_s, \mathbf{e}_o)$ | .413 | .381 | .401 | .363 |
| $d(\mathbf{R}_s\mathbf{e}_s, \mathbf{R}_o\mathbf{e}_o + \mathbf{r})$ | .341 | .299 | .367 | .335 |
| $d(\mathbf{e}_s + \mathbf{r}, \mathbf{Re}_o)$ | .442 | .410 | .454 | .413 |

(b) Biases.

| Bias choice | MuRE | | MuRP | |
|---|---|---|---|---|
| | MRR | H@1 | MRR | H@1 |
| $b_s$ & $b_o$ | **.459** | **.429** | **.477** | **.438** |
| $b_s$ only | .455 | .414 | .463 | .415 |
| $b_o$ only | .453 | .412 | .460 | .409 |
| $b_x = \|\mathbf{e}_x\|^2$ | .414 | .393 | .414 | .352 |
| $b_x = \|\mathbf{e}_x^r\|^2$ | .443 | .404 | .434 | .372 |

**Performance per relation** Since not every relation in WN18RR induces a hierarchical structure over the entities, we report the *Krackhardt hierarchy score (Khs)* [17] of the entity graph formed by each relation to obtain a measure of the hierarchy induced. The score is defined only for directed networks and measures the proportion of node pairs $(x, y)$ where there exists a directed path $x \to y$, but not $y \to x$ (see Appendix D for further details). The score takes a value of one for all directed acyclic graphs, and zero for cycles and cliques. We also report the maximum and average shortest path between any two nodes in the graph for hierarchical relations. To gain insight as to which relations benefit most from embedding entities in hyperbolic space, we compare hits@10 per relation of MuRE and MuRP for entity embeddings of low dimensionality ($d = 20$). From Table 3 we see that both models achieve comparable performance on non-hierarchical, symmetric relations with the Krackhardt hierarchy score 0, such as *verb_group*, whereas MuRP generally outperforms MuRE on hierarchical relations. We also see that the difference between the performances of MuRE and MuRP is generally larger for relations that form deeper trees, fitting the hypothesis that hyperbolic space is of most benefit for modelling hierarchical relations.

Computing the Krackhardt hierarchy score for FB15k-237, we find that 80% of the relations have Khs = 1, however, the average of maximum path lengths over those relations is 1.14 with only 2.7% relations having paths longer than 2, meaning that the vast majority of relational sub-graphs consist of directed edges between *pairs of nodes*, rather than trees.

Table 3: Comparison of hits@10 per relation for MuRE and MuRP on WN18RR for $d = 20$.

| Relation Name | MuRE | MuRP | $\Delta$ | Khs | Max Path | Avg Path |
|---|---|---|---|---|---|---|
| also_see | .634 | **.705** | .071 | 0.24 | 44 | 15.2 |
| hypernym | .161 | **.228** | .067 | 0.99 | 18 | 4.5 |
| has_part | .215 | **.282** | .067 | 1 | 13 | 2.2 |
| member_meronym | .272 | **.346** | .074 | 1 | 10 | 3.9 |
| synset_domain_topic_of | .316 | **.430** | .114 | 0.99 | 3 | 1.1 |
| instance_hypernym | **.488** | .471 | −.017 | 1 | 3 | 1.0 |
| member_of_domain_region | .308 | **.347** | .039 | 1 | 2 | 1.0 |
| member_of_domain_usage | .396 | **.417** | .021 | 1 | 2 | 1.0 |
| derivationally_related_form | .954 | **.967** | .013 | 0.04 | – | – |
| similar_to | **1** | **1** | 0 | 0 | – | – |
| verb_group | **.974** | **.974** | 0 | 0 | – | – |

**Biases vs embedding vector norms** We plot the norms versus the biases $b_s$ for MuRP and MuRE in Figure 3. This shows an overall correlation between embedding vector norm and bias (or radius of the sphere of influence) for both MuRE and MuRP. This makes sense intuitively, as the sphere of influence increases to "fill out the space" in regions that are less cluttered, i.e. further from the origin.

**Spatial layout** Figure 4 shows a 40-dimensional subject embedding for the word *asia* and a random subset of 1500 object embeddings for the hierarchical WN18RR relation *has_part*, projected to 2 dimensions so that distances and angles of object entity embeddings *relative to the subject entity embedding* are preserved (see Appendix E for details on the projection method). We show subject and object entity embeddings before and after relation-specific transformation. For both MuRE and MuRP, we see that applying the relation-specific transformation separates true object entities from false ones. However, in the Poincaré model, where distances increase further from the origin, embeddings are moved further towards the boundary of the disk, where, loosely speaking, there is *more space* to separate and therefore distinguish them.

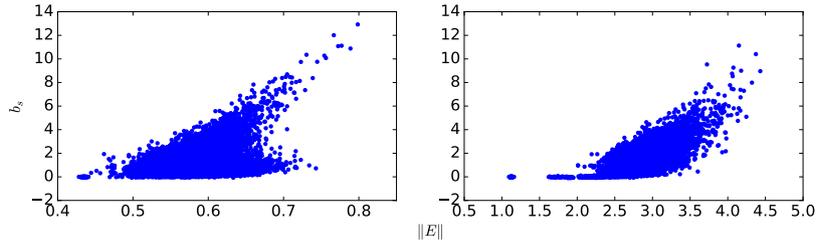

Figure 3: Scatter plot of norms vs biases for MuRP (left) and MuRE (right). Entities with larger embedding vector norms generally have larger biases for both MuRE and MuRP.

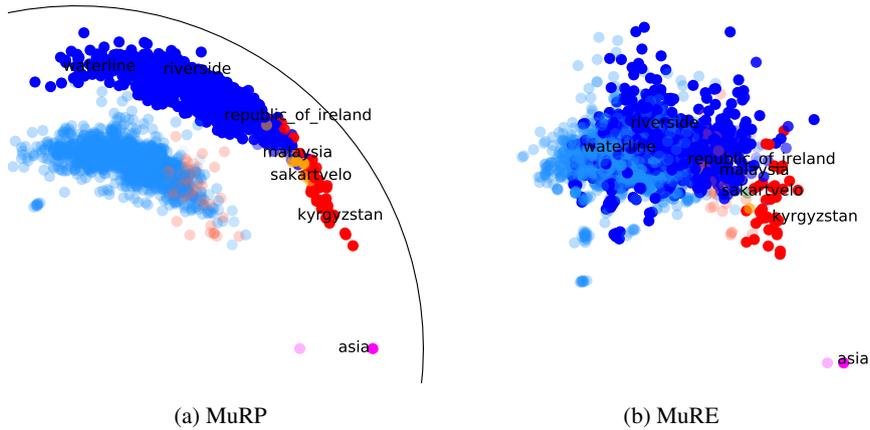

|                      |                      |
|:--------------------:|:--------------------:|
| (a) MuRP             | (b) MuRE             |

Figure 4: Learned 40-dimensional MuRP and MuRE embeddings for WN18RR relation *has_part*, projected to 2 dimensions. ● indicates the subject entity embedding, ● indicates true positive object entities predicted by the model, ● true negatives, ● false positives and ● false negatives. Lightly shaded blue and red points indicate object entity embeddings before applying the relation-specific transformation. The line in the left figure indicates the boundary of the Poincaré disk. The supposed false positives predicted by MuRP are actually true facts missing from the dataset (e.g. *malaysia*).

**Analysis of wrong predictions** Here we analyze the false positives and false negatives predicted by both models. MuRP predicts 15 false positives and 0 false negatives, whereas MuRE predicts only 2 false positives and 1 false negative, so seemingly performs better. However, inspecting the alleged false positives predicted by MuRP, we find they are all countries on the Asian continent (e.g. *sri_lanka*, *palestine*, *malaysia*, *sakartvelo*, *thailand*), so are actually correct, but missing from the dataset. MuRE's predicted false positives (*philippines* and *singapore*) are both also correct but missing, whereas the false negative (*bahrain*) is indeed falsely predicted. We note that this suggests current evaluation methods may be unreliable.

## 6   Conclusion and future work

We introduce a novel, theoretically inspired, translational method for embedding multi-relational graph data in the Poincaré ball model of hyperbolic geometry. Our multi-relational Poincaré model MuRP learns relation-specific parameters to transform entity embeddings by Möbius matrix-vector multiplication and Möbius addition. We show that MuRP outperforms its Euclidean counterpart MuRE and existing models on the link prediction task on the hierarchical WN18RR knowledge graph dataset, and requires far lower dimensionality to achieve comparable performance to its Euclidean analogue. We analyze various properties of the Poincaré model compared to its Euclidean analogue and provide insight through a visualization of the learned embeddings.

Future work may include investigating the impact of recently introduced Riemannian adaptive optimization methods compared to Riemannian SGD. Also, given not all relations in a knowledge graph are hierarchical, we may look into combining the Euclidean and hyperbolic models to produce mixed-curvature embeddings that best fit the curvature of the data.

**Acknowledgements**

We thank Rik Sarkar, Ivan Titov, Jonathan Mallinson, Eryk Kopczyński and the anonymous reviewers for helpful comments. Ivana Balažević and Carl Allen were supported by the Centre for Doctoral Training in Data Science, funded by EPSRC (grant EP/L016427/1) and the University of Edinburgh.

# References

[1] Carl Allen and Timothy Hospedales. Analogies Explained: Towards Understanding Word Embeddings. In *International Conference on Machine Learning*, 2019.

[2] Ivana Balažević, Carl Allen, and Timothy M Hospedales. TuckER: Tensor Factorization for Knowledge Graph Completion. In *Empirical Methods in Natural Language Processing*, 2019.

[3] Gary Bécigneul and Octavian-Eugen Ganea. Riemannian Adaptive Optimization Methods. In *International Conference on Learning Representation*, 2019.

[4] Kurt Bollacker, Colin Evans, Praveen Paritosh, Tim Sturge, and Jamie Taylor. Freebase: A Collaboratively Created Graph Database for Structuring Human Knowledge. In *ACM SIGMOD International Conference on Management of Data*, 2008.

[5] Silvere Bonnabel. Stochastic Gradient Descent on Riemannian Manifolds. *IEEE Transactions on Automatic Control*, 2013.

[6] Antoine Bordes, Nicolas Usunier, Alberto Garcia-Duran, Jason Weston, and Oksana Yakhnenko. Translating Embeddings for Modeling Multi-relational Data. In *Advances in Neural Information Processing Systems*, 2013.

[7] James W Cannon, William J Floyd, Richard Kenyon, Walter R Parry, et al. Hyperbolic Geometry. *Flavors of Geometry*, 31:59–115, 1997.

[8] Andrew Carlson, Justin Betteridge, Bryan Kisiel, Burr Settles, Estevam R Hruschka, and Tom M Mitchell. Toward an Architecture for Never-ending Language Learning. In *AAAI Conference on Artificial Intelligence*, 2010.

[9] Rajarshi Das, Shehzaad Dhuliawala, Manzil Zaheer, Luke Vilnis, Ishan Durugkar, Akshay Krishnamurthy, Alex Smola, and Andrew McCallum. Go for a Walk and Arrive at the Answer: Reasoning over Paths in Knowledge Bases Using Reinforcement Learning. In *International Conference on Learning Representations*, 2018.

[10] Christopher De Sa, Albert Gu, Christopher Ré, and Frederic Sala. Representation Tradeoffs for Hyperbolic Embeddings. In *International Conference on Machine Learning*, 2018.

[11] Tim Dettmers, Pasquale Minervini, Pontus Stenetorp, and Sebastian Riedel. Convolutional 2D Knowledge Graph Embeddings. In *AAAI Conference on Artificial Intelligence*, 2018.

[12] Jun Feng, Minlie Huang, Mingdong Wang, Mantong Zhou, Yu Hao, and Xiaoyan Zhu. Knowledge Graph Embedding by Flexible Translation. In *Principles of Knowledge Representation and Reasoning*, 2016.

[13] Octavian Ganea, Gary Bécigneul, and Thomas Hofmann. Hyperbolic Neural Networks. In *Advances in Neural Information Processing Systems*, 2018.

[14] Octavian-Eugen Ganea, Gary Bécigneul, and Thomas Hofmann. Hyperbolic Entailment Cones for Learning Hierarchical Embeddings. In *International Conference on Machine Learning*, 2018.

[15] Albert Gu, Frederic Sala, Beliz Gunel, and Christopher Ré. Learning Mixed-Curvature Representations in Product Spaces. In *International Conference on Learning Representations*, 2019.

[16] Caglar Gulcehre, Misha Denil, Mateusz Malinowski, Ali Razavi, Razvan Pascanu, Karl Moritz Hermann, Peter Battaglia, Victor Bapst, David Raposo, Adam Santoro, and Nando de Freitas. Hyperbolic Attention Networks. In *International Conference on Learning Representations*, 2019.

[17]  David Krackhardt. Graph Theoretical Dimensions of Informal Organizations. In *Computational Organization Theory*. Psychology Press, 1994.

[18]  Timothée Lacroix, Nicolas Usunier, and Guillaume Obozinski. Canonical Tensor Decomposition for Knowledge Base Completion. In *International Conference on Machine Learning*, 2018.

[19]  Omer Levy and Yoav Goldberg. Linguistic Regularities in Sparse and Explicit Word Representations. In *Computational Natural Language Learning*, 2014.

[20]  Tomas Mikolov, Ilya Sutskever, Kai Chen, Greg S Corrado, and Jeff Dean. Distributed Representations of Words and Phrases and their Compositionality. In *Advances in Neural Information Processing Systems*, 2013.

[21]  Tomas Mikolov, Wen-tau Yih, and Geoffrey Zweig. Linguistic Regularities in Continuous Space Word Representations. In *North American Chapter of the Association for Computational Linguistics: Human Language Technologies*, 2013.

[22]  George A Miller. WordNet: a Lexical Database for English. *Communications of the ACM*, 1995.

[23]  Dat Quoc Nguyen, Kairit Sirts, Lizhen Qu, and Mark Johnson. STransE: a Novel Embedding Model of Entities and Relationships in Knowledge Bases. In *North American Chapter of the Association for Computational Linguistics: Human Language Technologies*, 2016.

[24]  Maximilian Nickel, Volker Tresp, and Hans-Peter Kriegel. A Three-Way Model for Collective Learning on Multi-Relational Data. In *International Conference on Machine Learning*, 2011.

[25]  Maximillian Nickel and Douwe Kiela. Poincaré Embeddings For Learning Hierarchical Representations. In *Advances in Neural Information Processing Systems*, 2017.

[26]  Maximillian Nickel and Douwe Kiela. Learning Continuous Hierarchies in the Lorentz Model of Hyperbolic Geometry. In *International Conference on Machine Learning*, 2018.

[27]  Jeffrey Pennington, Richard Socher, and Christopher Manning. GloVe: Global Vectors for Word Representation. In *Empirical Methods in Natural Language Processing*, 2014.

[28]  Rik Sarkar. Low Distortion Delaunay Embedding of Trees in Hyperbolic Plane. In *International Symposium on Graph Drawing*, 2011.

[29]  Yelong Shen, Jianshu Chen, Po-Sen Huang, Yuqing Guo, and Jianfeng Gao. M-Walk: Learning to Walk over Graphs using Monte Carlo Tree Search. In *Advances in Neural Information Processing Systems*, 2018.

[30]  Zhiqing Sun, Zhi-Hong Deng, Jian-Yun Nie, and Jian Tang. RotatE: Knowledge Graph Embedding by Relational Rotation in Complex Space. In *International Conference on Learning Representations*, 2019.

[31]  Atsushi Suzuki, Yosuke Enokida, and Kenji Yamanishi. Riemannian TransE: Multi-relational Graph Embedding in Non-Euclidean Space, 2019. URL https://openreview.net/forum?id=r1xRW3A9YX.

[32]  Alexandru Tifrea, Gary Bécigneul, and Octavian-Eugen Ganea. Poincaré GloVe: Hyperbolic Word Embeddings. In *International Conference on Learning Representations*, 2019.

[33]  Kristina Toutanova, Danqi Chen, Patrick Pantel, Hoifung Poon, Pallavi Choudhury, and Michael Gamon. Representing Text for Joint Embedding of Text and Knowledge Bases. In *Empirical Methods in Natural Language Processing*, 2015.

[34]  Théo Trouillon, Johannes Welbl, Sebastian Riedel, Éric Gaussier, and Guillaume Bouchard. Complex Embeddings for Simple Link Prediction. In *International Conference on Machine Learning*, 2016.

[35]  Abraham A Ungar. Hyperbolic Trigonometry and its Application in the Poincaré Ball Model of Hyperbolic Geometry. *Computers & Mathematics with Applications*, 41(1-2):135–147, 2001.

[36] Wenhan Xiong, Thien Hoang, and William Yang Wang. DeepPath: A Reinforcement Learning Method for Knowledge Graph Reasoning. In *Empirical Methods in Natural Language Processing*, 2017.

[37] Bishan Yang, Wen-tau Yih, Xiaodong He, Jianfeng Gao, and Li Deng. Embedding Entities and Relations for Learning and Inference in Knowledge Bases. In *International Conference on Learning Representations*, 2015.

[38] Fan Yang, Zhilin Yang, and William W Cohen. Differentiable Learning of Logical Rules for Knowledge Base Reasoning. In *Advances in Neural Information Processing Systems*, 2017.

## A    Poincaré ball model of hyperbolic geometry

The Poincaré ball model is one of five *isometric* models of hyperbolic geometry [7], each offering different perspectives for performing mathematical operations in hyperbolic space. The isometry means there exists a one-to-one distance-preserving mapping from the *metric space* of one model $(\mathcal{X}, d)$ onto that of another $(\mathcal{X}', d')$, where $\mathcal{X}, \mathcal{X}'$ are sets and $d, d'$ distance functions, or *metrics*, providing a notion of equivalence between the models.

Each point on the Poincaré ball $\mathbf{x} \in \mathbb{B}_c^d$ has a *tangent space* $T_\mathbf{x} \mathbb{B}_c^d$, a $d$-dimensional vector space, that is a local first-order approximation of the manifold $\mathbb{B}_c^d$ around $\mathbf{x}$, which for the Poincaré ball $\mathbb{B}_c^d$ is a $d$-dimensional Euclidean space, i.e. $T_\mathbf{x} \mathbb{B}_c^d = \mathbb{R}^d$. The *exponential map* $\exp_\mathbf{x}^c : T_\mathbf{x} \mathbb{B}_c^d \to \mathbb{B}_c^d$ allows one to move on the manifold from $\mathbf{x}$ in the direction of a vector $\mathbf{v} \in T_\mathbf{x} \mathbb{B}_c^d$, tangential to $\mathbb{B}_c^d$ at $\mathbf{x}$. The inverse is the *logarithmic map* $\log_\mathbf{x}^c : \mathbb{B}_c^d \to T_\mathbf{x} \mathbb{B}_c^d$. For the Poincaré ball, these are defined [13] as:

$$\exp_\mathbf{x}^c(\mathbf{v}) = \mathbf{x} \oplus_c \left( \tanh\left( \sqrt{c} \frac{\lambda_\mathbf{x}^c \|\mathbf{v}\|}{2} \right) \frac{\mathbf{v}}{\sqrt{c}\|\mathbf{v}\|} \right) \tag{7}$$

$$\log_\mathbf{x}^c(\mathbf{y}) = \frac{2}{\sqrt{c}\lambda_\mathbf{x}^c} \tanh^{-1}(\sqrt{c}\| - \mathbf{x} \oplus_c \mathbf{y}\|) \frac{-\mathbf{x} \oplus_c \mathbf{y}}{\| - \mathbf{x} \oplus_c \mathbf{y}\|}. \tag{8}$$

## B    NELL-995-h{100, 75, 50, 25} dataset splits

NELL-995 [36] is a subset of the Never-Ending Language Learner (NELL) [8]. The commonly used test set of NELL-995 [36] contains only 12 out of 200 relations present in the training set, none of which are hierarchical. To ensure a fair representation of all training set relations in the validation and test sets, we create new validation and test set splits by combining the initial validation and test sets with the training set and randomly selecting 10,000 triples each from the combined dataset.

To evaluate the influence of the proportion of hierarchical relations in a dataset on the difference in performance between MuRE and MuRP, we create four subsets of the newly created NELL-995 dataset split, containing 100%, 75%, 50% and 25% hierarchical relations, named NELL-995-h100, NELL-995-h75, NELL-995-h50 and NELL-995-h25, containing 43 hierarchical relations each and 0, 14, 43 and 129 non-hierarchical relations respectively.

## C    NELL-995-h{100, 75, 50, 25} experiments

Table 4 shows link prediction results on the NELL-995-h{100, 75, 50, 25} datasets for MuRE and MuRP at $d = 40$ and $d = 200$. At $d = 40$, MuRP consistently outperforms MuRE on all four datasets. As expected, the difference between model performances gets smaller as we increase the number of non-hierarchical relations and is the smallest on NELL-995-h25. For $d = 200$, MuRE starts to perform comparably (or even outperforms on some metrics) to MuRP on NELL-995-h50 and outperforms on NELL-995-h25. These results substantiate our claim on the advantage of embedding hierarchical data in hyperbolic space, particularly in a scenario where low embedding dimensionality is required.

Table 4: Link prediction results on the NELL-995-h{100, 75, 50, 25} datasets for $d = 40$ and $d = 200$.

| Dataset | Model | $d = 40$ | | | | $d = 200$ | | | |
|---|---|---|---|---|---|---|---|---|---|
| | | MRR | Hits@10 | Hits@3 | Hits@1 | MRR | Hits@10 | Hits@3 | Hits@1 |
| NELL-995-h100 | MuRE | .330 | .502 | .366 | .245 | .355 | .527 | .398 | .266 |
| | MuRP | **.344** | **.511** | **.383** | **.261** | **.360** | **.529** | **.401** | **.274** |
| NELL-995-h75 | MuRE | .330 | .497 | .368 | .246 | .356 | **.526** | .396 | .269 |
| | MuRP | **.345** | **.506** | **.382** | **.263** | **.359** | .524 | **.401** | **.275** |
| NELL-995-h50 | MuRE | .342 | .510 | .383 | .256 | **.372** | **.544** | **.415** | **.284** |
| | MuRP | **.356** | **.519** | **.399** | **.271** | .371 | .539 | **.415** | **.284** |
| NELL-995-h25 | MuRE | .337 | .489 | .374 | .259 | **.365** | **.515** | **.404** | **.287** |
| | MuRP | **.343** | **.494** | **.379** | **.266** | .359 | .507 | .397 | .282 |

Figure 5 emphasizes the difference in performance (MRR) of MuRP and MuRE (taken from Table 4). We can see that on the purely hierarchical NELL-995-h100 (43 hierarchical and 0 non-hierarchical relations), MuRP outperforms MuRE both at lower and higher dimensionality. On the other hand, on NELL-995-h25 which is mostly non-hierarchical (43 hierarchical and 129 non-hierarchical relations), MuRP is only slightly better than MuRE at $d = 40$, while MuRE outperforms at $d = 200$.

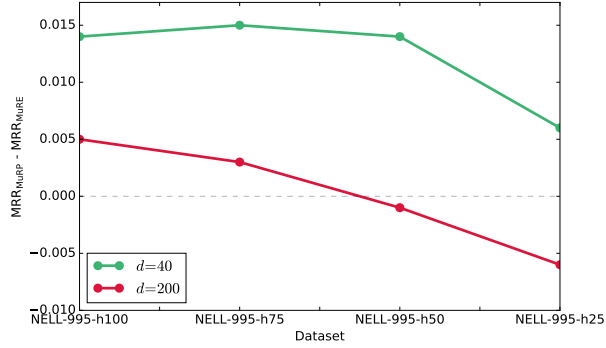

Figure 5: Difference in performance (MRR) of MuRP and MuRE on the NELL-995-h{100, 75, 50, 25} datasets for $d = 40$ and $d = 200$. The difference becomes smaller (turning negative for $d = 200$) as the number of non-hierarchical relations increases.

## D  Krackhardt hierarchy score

Let $\mathbf{R} \in \mathbb{R}^{n \times n}$ be the binary *reachability matrix* of a directed graph $\mathcal{G}$ with $n$ nodes, with $\mathbf{R}_{i,j} = 1$ if there exists a directed path from node $i$ to node $j$ and 0 otherwise. The Krackhardt hierarchy score of $\mathcal{G}$ [17] is defined as:

$$\text{Khs}_{\mathcal{G}} = \frac{\sum_{i=1}^{n} \sum_{j=1}^{n} \mathbb{1}(\mathbf{R}_{i,j} == 1 \wedge \mathbf{R}_{j,i} == 0)}{\sum_{i=1}^{n} \sum_{j=1}^{n} \mathbb{1}(\mathbf{R}_{i,j} == 1)}. \tag{9}$$

## E  Dimensionality reduction method

To project high-dimensional embeddings to 2 dimensions for visualization purposes, we use the following method to compute dimensions $x, y$ for projection $\mathbf{e}'_i$ of entity $\mathbf{e}_i$:

- $e_i^{x\prime} = \frac{\mathbf{e}_s}{\|\mathbf{e}_s\|} \mathbf{e}_i, i \in \{s, o_0, o_1, ..., o_N\}$, where $\mathbf{e}_s$ is the original high-dimensional subject entity embedding and $N$ is the number of object entity embeddings.
- $e_i^{y\prime} = \sqrt{\|\mathbf{e}_i\|^2 - \|e_i^{x\prime}\|^2}, i \in \{s, o_0, o_1, ..., o_N\}$.

This projects the reference subject entity embedding onto the $x$-axis ($e_s^{x\prime} = \|\mathbf{e}_s\|, e_s^{y\prime} = 0$) and all object entity embeddings are positioned relative to it, according to their $e_i^{x\prime}$ component aligned with the subject entity and their "remaining" component $e_i^{y\prime}$.

## 5.3 Impact

According to Google Scholar, the paper has received 89 citations as of November 2021. It was featured in various surveys (e.g. Ji et al., 2021; Saxena et al., 2020; Chami et al., 2020a) and works on hyperbolic representation learning (e.g. Shimizu et al., 2021; Wang et al., 2020; Zhang et al., 2021b).

MuRP has directly influenced several followup models. Chami et al. (2020b) improve upon the link prediction performance of MuRP by (i) replacing the diagonal matrix **R** with a rotation matrix; and (ii) learning relation-specific curvature. Given not all knowledge graph relations are hierarchical, Wang et al. (2021) and Zhang et al. (2021a) propose extensions of MuRP embedded in the mixed-curvature space which is a product of hyperbolic, Euclidean and spherical components. Han et al. (2020) generalise MuRP to temporal knowledge graphs. Yan et al. (2021) extend the MuRP score function to n-ary relational knowledge graphs.

## 5.4 Discussion

The presented paper demonstrated the utility of embedding hierarchical multi-relational graphs in hyperbolic rather than Euclidean space. However, being the first work to embed multi-relational data in hyperbolic space, there is still plenty of room for improvement, some of which has been addressed by follow-up work.

The Poincaré ball curvature is fixed for MuRP, despite different relations exhibiting different levels of hierarchy — and in the extreme, some may not be hierarchical at all — which suggests learnable, relation-specific curvature is needed. Chami et al. (2020b) show that learning relation-specific curvature leads to improved link prediction performance. Zhang et al. (2021a) and Wang et al. (2021) learn mixed-curvature embeddings as a product of hyperbolic, spherical and Euclidean components.

Chami et al. (2020b) further demonstrate the benefit of using a relation-specific rotation for the multiplicative component of the score function, compared to the simple stretch through the diagonal matrix **R** used by MuRP. This suggests further research into different types of multiplicative transformations is needed.

Lastly, both MuRP and MuRE are outperformed by TuckER (Balažević et al., 2019b) on FB15k-237 (Toutanova et al., 2015) — a predominantly non-hierarchical dataset with a large number of relations — primarily due to multi-task learning across relations. Future work should aim to incorporate the multi-task learning aspect into the hyperbolic score function.

# Chapter 6

# Interpreting Knowledge Graph Relation Representation from Word Embeddings

The main contribution of this chapter is the paper *Interpreting Knowledge Graph Relation Representation from Word Embeddings* which was published at the *International Conference on Learning Representations* in May 2021. We first outline the motivation for this work (Section 6.1), followed by the paper itself (Section 6.2), the impact that it has had so far (Section 6.3), and discussion (Section 6.4).

## 6.1  Introduction

Despite the large number of knowledge graph representation models and their gradual improvements in predictive performance on the link prediction task, relatively little is understood of the low-rank latent structure that underpins them. The main aim of the paper in the following section is to address this shortcoming. The outcomes of the presented analysis can be used to (i) aid and direct future knowledge graph representation model design, (ii) provide a basis for transforming logical rules between relations into constraints on the geometric interactions of relation representations; and (iii) offer insight for joint learning of knowledge graph representation and word embedding models using data from multiple sources (e.g. text and knowledge graphs).

The motivating premise for this work is that the same latent word features (e.g. meaning, tense, grammatical type) give rise to the patterns found in different data sources, i.e. determining which words co-occur in a text corpus and which words relate to which in a knowledge graph. We build on recent theoretical work that explains how semantic relationships of similarity, paraphrase and analogy are encoded in geometric interactions between PMI-based word embeddings (Allen and Hospedales, 2019; Allen et al., 2019) and draw a parallel between analogies as found in word embeddings (see Section 2.3.2) and relations between knowledge graph entities.

Analogies contain *common binary word relations*, which are similar to relations found in knowledge graphs. For certain types of analogies, a shared *semantic difference* between word pairs (e.g. (*man*, *king*), (*woman*, *queen*) and (*boy*, *prince*)) corresponds to a common *relation-specific vector offset* between the corresponding word embedding pairs (see Section 2.3.2). However, not all relations fit this semantic pattern and thus cannot be identified simply by a vector offset between embeddings, but we leverage this insight to identify more general geometric relationships that correspond to other semantic relation types, i.e. those of knowledge graphs. Such *relation conditions* — required for defining relation-specific mappings from subject word embeddings to all related object word embeddings (or vice versa) — provide a "blue-print" for representing relations in knowledge graph representation models.

Based on the relation conditions identified by analysing PMI-based word embeddings, we make several predictions relating to properties of relation representation and relative link prediction performance of leading knowledge graph representation models (introduced in Section 2.2.1). Our empirical findings accord with predictions based on these relation conditions, supporting the premise that there is *commonality in latent structure* learned by word embedding and knowledge graph representation models, despite the significant differences between their training data and methodology.

## 6.2 The Paper

### Author Contributions

The paper is co-authored by Carl Allen, myself and Timothy Hospedales, with Carl and myself being joint lead authors with equal contribution. The paper is an intersection of Carl Allen's PhD work on theoretically linking the semantics with the geometry of word representations and my PhD work on representation learning for knowledge graphs. We co-wrote the paper and had many joint discussions where we developed the semantic relation types, their relation conditions and the corresponding theoretical score functions. Carl Allen had the initial idea for this work and discussed all the experiments. I wrote the code and performed all the experiments. Timothy Hospedales offered suggestions and helped revise the final version.

# INTERPRETING KNOWLEDGE GRAPH RELATION REPRESENTATION FROM WORD EMBEDDINGS

**Carl Allen**[1]*, **Ivana Balažević**[1]* **& Timothy Hospedales**[1,2]
[1] University of Edinburgh, UK  [2] Samsung AI Centre, Cambridge, UK
{carl.allen, ivana.balazevic, t.hospedales}@ed.ac.uk

## ABSTRACT

Many models learn representations of knowledge graph data by exploiting its low-rank latent structure, encoding known relations between entities and enabling unknown facts to be inferred. To predict whether a relation holds between entities, embeddings are typically compared in the latent space following a relation-specific mapping. Whilst their predictive performance has steadily improved, how such models capture the underlying latent structure of semantic information remains unexplained. Building on recent theoretical understanding of word embeddings, we categorise knowledge graph relations into three types and for each derive explicit requirements of their representations. We show that empirical properties of relation representations and the relative performance of leading knowledge graph representation methods are justified by our analysis.

## 1 INTRODUCTION

Knowledge graphs are large repositories of binary relations between words (or entities) in the form of *(subject, relation, object)* triples. Many models for representing entities and relations have been developed, so that known facts can be recalled and previously unknown facts can be inferred, a task known as *link prediction*. Recent link prediction models (e.g. Bordes et al., 2013; Trouillon et al., 2016; Balažević et al., 2019b) learn entity representations, or *embeddings*, of far lower dimensionality than the number of entities, by capturing latent structure in the data. Relations are typically represented as a mapping from the embedding of a subject entity to those of related object entities. Although the performance of link prediction models has steadily improved for nearly a decade, relatively little is understood of the low-rank latent structure that underpins them, which we address in this work. The outcomes of our analysis can be used to aid and direct future knowledge graph model design.

We start by drawing a parallel between the entity embeddings of knowledge graphs and context-free word embeddings, e.g. as learned by Word2Vec (W2V) (Mikolov et al., 2013a) and GloVe (Pennington et al., 2014). Our motivating premise is that the same latent word features (e.g. meaning(s), tense, grammatical type) give rise to the patterns found in different data sources, i.e. manifesting in word co-occurrence statistics and determining which words relate to which. Different embedding approaches may capture such structure in different ways, but if it is fundamentally the same, an understanding gained from one embedding task (e.g. word embedding) may benefit another (e.g. knowledge graph representation). Furthermore, the relatively limited but accurate data used in knowledge graph representation differs materially from the highly abundant but statistically noisy text data used for word embeddings. As such, theoretically reconciling the two embedding methods may lead to unified and improved embeddings learned jointly from both data sources.

Recent work (Allen & Hospedales, 2019; Allen et al., 2019) theoretically explains how semantic properties are encoded in word embeddings that (approximately) factorise a matrix of *pointwise mutual information* (PMI) from word co-occurrence statistics, as known for W2V (Levy & Goldberg, 2014). *Semantic* relationships between words, specifically similarity, relatedness, paraphrase and analogy, are proven to manifest as linear *geometric* relationships between rows of the PMI matrix (subject to known error terms), of which word embeddings can be considered low-rank projections. This explains, for example, the observations that similar words have similar embeddings and that embeddings of analogous word pairs share a common "vector offset" (e.g. Mikolov et al., 2013b).

---

*Equal contribution

**Table 1:** Score functions of representative linear link prediction models. $\boldsymbol{R} \in \mathbb{R}^{d_e \times d_e}$ and $\boldsymbol{r} \in \mathbb{R}^{d_e}$ are the relation matrix and translation vector, $\mathsf{W} \in \mathbb{R}^{d_e \times d_r \times d_e}$ is the core tensor and $b_s, b_o \in \mathbb{R}$ are the entity biases.

| Model | | Linear Subcategory | Score Function |
|---|---|---|---|
| TransE | (Bordes et al., 2013) | additive | $-\|\boldsymbol{e}_s + \boldsymbol{r} - \boldsymbol{e}_o\|_2^2$ |
| DistMult | (Yang et al., 2015) | multiplicative (diagonal) | $\boldsymbol{e}_s^\top \boldsymbol{R} \boldsymbol{e}_o$ |
| TuckER | (Balažević et al., 2019b) | multiplicative | $\mathsf{W} \times_1 \boldsymbol{e}_s \times_2 \boldsymbol{r} \times_3 \boldsymbol{e}_o$ |
| MuRE | (Balažević et al., 2019a) | multiplicative (diagonal) + additive | $-\|\boldsymbol{R}\boldsymbol{e}_s + \boldsymbol{r} - \boldsymbol{e}_o\|_2^2 + b_s + b_o$ |

We extend this insight to identify geometric relationships between PMI-based word embeddings that correspond to other relations, i.e. those of knowledge graphs. Such *relation conditions* define relation-specific mappings between entity embeddings (i.e. *relation representations*) and so provide a "blue-print" for knowledge graph representation models. Analysing the relation representations of leading knowledge graph representation models, we find that various properties, including their relative link prediction performance, accord with predictions based on these relation conditions, supporting the premise that a *common latent structure* is learned by word and knowledge graph embedding models, despite the significant differences between their training data and methodology.

In summary, the key contributions of this work are:

- to use recent understanding of PMI-based word embeddings to derive geometric attributes of a relation representation for it to map subject word embeddings to all related object word embeddings (*relation conditions*), which partition relations into three *types* (§3);
- to show that both per-relation ranking as well as classification performance of leading link prediction models corresponds to the model satisfying the appropriate relation conditions, i.e. how closely its relation representations match the geometric form derived theoretically (§4.1); and
- to show that properties of knowledge graph representation models fit predictions based on relation conditions, e.g. the strength of a relation's *relatedness* aspect is reflected in the eigenvalues of its relation matrix (§4.2).

## 2 BACKGROUND

**Knowledge graph representation:** Recent knowledge graph models typically represent entities $e_s, e_o$ as vectors $\boldsymbol{e}_s, \boldsymbol{e}_o \in \mathbb{R}^{d_e}$, and relations as transformations in the latent space from subject to object entity embedding, where the dimension $d_e$ is far lower (e.g. 200) than the number of entities $n_e$ (e.g. $> 10^4$). Such models are distinguished by their *score function*, which defines (i) the form of the relation transformation, e.g. matrix multiplication and/or vector addition; and (ii) the measure of proximity between a transformed subject embedding and an object embedding, e.g. dot product or Euclidean distance. Score functions can be non-linear (e.g. Dettmers et al., 2018), or linear and sub-categorised as *additive*, *multiplicative* or both. We focus on linear models due to their simplicity and strong performance at link prediction (including state-of-the-art). Table 1 shows the score functions of competitive linear knowledge graph embedding models spanning the sub-categories: TransE (Bordes et al., 2013), DistMult (Yang et al., 2015), TuckER (Balažević et al., 2019b) and MuRE (Balažević et al., 2019a).

*Additive models* apply a relation-specific translation to a subject entity embedding and typically use Euclidean distance to evaluate proximity to object embeddings. A generic additive score function is given by $\phi(e_s, r, e_o) = -\|\boldsymbol{e}_s + \boldsymbol{r} - \boldsymbol{e}_o\|_2^2 + b_s + b_o$. A simple example is TransE, where $b_s = b_o = 0$.

*Multiplicative models* have the generic score function $\phi(e_s, r, e_o) = \boldsymbol{e}_s^\top \boldsymbol{R} \boldsymbol{e}_o$, i.e. a bilinear product of the entity embeddings and a relation-specific matrix $\boldsymbol{R}$. DistMult is a simple example with $\boldsymbol{R}$ diagonal and so cannot model asymmetric relations (Trouillon et al., 2016). In TuckER, each relation-specific $\boldsymbol{R} = \mathsf{W} \times_3 \boldsymbol{r}$ is a linear combination of $d_r$ "prototype" relation matrices in a core tensor $\mathsf{W} \in \mathbb{R}^{d_e \times d_r \times d_e}$ ($\times_n$ denoting tensor product along mode $n$), facilitating *multi-task learning* across relations.

Some models, e.g. MuRE, combine both multiplicative ($\boldsymbol{R}$) and additive ($\boldsymbol{r}$) components.

**Word embedding:** Algorithms such as Word2Vec (Mikolov et al., 2013a) and GloVe (Pennington et al., 2014) generate low-dimensional word embeddings that perform well on downstream tasks (Baroni et al., 2014). Such models predict the context words ($c_j$) observed around a target word ($w_i$) in a text corpus using shallow neural networks. Whilst recent language models (e.g. Devlin et al., 2018; Peters et al., 2018) achieve strong performance using *contextualised* word embeddings, we focus on "context-free" embeddings since knowledge graph entities have no obvious context and, importantly, they offer insight into embedding interpretability.

Levy & Goldberg (2014) show that, for a dictionary of $n_e$ unique words and embedding dimension $d_e \ll n_e$, W2V's loss function is minimised when its embeddings $\boldsymbol{w}_i, \boldsymbol{c}_j$ form matrices $\boldsymbol{W}, \boldsymbol{C} \in \mathbb{R}^{d_e \times n_e}$ that factorise a *pointwise mutual information* (PMI) matrix of word co-occurrence statistics $(\text{PMI}(w_i, c_j) = \log \frac{P(w_i, c_j)}{P(w_i)P(c_j)})$, subject to a shift term. This result relates W2V to earlier count-based embeddings and specifically PMI, which has a history in linguistic analysis (Turney & Pantel, 2010). From its loss function, GloVe can be seen to perform a related factorisation.

Recent work (Allen & Hospedales, 2019; Allen et al., 2019) shows how the semantic relationships of *similarity*, *relatedness*, *paraphrase* and *analogy* are encoded in PMI-based word embeddings by recognising such embeddings as low-rank projections of high dimensional rows of the PMI matrix, termed **PMI vectors**. Those semantic relationships are described in terms of *multiplicative* interactions between co-occurrence probabilities (subject to defined error terms), that correspond to *additive* interactions between (logarithmic) PMI statistics, and hence PMI vectors. Thus, under a sufficiently linear projection, those semantic relationships correspond to linear relationships between word embeddings. Note that although the relative geometry reflecting semantic relationships is preserved, the direct interpretability of dimensions, as in PMI vectors, is lost since the embedding matrices can be arbitrarily scaled/rotated if the other is inversely transformed. We state the relevant semantic relationships on which we build, denoting the set of unique dictionary words by $\mathcal{E}$:

- **Paraphrase**: word subsets $\mathcal{W}, \mathcal{W}^* \subseteq \mathcal{E}$ are said to *paraphrase* if they induce similar distributions over nearby words, i.e. $p(\mathcal{E}|\mathcal{W}) \approx p(\mathcal{E}|\mathcal{W}^*)$, e.g. $\{king\}$ paraphrases $\{man, royal\}$.
- **Analogy**: a common example of an *analogy* is "*woman* is to *queen* as *man* is to *king*" and can be defined as any set of word pairs $\{(w_i, w_i^*)\}_{i \in \mathcal{I}}$ for which it is semantically meaningful to say "$w_a$ is to $w_a^*$ as $w_b$ is to $w_b^*$" $\forall a, b \in \mathcal{I}$.

Where one word subset paraphrases another, the sums of their embeddings are shown to be equal (subject to the independence of words within each set), e.g. $\boldsymbol{w}_{king} \approx \boldsymbol{w}_{man} + \boldsymbol{w}_{royal}$. An interesting connection is established between the two semantic relationships: a set of word pairs $\mathcal{A} = \{(w_a, w_a^*), (w_b, w_b^*)\}$ is an analogy if $\{w_a, w_b^*\}$ paraphrases $\{w_a^*, w_b\}$, in which case the embeddings satisfy $\boldsymbol{w}_{a^*} - \boldsymbol{w}_a \approx \boldsymbol{w}_{b^*} - \boldsymbol{w}_b$ ("vector offset").

## 3 FROM ANALOGIES TO KNOWLEDGE GRAPH RELATIONS

Analogies from the field of word embeddings are our starting point for developing a theoretical basis for representing knowledge graph relations. The relevance of analogies stems from the observation that for an analogy to hold (see §2), its word pairs, e.g $\{(man, king), (woman, queen), (girl, princess)\}$, must be *related* in the same way, comparably to subject-object entity pairs under a common knowledge graph relation. Our aim is to develop the understanding of PMI-based word embeddings (henceforth *word embeddings*), to identify the mathematical properties necessary for a relation representation to map subject word embeddings to all related object word embeddings.

Considering the paraphrasing word sets $\{king\}$ and $\{man, royal\}$ corresponding to the word embedding relationship $\boldsymbol{w}_{king} \approx \boldsymbol{w}_{man} + \boldsymbol{w}_{royal}$ (§2), *royal* can be interpreted as the semantic difference between *man* and *king*, fitting intuitively with the relationship $\boldsymbol{w}_{royal} \approx \boldsymbol{w}_{king} - \boldsymbol{w}_{man}$. Fundamentally, this relationship holds because the difference between words that co-occur (i.e. occur more frequently than if independent) with *king* and those that co-occur with *man*, reflects those words that co-occur with *royal*. We refer to this difference in co-occurrence distribution as a "context shift", from *man* (subject) to *king* (object). Allen & Hospedales (2019) effectively show that where multiple word pairs share a common context shift, they form an analogy whose embeddings satisfy the vector offset relationship. This result seems obvious where the context shift mirrors an identifiable word, the embedding of which is approximated by the common vector offset, e.g. *queen* and *woman* are related by the same context shift, i.e. $\boldsymbol{w}_{queen} \approx \boldsymbol{w}_{woman} + \boldsymbol{w}_{royal}$, thus $\boldsymbol{w}_{queen} - \boldsymbol{w}_{woman} \approx \boldsymbol{w}_{king} - \boldsymbol{w}_{man}$. However, the same result holds, i.e. an analogy is formed with a common vector offset between embeddings, for an arbitrary (common) context shift that may reflect no particular word. Importantly, these context shift relations evidence a case in which it is known how a relation can be represented, i.e. by an additive vector (comparable to TransE) *if* entities are represented by word embeddings. More generally, this provides an interpretable foothold into relation representation.

Note that not all sets of word pairs considered analogies exhibit a clear context shift relation, e.g. in the analogy $\{(car, engine), (bus, seats)\}$, the difference between words co-occurring with *engine* and *car* is not expected to reflect the corresponding difference between *bus* and *seats*. This illustrates how analogies are a loosely defined concept, e.g. their implicit relation may be semantic or syntactic, with several sub-categories of each (e.g. see Gladkova et al. (2016)). The same is readily observed for the relations of knowledge graphs. This likely explains the observed variability in "solving" analogies by use of vector offset (e.g. Köper et al., 2015; Karpinska et al., 2018; Gladkova et al., 2016) and suggests that further consideration is required to represent relations (or solve analogies) in general.

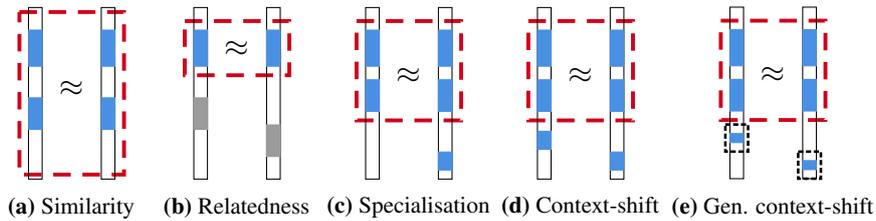

**Figure 1:** Relationships between PMI vectors (black rectangles) of subject/object words for different relation *types*. PMI vectors capture co-occurrence with every dictionary word: strong associations (PMI > 0) are shaded (blue define the relation, grey are random other associations); red dash = *relatedness*; black dash = *context sets*.

We have seen that the existence of a context shift relation between a subject and object word implies a (relation-specific) geometric relationship between word embeddings, thus the latter provides a *necessary condition for the relation to hold*. We refer to this as a "relation condition" and aim to identify relation conditions for other classes of relation. Once identified, relation conditions define a mapping from subject embeddings to all related object embeddings, by which related entities might be identified with a proximity measure (e.g. Euclidean distance or dot product). This is the precise aim of a knowledge graph representation model, but loss functions are typically developed heuristically. Given the existence of many representation models, we can verify identified relation conditions by contrasting the per-relation performance of various models with the extent to which their loss function reflects the appropriate relation conditions. Note that since relation conditions are necessary rather than sufficient, they do not guarantee a relation holds, i.e. false positives may arise.

Whilst we seek to establish relation conditions based on PMI word embeddings, the data used to train knowledge graph embeddings differs significantly to the text data used by word embeddings, and the relevance of conditions ultimately based on PMI statistics may seem questionable. However, where a knowledge graph representation model implements relation conditions and measures proximity between embeddings, the parameters of word embeddings necessarily provide *a potential* solution that minimises the loss function. Many equivalent solutions may exist due to symmetry as typical for neural network architectures. We now define relation types and identify their relation conditions (underlined); we then consider the completeness of this categorisation.

• **Similarity:** Semantically similar words induce similar distributions over the words they co-occur with. Thus their PMI vectors and <u>word embeddings are similar</u> (Fig 1a).

• **Relatedness:** The relatedness of two words can be considered in terms of the words $\mathcal{S} \subseteq \mathcal{E}$ with which both co-occur similarly. $\mathcal{S}$ defines the *nature* of relatedness, e.g. *milk* and *cheese* are related by $\mathcal{S} = \{dairy, breakfast, ...\}$; and $|\mathcal{S}|$ reflects the *strength* of relatedness. Since PMI vector components corresponding to $\mathcal{S}$ are similar (Fig 1b), embeddings of $\mathcal{S}$-*related* words <u>have similar components in the subspace</u> $\mathbb{V}_{\mathcal{S}}$ that spans the projected PMI vector dimensions corresponding to $\mathcal{S}$. The rank of $\mathbb{V}_{\mathcal{S}}$ is thus anticipated to reflect relatedness strength. Relatedness can be seen as a weaker and more variable generalisation of similarity, its limiting case where $\mathcal{S} = \mathcal{E}$, hence rank($\mathbb{V}_{\mathcal{S}}$) $= d_e$.

• **Context-shift:** As discussed above, words related by a common difference between their distributions of co-occurring words, defined as *context-shifts*, share a <u>common vector offset between word embeddings</u>. Context might be considered *added* (e.g. *man* to *king*), termed a **specialisation** (Fig 1c), *subtracted* (e.g. *king* to *man*) or both (Fig 1d). These relations are 1-to-1 (subject to synonyms) and include an aspect of *relatedness* due to the word associations in common. Note that, specialisations include hyponyms/hypernyms and context shifts include meronyms.

• **Generalised context-shift:** Context-shift relations generalise to 1-to-many, many-to-1 and many-to-many relations where the added/subtracted context may be from a (relation-specific) *context set* (Fig 1e), e.g. *any* city or *anything* bigger. The potential scope and size of context sets adds variability to these relations. The limiting case in which the context set is "small" reduces to a 1-to-1 context-shift (above) and the <u>embedding difference is a known vector offset</u>. In the limiting case of a "large" context set, the added/subtracted context is essentially unrestricted such that only the relatedness aspect of the relation, and thus a <u>common subspace component of embeddings</u>, is fixed.

**Categorisation completeness:** Taking intuition from Fig 1 and considering PMI vectors as *sets of word features*, these relation types can be interpreted as set operations: similarity as set equality; relatedness as subset equality; and context-shift as a relation-specific set difference. Since for any relation each feature must either remain unchanged (relatedness), change (context shift) or else be irrelevant, we conjecture that the above relation types give a complete partition of semantic relations.

**Table 2:** Categorisation of WN18RR relations.

| Type | Relation | Examples *(subject entity, object entity)* |
|------|----------|--------------------------------------------|
| R | verb_group<br>derivationally_related_form<br>also_see | *(trim_down_VB_1, cut_VB_35), (hatch_VB_1, incubate_VB_2)*<br>*(lodge_VB_4, accommodation_NN_4), (question_NN_1, inquire_VB_1)*<br>*(clean_JJ_1, tidy_JJ_1), (ram_VB_2, screw_VB_3)* |
| S | hypernym<br>instance_hypernym | *(land_reform_NN_1, reform_NN_1), (prickle-weed_NN_1, herbaceous_plant_NN_1)*<br>*(yellowstone_river_NN_1, river_NN_1), (leipzig_NN_1, urban_center_NN_1)* |
| C | member_of_domain_usage<br>member_of_domain_region<br>member_meronym<br>has_part<br>synset_domain_topic_of | *(colloquialism_NN_1, figure_VB_5), (plural_form_NN_1, authority_NN_2)*<br>*(rome_NN_1, gladiator_NN_1), (usa_NN_1, multiple_voting_NN_1)*<br>*(south_NN_2, sunshine_state_NN_1), (genus_carya_NN_1, pecan_tree_NN_1)*<br>*(aircraft_NN_1, cabin_NN_3), (morocco_NN_1, atlas_mountains_NN_1)*<br>*(quark_NN_1, physics_NN_1), (harmonize_VB_3, music_NN_4)* |

### 3.1 CATEGORISING REAL KNOWLEDGE GRAPH RELATIONS

Analysing the relations of popular knowledge graph datasets, we observe that they indeed imply (i) a relatedness aspect reflecting a common theme (e.g. both entities are animals or geographic terms); and (ii) contextual themes specific to the subject and/or object entities. Further, relations fall under a hierarchy of three *relation types*: highly related (**R**); generalised specialisation (**S**); and generalised context-shift (**C**). As above, "generalised" indicates that context differences are not restricted to be 1-1. From Fig 1, it can be seen that type R relations are a special case of S, which are a special case of C. Thus type C encompasses all considered relations. Whilst there are many ways to classify relations, e.g. by hierarchy, transitivity, the proposed relation conditions delineate relations by the required mathematical form (and complexity) of their representation. Table 2 shows a categorisation of the relations of the WN18RR dataset (Dettmers et al., 2018) comprising 11 relations and 40,943 entities.[1] An explanation for the category assignment is in Appx. A. Analysing the commonly used FB15k-237 dataset (Toutanova et al., 2015) reveals relations to be almost exclusively of type C, precluding a contrast of performance per relation type and hence that dataset is omitted from our analysis. Instead, we categorise a random subsample of 12 relations from the NELL-995 dataset (Xiong et al., 2017), containing 75,492 entities and 200 relations (see Tables 8 and 9 in Appx. B).

### 3.2 RELATIONS AS MAPPINGS BETWEEN EMBEDDINGS

Given the relation conditions of a relation type, we now consider mappings that satisfy them and thereby loss functions able to identify relations of each type, evaluating proximity between mapped entity embeddings by dot product or Euclidean distance. We then contrast our theoretically derived loss functions, specific to a relation type, with those of several knowledge graph models (Table 1) to predict identifiable properties and the relative performance of different knowledge graph models for each relation type.

**R:** Identifying $\mathcal{S}$-relatedness requires testing both entity embeddings $e_s, e_o$ for a common subspace component $\mathbb{V}_{\mathcal{S}}$, which can be achieved by projecting both embeddings onto $\mathbb{V}_{\mathcal{S}}$ and comparing their images. Projection requires multiplication by a matrix $P_r \in \mathbb{R}^{d \times d}$ and cannot be achieved additively, except in the trivial limiting case of similarity ($P_r = I$) when $r \approx 0$ can be added.
Comparison by dot product gives $(P_r e_s)^\top (P_r e_o) = e_s^\top P_r^\top P_r e_o = e_s^\top M_r e_o$ (for relation-specific symmetric $M_r = P_r^\top P_r$). Euclidean distance gives $\|P_r e_s - P_r e_o\|^2 = (e_s - e_o)^\top M_r (e_s - e_o) = \|P_r e_s\|^2 - 2 e_s^\top M_r e_o + \|P_r e_o\|^2$.

**S/C:** The relation conditions require testing for both $\mathcal{S}$-relatedness and relation-specific entity component(s) ($v_r^s, v_r^o$). This is achieved by (i) multiplying both entity embeddings by a relation-specific projection matrix $P_r$ that projects onto the subspace that spans the low-rank projection of dimensions corresponding to $\mathcal{S}$, $v_r^s$ and $v_r^o$ (i.e. testing for $\mathcal{S}$-relatedness while preserving relation-specific entity components); and (ii) adding a relation-specific vector $r = v_r^o - v_r^s$ to the transformed subject entity embeddings.

---

[1] We omit the relation "similar_to" since its instances have no discernible structure, and only 3 occur in the test set, all of which are the inverse of a training example and trivial to predict.

Comparing the transformed entity embeddings by dot product equates to $(\boldsymbol{P}_r\boldsymbol{e}_s + \boldsymbol{r})^\top \boldsymbol{P}_r\boldsymbol{e}_o$; and by Euclidean distance to $\|\boldsymbol{P}_r\boldsymbol{e}_s + \boldsymbol{r} - \boldsymbol{P}_r\boldsymbol{e}_o\|^2 = \|\boldsymbol{P}_r\boldsymbol{e}_s + \boldsymbol{r}\|^2 - 2(\boldsymbol{P}_r\boldsymbol{e}_s + \boldsymbol{r})^\top \boldsymbol{P}_r\boldsymbol{e}_o + \|\boldsymbol{P}_r\boldsymbol{e}_o\|^2$ (*cf* MuRE: $\|\boldsymbol{R}\boldsymbol{e}_s + \boldsymbol{r} - \boldsymbol{e}_o\|^2$).

Contrasting these theoretically derived loss functions with those of knowledge graph models (Table 1), we make the following predictions:

**P1:** The ability to learn the representation of a relation is expected to reflect:
    (a) the complexity of its type (R<S<C) independently of model choice; and
    (b) whether relation conditions (e.g. additive/multiplicative interactions) are met by the model.
**P2:** Knowledge graph relation representations reflect the following type-specific properties:
    (a) relation matrices for relatedness (type R) relations are highly symmetric;
    (b) offset vectors for relatedness relations have low norm; and
    (c) as a proxy to the rank of $\mathbb{V}_{\mathcal{S}}$, the eigenvalues of a relation matrix reflect relatedness strength.

To elaborate, our core prediction P1(b) anticipates that: (i) additive-only models (e.g. TransE) are not suited to identifying the relatedness aspect of relations, except in limiting cases of similarity, requiring a zero vector); (ii) multiplicative-only models (e.g. DistMult) should perform well on type R relations, but are not suited to identifying entity-specific features of type S/C (an asymmetric relation matrix in TuckER may help compensate); and (iii) the loss function of MuRE closely resembles that derived for type C relations, which generalise all others, and is thus expected to perform best overall.

## 4 Evidence linking knowledge graph and word embeddings

We test whether the predictions P1 and P2, made on the basis of word embeddings, apply to knowledge graph relations by analysing the performance and properties of competitive knowledge graph models. We compare TransE, DistMult, TuckER and MuRE, which entail different forms of relation representation, on all WN18RR relations and a similar number of NELL-995 relations (spanning all relation types). All models have a comparable number of free parameters.

Since for TransE, the logistic sigmoid cannot be applied to the score function to give a probabilistic interpretation comparable to other models, for fair comparison we include MuRE$_\boldsymbol{I}$, a constrained variant of MuRE with $\boldsymbol{R}_s = \boldsymbol{R}_o = \boldsymbol{I}$, as a proxy to TransE. Implementation details are included in Appx. D. For evaluation, we generate $2n_e$ *evaluation triples* for each test triple ($n_e = |\mathcal{E}|$ denoting the number of entities) by fixing the subject entity $\boldsymbol{e}_s$ and relation $r$ and replacing the object entity $\boldsymbol{e}_o$ with each entity in turn and then replacing $\boldsymbol{e}_o$ and $r$ fixed and varying $\boldsymbol{e}_s$. Each model's scores for the evaluation triples are ranked to give the standard metric Hits@10 (Bordes et al., 2013), i.e. the fraction of times a true triple appears in the top 10 ranked evaluation triples.

### 4.1 P1: Justifying the relative performance of knowledge graph models

**Ranking performance:** Tables 3 and 4 report Hits@10 for each relation and include the relation type as well as known confounding influences: percentage of relation instances in the training and test sets (approximately equal), the actual number of instances in the test set (causing some results to be highly granular), Krackhardt hierarchy score (see Appx. E) (Krackhardt, 2014; Balaževič et al., 2019a) and maximum and average shortest path between any two related nodes. A further confounding effect is dependence between relations: Lacroix et al. (2018) and Balaževič et al. (2019b) independently show that constraining the rank of relation representations is beneficial for datasets with many relations due to *multi-task learning*, particularly when the number of instances per relation is low. This is expected to benefit TuckER on the NELL-995 dataset (200 relations).

As predicted by P1(a), all models tend to perform best at type R relations, with a clear performance gap to other relation types. Also, performance on type S relations appears higher in general than type C. In accordance with P1(b), additive-only models (TransE, MuRE$_\boldsymbol{I}$) perform worst on average since all relation types involve (multiplicative) relatedness. Best performance is achieved on type R relations, which can be represented by a small/zero additive vector. Multiplicative-only DistMult performs well, sometimes best, on type R relations, fitting expectation as it can represent those relations and has no inessential parameters, e.g. that may overfit to noise, which may explain instances where MuRE performs slightly worse. As expected, MuRE, performs best overall (particularly on WN18RR), and most strongly on S and C type relations, predicted to require both multiplicative and additive components. Comparable performance of TuckER on NELL-995 may be explained by its multi-task learning ability.

**Table 3:** Hits@10 per relation on WN18RR.

| Relation Name | Type | % | # | Khs | Max/Avg Path | | TransE | MuRE$_I$ | DistMult | TuckER | MuRE |
|---|---|---|---|---|---|---|---|---|---|---|---|
| verb_group | R | 1% | 39 | 0.00 | - | - | 0.87 | 0.95 | **0.97** | **0.97** | **0.97** |
| derivationally_related_form | R | 34% | 1074 | 0.04 | - | - | 0.93 | 0.96 | 0.96 | 0.96 | **0.97** |
| also_see | R | 2% | 56 | 0.24 | 44 | 15.2 | 0.59 | **0.73** | 0.67 | 0.72 | **0.73** |
| instance_hypernym | S | 4% | 122 | 1.00 | 3 | 1.0 | 0.22 | 0.52 | 0.47 | 0.53 | **0.54** |
| synset_domain_topic_of | C | 4% | 114 | 0.99 | 3 | 1.1 | 0.19 | 0.43 | 0.42 | 0.45 | **0.53** |
| member_of_domain_usage | C | 1% | 24 | 1.00 | 2 | 1.0 | 0.42 | 0.42 | 0.48 | 0.38 | **0.50** |
| member_of_domain_region | C | 1% | 26 | 1.00 | 2 | 1.0 | 0.35 | 0.40 | 0.40 | 0.35 | **0.46** |
| member_meronym | C | 8% | 253 | 1.00 | 10 | 3.9 | 0.04 | 0.38 | 0.30 | **0.39** | **0.39** |
| has_part | C | 6% | 172 | 1.00 | 13 | 2.2 | 0.04 | 0.31 | 0.28 | 0.29 | **0.35** |
| hypernym | S | 40% | 1251 | 0.99 | 18 | 4.5 | 0.02 | 0.20 | 0.19 | 0.20 | **0.28** |
| all | | 100% | 3134 | | | | 0.38 | 0.52 | 0.51 | 0.53 | **0.57** |

**Table 4:** Hits@10 per relation on NELL-995.

| Relation Name | Type | % | # | Khs | Max/Avg Path | | TransE | MuRE$_I$ | DistMult | TuckER | MuRE |
|---|---|---|---|---|---|---|---|---|---|---|---|
| teamplaysagainstteam | R | 2% | 243 | 0.11 | 10 | 3.5 | 0.76 | 0.84 | **0.90** | 0.89 | 0.89 |
| clothingtogowithclothing | R | 1% | 132 | 0.17 | 5 | 2.6 | 0.72 | 0.80 | **0.88** | 0.85 | 0.84 |
| professionistypeofprofession | S | 1% | 143 | 0.38 | 7 | 2.5 | 0.37 | 0.55 | 0.62 | 0.65 | **0.66** |
| animalistypeofanimal | S | 1% | 103 | 0.68 | 9 | 3.1 | 0.50 | 0.56 | 0.64 | **0.68** | 0.65 |
| athleteplayssport | C | 1% | 113 | 1.00 | 1 | 1.0 | 0.54 | 0.58 | 0.58 | 0.60 | **0.64** |
| chemicalistypeofchemical | S | 1% | 115 | 0.53 | 6 | 3.0 | 0.23 | 0.43 | 0.49 | 0.51 | **0.60** |
| itemfoundinroom | C | 2% | 162 | 1.00 | 1 | 1.0 | 0.39 | 0.57 | 0.53 | 0.56 | **0.59** |
| agentcollaborateswithagent | R | 1% | 119 | 0.51 | 14 | 4.7 | 0.44 | 0.58 | **0.64** | 0.61 | 0.58 |
| bodypartcontainsbodypart | C | 1% | 103 | 0.60 | 7 | 3.2 | 0.30 | 0.38 | 0.54 | **0.58** | 0.55 |
| atdate | C | 10% | 967 | 0.99 | 4 | 1.1 | 0.41 | 0.50 | 0.48 | 0.48 | **0.52** |
| locationlocatedwithinlocation | C | 2% | 157 | 1.00 | 6 | 1.9 | 0.35 | 0.37 | 0.46 | **0.48** | **0.48** |
| atlocation | C | 1% | 294 | 0.99 | 6 | 1.4 | 0.22 | 0.33 | 0.39 | 0.43 | **0.44** |
| all | | 100% | 20000 | | | | 0.36 | 0.48 | 0.51 | **0.52** | **0.52** |

Other anomalous results also closely align with confounding factors. For example, all models perform poorly on the *hypernym* relation, despite it having a relative abundance of training data (40% of all instances), which may be explained by its *hierarchical* nature (Khs ≈ 1 and long paths). The same may explain the reduced performance on relations *also_see* and *agentcollaborateswithagent*. As found previously (Balažević et al., 2019a), none of the models considered are well suited to modelling hierarchical structures. We also note that the percentage of training instances of a relation is not a dominant factor on performance, as would be expected if all relations could be equally represented.

**Classification performance:** We further evaluate whether P1 holds when comparing knowledge graph models by classification accuracy on WN18RR. Independent predictions of whether a given triple is true or false are not commonly evaluated, instead metrics such as mean reciprocal rank and Hits@$k$ are reported that compare the prediction of a test triple against all evaluation triples. Not only is this computationally costly, the evaluation is flawed if an entity is related to $l > k$ others ($k$ is often 1 or 3). A correct prediction validly falling within the top $l$ but not the top $k$ would appear incorrect under the metric. Some recent works also note the importance of standalone predictions (Speranskaya et al., 2020; Pezeshkpour et al., 2020).

Since for each relation there are $n_e^2$ possible entity-entity relationships, we sub-sample by computing predictions only for those $(e_s, r, e_o)$ triples for which the $e_s, r$ pairs appear in the test set. We split positive predictions ($\sigma(\phi(e_s, r, e_o)) > 0.5$) between (i) *known truths* – training or test/validation instances; and (ii) *other*, the truth of which is not known. We then compute per-relation accuracy over the true training instances ("train") and true test/validation instances ("test"); and the average number of "other" triples predicted true per $e_s, r$ pair. Table 5 shows results for MuRE$_I$, DistMult, TuckER and MuRE. All models achieve near perfect training accuracy. The additive-multiplicative MuRE gives best test set performance, followed (surprisingly) closely by MuRE$_I$, with multiplicative models (DistMult and TuckER) performing poorly on all but type R relations in line with P1(b), with near-zero performance on most type S/C relations. Since the ground truth labels for "other" triples predicted to be true are not in the dataset, we analyse a sample of "other" true predictions for one relation of each type (see Appx. G). From this, we estimate that TuckER is relatively accurate but pessimistic (~0.3 correct of the 0.5 predictions≈60%), MuRE$_I$ is optimistic but inaccurate (~2.3 of 7.5≈31%), whereas MuRE is both optimistic and accurate (~1.1 of 1.5≈73%).

**Summary:** Our analysis identifies the best performing model per relation type as predicted by P1(b): multiplicative-only DistMult for type R, additive-multiplicative MuRE for types S/C; providing a basis for *dataset-dependent model selection*. The per-relation insight into where models perform poorly, e.g. hierarchical or type C relations, can be used to aid and direct future model design. Analysis of the classification performance: (i) shows that MuRE is the most reliable fact prediction model; and (ii) emphasises the poorer ability of multiplicative-only models to represent S/C relations.

**Table 5:** Per relation prediction accuracy for MuRE$_I$ (M$_I$), (D)istMult, (T)uckER and (M)uRE (WN18RR).

| Relation Name | Type | #$_{train}$ | #$_{test}$ | Accuracy (train) | | | | Accuracy (test) | | | | # Other "True" | | | |
|---|---|---|---|---|---|---|---|---|---|---|---|---|---|---|---|
| | | | | M$_I$ | D | T | M | M$_I$ | D | T | M | M$_I$ | D | T | M |
| verb_group | R | 15 | 39 | 1.00 | 1.00 | 1.00 | 1.00 | 0.97 | 0.97 | 0.97 | 0.97 | 8.3 | 1.7 | 0.9 | 2.7 |
| derivationally_related_form | R | 1714 | 1127 | 1.00 | 1.00 | 1.00 | 1.00 | 0.96 | 0.94 | 0.95 | 0.95 | 8.8 | 0.5 | 0.6 | 1.7 |
| also_see | R | 95 | 61 | 1.00 | 1.00 | 1.00 | 1.00 | 0.64 | 0.64 | 0.61 | 0.59 | 7.9 | 1.6 | 0.9 | 1.9 |
| instance_hypernym | S | 52 | 122 | 1.00 | 1.00 | 1.00 | 1.00 | 0.32 | 0.32 | 0.23 | 0.43 | 1.3 | 0.4 | 0.3 | 0.9 |
| member_of_domain_usage | C | 545 | 43 | 0.98 | 1.00 | 1.00 | 1.00 | 0.02 | 0.00 | 0.02 | 0.00 | 1.5 | 0.6 | 0.0 | 0.3 |
| member_of_domain_region | C | 543 | 42 | 0.88 | 0.89 | 1.00 | 0.93 | 0.02 | 0.02 | 0.00 | 0.02 | 1.0 | 0.4 | 0.8 | 0.7 |
| synset_domain_topic_of | C | 13 | 115 | 1.00 | 1.00 | 1.00 | 1.00 | 0.42 | 0.10 | 0.14 | 0.47 | 0.7 | 0.6 | 0.1 | 0.2 |
| member_meronym | C | 1402 | 307 | 1.00 | 1.00 | 1.00 | 1.00 | 0.22 | 0.02 | 0.01 | 0.22 | 7.9 | 3.4 | 1.5 | 5.6 |
| has_part | C | 848 | 196 | 1.00 | 1.00 | 1.00 | 1.00 | 0.24 | 0.05 | 0.09 | 0.22 | 7.1 | 2.4 | 1.3 | 3.9 |
| hypernym | S | 57 | 1254 | 1.00 | 1.00 | 1.00 | 1.00 | 0.15 | 0.02 | 0.02 | 0.22 | 3.7 | 1.2 | 0.0 | 1.7 |
| all | | 5284 | 3306 | 0.99 | 0.99 | 1.00 | 0.99 | 0.47 | 0.37 | 0.37 | 0.50 | 5.9 | 1.2 | 0.5 | 2.1 |

### 4.2 P2: Properties of relation representation

**P2(a)-(b):** Table 6 shows the symmetry score ($\in [-1, 1]$ indicating perfect anti-symmetry to symmetry; see Appx. F) for the relation matrix of TuckER and the norm of relation vectors of TransE, MuRE$_I$ and MuRE on the WN18RR dataset. As expected, type R relations have materially higher symmetry than both other relation types, fitting the prediction of how TuckER compensates for having no additive component. All additive models learn relation vectors of a noticeably lower norm for type R relations, which in the limiting case (similarity) require no additive component, than for types S or C.

**P2(c):** Fig 2 shows eigenvalue magnitudes (scaled relative to the largest and ordered) of relation-specific matrices $R$ of MuRE, labelled by relation type, as predicted to reflect the strength of a relation's *relatedness* aspect. As expected, values are highest for type R relations. For relation types S and C the profiles are more varied, supporting the understanding that relatedness of such relations is highly variable, both in its nature (components of $\mathcal{S}$) and strength (cardinality of $\mathcal{S}$).

**Table 6:** Relation matrix symmetry score [-1.1] for TuckER; and relation vector norm for TransE, MuRE$_I$ and MuRE (WN18RR).

| Relation | Type | Symmetry Score TuckER | Vector Norm TransE | MuRE$_I$ | MuRE |
|---|---|---|---|---|---|
| verb_group | R | 0.52 | 5.65 | 0.76 | 0.89 |
| derivationally_related_form | R | 0.54 | 2.98 | 0.45 | 0.69 |
| also_see | R | 0.50 | 7.20 | 0.97 | 0.97 |
| instance_hypernym | S | 0.13 | 18.26 | 2.98 | 1.88 |
| member_of_domain_usage | C | 0.10 | 11.24 | 3.18 | 1.88 |
| member_of_domain_region | C | 0.06 | 12.52 | 3.07 | 2.11 |
| synset_domain_topic_of | C | 0.12 | 23.29 | 2.65 | 1.52 |
| member_meronym | C | 0.12 | 4.97 | 1.91 | 1.97 |
| has_part | C | 0.13 | 6.44 | 1.69 | 1.25 |
| hypernym | S | 0.04 | 9.64 | 1.53 | 1.03 |

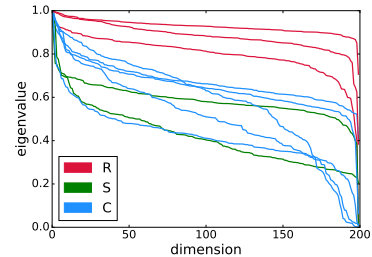

**Figure 2:** Eigenvalue magnitudes of relation-specific matrices $R$ for MuRE by relation type (WN18RR).

## 5 Conclusion

Many low-rank knowledge graph representation models have been developed, yet little is known of the latent structure they learn. We build on recent understanding of PMI-based word embeddings to theoretically establish a set of geometric properties of relation representations (relation conditions) required to map PMI-based word embeddings of subject entities to related object entities under knowledge graph relations. These conditions partition relations into three types and provide a basis to consider the loss functions of existing knowledge graph models. Models that satisfy the relation conditions of a particular type have a known set of model parameters that minimise the loss function, i.e. the parameters of PMI embeddings, together with potentially many equivalent solutions. We show that the better a model's architecture satisfies a relation's conditions, the better its performance at link prediction, evaluated under both rank-based metrics and accuracy. Overall, we generalise recent theoretical understanding of how particular semantic relations, e.g. similarity and analogy, are encoded between PMI-based word embeddings to the general relations of knowledge graphs. In doing so, we provide evidence in support of our initial premise: that common latent structure is exploited by both PMI-based word embeddings (e.g. W2V) and knowledge graph representation.

ACKNOWLEDGEMENTS

Carl Allen and Ivana Balažević were supported by the Centre for Doctoral Training in Data Science, funded by EPSRC (grant EP/L016427/1) and the University of Edinburgh.

REFERENCES

Carl Allen and Timothy Hospedales. Analogies Explained: Towards Understanding Word Embeddings. In *International Conference on Machine Learning*, 2019.

Carl Allen, Ivana Balažević, and Timothy Hospedales. What the Vec? Towards Probabilistically Grounded Embeddings. In *Advances in Neural Information Processing Systems*, 2019.

Ivana Balažević, Carl Allen, and Timothy M Hospedales. Multi-relational Poincaré Graph Embeddings. In *Advances in Neural Information Processing Systems*, 2019a.

Ivana Balažević, Carl Allen, and Timothy M Hospedales. TuckER: Tensor Factorization for Knowledge Graph Completion. In *Empirical Methods in Natural Language Processing*, 2019b.

Marco Baroni, Georgiana Dinu, and Germán Kruszewski. Don't Count, Predict! A Systematic Comparison of Context-Counting vs. Context-Predicting Semantic Vectors. In *Association for Computational Linguistics*, 2014.

Antoine Bordes, Nicolas Usunier, Alberto Garcia-Duran, Jason Weston, and Oksana Yakhnenko. Translating Embeddings for Modeling Multi-relational Data. In *Advances in Neural Information Processing Systems*, 2013.

Tim Dettmers, Pasquale Minervini, Pontus Stenetorp, and Sebastian Riedel. Convolutional 2D Knowledge Graph Embeddings. In *Association for the Advancement of Artificial Intelligence*, 2018.

Jacob Devlin, Ming-Wei Chang, Kenton Lee, and Kristina Toutanova. BERT: Pre-training of Deep Bidirectional Transformers for Language Understanding. In *North American Chapter of the Association for Computational Linguistics*, 2018.

Anna Gladkova, Aleksandr Drozd, and Satoshi Matsuoka. Analogy-based Detection of Morphological and Semantic Relations With Word Embeddings: What Works and What Doesn't. In *NAACL Student Research Workshop*, 2016.

Lawrence J Hubert and Frank B Baker. Evaluating the Symmetry of a Proximity Matrix. *Quality & Quantity*, 13(1):77–84, 1979.

Marzena Karpinska, Bofang Li, Anna Rogers, and Aleksandr Drozd. Subcharacter Information in Japanese Embeddings: When Is It Worth It? In *Workshop on the Relevance of Linguistic Structure in Neural Architectures for NLP*, 2018.

Diederik P Kingma and Jimmy Ba. Adam: A Method for Stochastic Optimization. In *International Conference on Learning Representations*, 2015.

Maximilian Köper, Christian Scheible, and Sabine Schulte im Walde. Multilingual reliability and "semantic" structure of continuous word spaces. In *International Conference on Computational Semantics*, 2015.

David Krackhardt. Graph Theoretical Dimensions of Informal Organizations. In *Computational Organization Theory*. Psychology Press, 2014.

Timothée Lacroix, Nicolas Usunier, and Guillaume Obozinski. Canonical Tensor Decomposition for Knowledge Base Completion. In *International Conference on Machine Learning*, 2018.

Omer Levy and Yoav Goldberg. Neural Word embedding as Implicit Matrix Factorization. In *Advances in Neural Information Processing Systems*, 2014.

Tomas Mikolov, Ilya Sutskever, Kai Chen, Greg S Corrado, and Jeff Dean. Distributed Representations of Words and Phrases and their Compositionality. In *Advances in Neural Information Processing Systems*, 2013a.

Tomas Mikolov, Wen-tau Yih, and Geoffrey Zweig. Linguistic regularities in continuous space word representations. In *North American Chapter of the Association for Computational Linguistics*, 2013b.

Jeffrey Pennington, Richard Socher, and Christopher Manning. Glove: Global Vectors for Word Representation. In *Empirical Methods in Natural Language Processing*, 2014.

Matthew E Peters, Mark Neumann, Mohit Iyyer, Matt Gardner, Christopher Clark, Kenton Lee, and Luke Zettlemoyer. Deep Contextualized Word Representations. In *North American Chapter of the Association for Computational Linguistics*, 2018.

Pouya Pezeshkpour, Yifan Tian, and Sameer Singh. Revisiting Evaluation of Knowledge Base Completion Models. In *Automated Knowledge Base Construction*, 2020.

Marina Speranskaya, Martin Schmitt, and Benjamin Roth. Ranking vs. Classifying: Measuring Knowledge Base Completion Quality. In *Automated Knowledge Base Construction*, 2020.

Kristina Toutanova, Danqi Chen, Patrick Pantel, Hoifung Poon, Pallavi Choudhury, and Michael Gamon. Representing Text for Joint Embedding of Text and Knowledge Bases. In *Empirical Methods in Natural Language Processing*, 2015.

Théo Trouillon, Johannes Welbl, Sebastian Riedel, Éric Gaussier, and Guillaume Bouchard. Complex Embeddings for Simple Link Prediction. In *International Conference on Machine Learning*, 2016.

Peter D Turney and Patrick Pantel. From Frequency to Meaning: Vector Space Models of Semantics. *Journal of Artificial Intelligence Research*, 37:141–188, 2010.

Wenhan Xiong, Thien Hoang, and William Yang Wang. DeepPath: A Reinforcement Learning Method for Knowledge Graph Reasoning. In *Empirical Methods in Natural Language Processing*, 2017.

Bishan Yang, Wen-tau Yih, Xiaodong He, Jianfeng Gao, and Li Deng. Embedding Entities and Relations for Learning and Inference in Knowledge Bases. In *International Conference on Learning Representations*, 2015.

## A  CATEGORISING WN18RR RELATIONS

Table 7 describes how each WN18RR relation was assigned to its respective category.

**Table 7:** Explanation for the WN18RR relation category assignment.

| Type | Relation | Relatedness | Subject Specifics | Object Specifics |
|---|---|---|---|---|
| R | verb_group | both verbs; similar in meaning | - | - |
| | derivationally_related_form | different syntactic categories; semantically related | - | - |
| | also_see | semantically similar | - | - |
| S | hypernym | semantically similar | instance of the object | - |
| | instance_hypernym | semantically similar | instance of the object | - |
| C | member_of_domain_usage | loosely semantically related (word usage features) | usage descriptor | broad feature set |
| | member_of_domain_region | loosely semantically related (regional features) | region descriptor | broad feature set |
| | member_meronym | semantically related | collection of objects | part of the subject |
| | has_part | semantically related | collection of objects | part of the subject |
| | synset_domain_topic_of | semantically related | broad feature set | domain descriptor |

## B  CATEGORISING NELL-995 RELATIONS

Categorisation of NELL-995 relations and the explanation for the category assignment of are shown in Tables 8 and 9 respectively.

**Table 8:** Categorisation of NELL-995 relations.

| Type | Relation | Examples *(subject entity, object entity)* |
|---|---|---|
| R | teamplaysagainstteam | *(rangers, mariners), (phillies, tampa_bay_rays)* |
| | clothingtogowithclothing | *(shirts, trousers), (shoes, black_shirt)* |
| | agentcollaborateswithagent | *(white_stripes, jack_white), (barack_obama, hillary_clinton)* |
| S | professionistypeofprofession | *(trial_lawyers, attorneys), (engineers, experts)* |
| | animalistypeofanimal | *(cats, small_animals), (chickens, livestock)* |
| | chemicalistypeofchemical | *(moisture, gas), (oxide, materials)* |
| C | athleteplayssport | *(joe_smith, baseball), (chris_cooley, football)* |
| | itemfoundinroom | *(bed, den), (refrigerator, kitchen_area)* |
| | bodypartcontainsbodypart | *(system002, eyes), (blood, left_ventricle)* |
| | atdate | *(scotland, n2009), (wto, n2003)* |
| | locationlocatedwithinlocation | *(medellin, colombia), (jackson, wyoming)* |
| | atlocation | *(ogunquin, maine), (palmer_lake, colorado)* |

**Table 9:** Explanation for the NELL-995 relation category assignment.

| Type | Relation | Relatedness | Subject Specifics | Object Specifics |
|---|---|---|---|---|
| R | teamplaysagainstteam | both sport teams | - | - |
| | clothingtogowithclothing | both items of clothing that go together | - | - |
| | agentcollaborateswithagent | both people or companies; related industries | - | - |
| S | professionistypeofprofession | semantically related (both profession types) | instance of the object | - |
| | animalistypeofanimal | semantically related (both animals) | instance of the object | - |
| | chemicalistypeofchemical | semantically related (both chemicals) | instance of the object | - |
| C | athleteplayssport | semantically related (sports features) | athlete descriptor | sport descriptor |
| | itemfoundinroom | semantically related (room/furniture features) | item descriptor | room descriptor |
| | bodypartcontainsbodypart | enamically related (specific body part features) | collection of objects | part of the subject |
| | atdate | loosely semantically related (start date features) | broad feature set | date descriptor |
| | locationlocatedwithinlocation | semantically related (geographical features) | part of the subject | collection of objects |
| | atlocation | semantically related (geographical features) | part of the subject | collection of objects |

## C  SPLITTING THE NELL-995 DATASET

The test set of NELL-995 created by Xiong et al. (2017) contains only 10 out of 200 relations present in the training set. To ensure a fair representation of all training set relations in the validation and test sets, we create new validation and test set splits by combining the initial validation and test sets with the training set and randomly selecting 10,000 triples each from the combined dataset.

## D  IMPLEMENTATION DETAILS

All algorithms are re-implemented in PyTorch with the Adam optimizer (Kingma & Ba, 2015) that minimises binary cross-entropy loss, using hyper-parameters that work well for all models (learning rate: 0.001, batch size: 128, number of negative samples: 50). Entity and relation embedding dimensionality is set to $d_e = d_r = 200$ for all models except TuckER, for which $d_r = 30$ (Balažević et al., 2019b).

## E  KRACKHARDT HIERARCHY SCORE

The Krackhardt hierarchy score measures the proportion of node pairs $(x, y)$ where there exists a directed path $x \rightarrow y$, but not $y \rightarrow x$; and it takes a value of one for all directed acyclic graphs, and zero for cycles and cliques (Krackhardt, 2014; Balažević et al., 2019a).

Let $\boldsymbol{M} \in \mathbb{R}^{n \times n}$ be the binary *reachability matrix* of a directed graph $\mathcal{G}$ with $n$ nodes, with $\boldsymbol{M}_{i,j} = 1$ if there exists a directed path from node $i$ to node $j$ and 0 otherwise. The Krackhardt hierarchy score of $\mathcal{G}$ is defined as:

$$\text{Khs}_{\mathcal{G}} = \frac{\sum_{i=1}^{n} \sum_{j=1}^{n} \mathbb{1}(\boldsymbol{M}_{i,j} == 1 \wedge \boldsymbol{M}_{j,i} == 0)}{\sum_{i=1}^{n} \sum_{j=1}^{n} \mathbb{1}(\boldsymbol{M}_{i,j} == 1)}. \tag{1}$$

## F  SYMMETRY SCORE

The symmetry score $\in [-1, 1]$ (Hubert & Baker, 1979) for a relation matrix $\boldsymbol{R} \in \mathbb{R}^{d_e \times d_e}$ is defined as:

$$s = \frac{\sum \sum_{i \neq j} \boldsymbol{R}_{ij} \boldsymbol{R}_{ji} - \frac{(\sum \sum_{i \neq j} \boldsymbol{R}_{ij})^2}{d_e(d_e - 1)}}{\sum \sum_{i \neq j} \boldsymbol{R}_{ij}^2 - \frac{(\sum \sum_{i \neq j} \boldsymbol{R}_{ij})^2}{d_e(d_e - 1)}}, \tag{2}$$

where 1 indicates a symmetric and -1 an anti-symmetric matrix.

## G  "OTHER" PREDICTED FACTS

Tables 10 to 13 shows a sample of the unknown triples (i.e. those formed using the WN18RR entities and relations, but not present in the dataset) for the *derivationally_related form* (R), *instance_hypernym* (S) and *synset_domain_topic_of* (C) relations at a range of probabilities $(\sigma(\phi(e_s, r, e_o)) \approx \{0.4, 0.6, 0.8, 1\})$, as predicted by each model. True triples are indicated in bold; instances where a model predicts an entity is related to itself are indicated in blue.

Table 10: "Other" facts as predicted by MuRE₁.

| Relation (Type) | $\sigma(\phi(e_h, r, e_o)) \approx 0.4$ | $\sigma(\phi(e_h, r, e_o)) \approx 0.6$ | $\sigma(\phi(e_h, r, e_o)) \approx 0.8$ | $\sigma(\phi(e_h, r, e_o)) \approx 1$ |
|---|---|---|---|---|
| derivationally_related_form (R) | (equalizer_NN_1, set_off_VB_5)<br>(constellation_NN_2, satellite_NN_3)<br>(**shrink_VB_3, subtraction_NN_2**)<br>(continue_VB_10, devoted_VB_1)<br>(support_VB_6, defend_VB_5)<br>(blurter_NN_1, fill_up_VB_3)<br>(yawning_NN_1, patellar_reflex_NN_1)<br>(**yaw_NN_1, spiral_VB_1**)<br>(stratum_NN_2, social_group_NN_1)<br>(duel_VB_1, scuffle_NN_3) | (extrapolation_NN_1, maths_NN_1)<br>(spread_VB_5, circularize_VB_3)<br>(flaunt_NN_1, showing_NN_2)<br>(**extrapolate_VB_3, synthesis_NN_3**)<br>(strategist_NN_1, machination_NN_1)<br>(crash_VB_4, grind_VB_2)<br>(tipple_VB_5, steady_VB_2)<br>(flicking_NN_1, vanquish_VB_1)<br>(**synthetical_JJ_1, synthesizer_NN_2**)<br>(realization_NN_2, embodiment_NN_3) | (sever_NN_2, sticher_NN_1)<br>(land_VB_1, vegetable_oil_NN_1)<br>(**snuggle_NN_1, draw_close_VB_3**)<br>(**train_VB_3, training_NN_1**)<br>(**tarnish_VB_3, skin_sensation_NN_1**)<br>(scheme_NN_5, schematization_NN_1)<br>(ordain_VB_3, vest_VB_1)<br>(lie_VB_1, front_end_NN_1)<br>(tread_NN_2, step_NN_9)<br>(**register_NN_3, file_away_NN_1**) | (trail_VB_2, trail_VB_2)<br>(worship_VB_1, worship_VB_1)<br>(steer_VB_1, steer_VB_1)<br>(sort_out_VB_1, sort_out_VB_1)<br>(make_full_VB_1, make_full_VB_1)<br>(utilize_VB_1, utilize_VB_1)<br>(geology_NN_1, geology_NN_1)<br>(zoology_NN_2, zoology_NN_2)<br>(uranology_NN_1, uranology_NN_1)<br>(travel_VB_1, travel_VB_1) |
| instance_hypernym (S) | (thomas_aquinas_NN_1, marry_NN_2)<br>(volcanic_islands_NN_1, volcano_NN_2)<br>(cape_horn_NN_1, urban_center_NN_1)<br>(bergen_NN_1, national_capital_NN_1)<br>(marshall_NN_2, generalship_NN_1)<br>(**tantaon_NN_1, venturer_NN_1**)<br>(wisconsin_NN_2, state_capital_NN_1)<br>(prussia_NN_1, stockade_NN_2)<br>(**de_mille_NN_1, dancing-master_NN_1**)<br>(aegean_sea_NN_1, aegean_island_NN_1) | (taiwan_NN_1, asian_nation_NN_1)<br>(**st._gregory_of_m._NN_1, canonization_NN_1**)<br>(st._gregory_of_m._NN_1, saint_VB_2)<br>(mccormick_NN_1, find_VB_8)<br>(**st._gregory_i_NN_1, bishop_NN_1**)<br>(richard_buckminster_f_NN_1, technological_JJ_2)<br>(thomas_aquinas_NN_1, archbishop_NN_1)<br>(**marshall_NN_2, general_officer_NN_1**)<br>(newman_NN_2, primateship_NN_1)<br>(thomas_the_apostle_NN_1, sanctify_VB_1) | (prophets_NN_1, gospels_NN_1)<br>(malcolm_x_NN_1, passive_resister_NN_1)<br>(taiwan_NN_1, national_capital_NN_1)<br>(truth_NN_5, abolitionism_NN_1)<br>(**thomas_aquinas_NN_1, saint_VB_2**)<br>(central_america_NN_1, s._am._nation_NN_1)<br>(de_mille_NN_1, dance_VB_1)<br>(st._gregory_i_NN_1, apostle_NN_3)<br>(fertile_crescent_NN_1, asian_nation_NN_1)<br>(robert_owen_NN_1, industry_NN_1) | (helsinki_NN_1, urban_center_NN_1)<br>(mannheim_NN_1, stockade_NN_2)<br>(nippon_NN_1, nippon_NN_1)<br>(victor_hugo_NN_1, novel_NN_1)<br>(regiomontanus_NN_1, uranology_NN_1)<br>(**prophets_NN_1, book_NN_10**)<br>(thomas_aquinas_NN_1, church_father_NN_1)<br>(woods_adhrie_NN_1, minstrel_VB_1)<br>(central_america_NN_1, s._am._nation_NN_1)<br>(aegean_sea_NN_1, island_NN_1) |
| synset_domain_topic_of (C) | (write_VB_8, tape_VB_3)<br>(introvert_NN_1, scientific_discipline_NN_1)<br>(**libel_NN_1, draw_VB_6**)<br>(**temple_NN_4, place_of_worship_NN_1**)<br>(trial_impression_NN_1, proof_VB_1)<br>(friend_of_the_court_NN_1, war_machine_NN_1)<br>(**multi_analysis_NN_1, applied_math_NN_1**)<br>(**sell_VB_1, transaction_NN_1**)<br>(draw_VB_6, represent_VB_9) | (draw_VB_6, creative_person_NN_1)<br>(**subsider_NN_1, taxonomic_group_NN_1**)<br>(draw_VB_6, draw_VB_6)<br>(**first_sacker_NN_1, ballplayer_NN_2**)<br>(alchemize_VB_1, modify_VB_3)<br>(**saint_VB_2, catholic_church_NN_1**)<br>(male_JJ_1, masculine_JJ_2)<br>(sell_VB_1, sell_VB_1) | (libel_NN_1, sully_VB_3)<br>(relationship_NN_4, relationship_NN_4)<br>(**turn_VB_2, drive_VB_8**)<br>(drybild_NN_1, mythologize_VB_2)<br>(**drybild_NN_1, myth_NN_1**)<br>(assist_NN_2, am._football_game_NN_1)<br>(initcult_NN_2, human_activity_NN_1)<br>(drive_NN_12, drive_VB_8)<br>(**relationship_NN_4, biology_NN_1**) | (**libel_NN_1, disparagement_NN_1**)<br>(**trolcon_rail-off_NN_1, transport_NN_1**)<br>(**prance_VB_4, equestrian_sport_NN_1**)<br>(**libel_NN_1, traducement_NN_1**)<br>(**sell_VB_1, selling_NN_1**)<br>(trot_VB_2, ride_horseback_VB_1)<br>(prance_VB_4, ride_horseback_VB_1)<br>(gallop_VB_1, ride_horseback_VB_1)<br>(**drybild_NN_1, mythology_NN_3**)<br>(**drive_VB_12, badminton_NN_1**) |

Table 11: "Other" facts as predicted by DistMult.

| Relation (Type) | $\sigma(\phi(e_i, r, c_{o_i})) \approx 0.4$ | $\sigma(\phi(e_i, r, c_{o_i})) \approx 0.6$ | $\sigma(\phi(e_i, r, c_{o_i})) \approx 0.8$ | $\sigma(\phi(e_i, r, c_{o_i})) \approx 1$ |
|---|---|---|---|---|
| derivationally_related_form (R) | *(sag_VB_3, undercover_work_NN_1)* <br> *(print_NN_4, publisher_NN_2)* <br> *(crier_NN_3, prichman_NN_2)* <br> *(play_VB_26, turn_NN_10)* <br> *(count_VB_4, recite_VB_2)* <br> *(vividness_NN_2, imbue_VB_3)* <br> *(sea_mew_NN_1, larus_NN_1)* <br> *(alkali_NN_2, acidify_VB_2)* <br> *(see_VB_17, understand_VB_2)* <br> *(shun_VB_1, helping_NN_2)* | *(dish_NN_2, stew_NN_2)* <br> *(expose_VB_3, show_NN_1)* <br> *(system_NN_9, orderliness_NN_1)* <br> *(spread_NN_4, strew_VB_1)* <br> *(take_down_VB_2, put_VB_2)* <br> *(wrestle_VB_4, wrestler_NN_1)* <br> *(autun_organism_NN_1, epiphytic_JJ_1)* <br> *(duel_VB_1, slugfest_NN_1)* <br> *(vocal_NN_2, rock_star_NN_1)* <br> *(tandling_NN_1, scrim_NN_1)* | *(shrink_NN_1, pedology_NN_1)* <br> *(finish_VB_6, finishing_NN_1)* <br> *(play_VB_26, playing_NN_3)* <br> *(centralization_NN_1, unite_VB_6)* <br> *(existence_NN_1, living_NN_3)* <br> *(mouth_VB_3, sassing_NN_1)* <br> *(constellation_NN_2, star_NN_1)* <br> *(print_VB_4, publishing_house_NN_1)* <br> *(puzzle_VB_3, secret_NN_3)* <br> *(urinology_NN_1, vt_NN_1)* | *(alliterate_VB_1, versifier_NN_1)* <br> *(geology_NN_1, structural_JJ_5)* <br> *(resect_VB_1, amputation_NN_2)* <br> *(nutrition_NN_3, man_NN_4)* <br> *(saint_NN_3, sanctify_VB_1)* <br> *(right_fielder_NN_1, leftfield_NN_1)* <br> *(list_VB_4, slope_NN_1)* <br> *(lieutenancy_NN_1, captain_NN_1)* <br> *(tread_NN_1, step_VB_7)* <br> *(cexateration_NN_1, enucleate_VB_2)* |
| instance_hypernym (S) | *(wisconsin_NN_2, urban_center_NN_1)* <br> *(marshall_NN_2, lieutenant_general_NN_1)* <br> *(abidjan_NN_1, cote_d'ivoire_NN_1)* <br> *(world_war_i_NN_1, urban_center_NN_1)* <br> *(st_paul_NN_2, evangelist_NN_2)* <br> *(deep_south_NN_1, urban_center_NN_1)* <br> *(naptse_NN_1, urban_center_NN_1)* <br> *(ticino_NN_1, urban_center_NN_1)* <br> *(aegean_sea_NN_1, war_of_am_ind_NN_1)* <br> *(compens_NN_1, war_of_am_ind_NN_1)* | *(mississippi_river_NN_1, american_state_NN_1)* <br> *(tc_e_byrd_NN_1, commissioned_officer_NN_1)* <br> *(the_gambia_NN_1, africa_NN_1)* <br> *(trich_mir_NN_1, urban_center_NN_1)* <br> *(tc_e_byrd_NN_1, military_advisor_NN_1)* <br> *(tc_e_byrd_NN_1, aide-de-camp_NN_1)* <br> *(tampa_bay_NN_1, urban_center_NN_1)* <br> *(tidewater_region_NN_1, south_NN_1)* <br> *(tc_e_byrd_NN_1, executive_officer_NN_1)* | *(deep_south_NN_1, south_NN_1)* <br> *(capital_of_gambia_NN_1, urban_center_NN_1)* <br> *(south_west_africa_NN_1, africa_NN_1)* <br> *(brandenburg_NN_1, urban_center_NN_1)* <br> *(sierra_nevada_NN_1, urban_center_NN_1)* <br> *(malcolm_x_NN_1, emancipationist_NN_1)* <br> *(north_platte_river_NN_1, urban_center_NN_1)* <br> *(adw_NN_1, urban_center_NN_1)* <br> *(zaire_river_NN_1, urban_center_NN_1)* <br> *(transylvanian_alps_NN_1, urban_center_NN_1)* | *(helsinki_NN_1, urban_center_NN_1)* <br> *(the_nazarene_NN_1, save_VB_2)* <br> *(irish_capital_NN_1, urban_center_NN_1)* <br> *(tc_e_byrd_NN_1, inspector_general_NN_1)* <br> *(tc_e_byrd_NN_1, chief_of_staff_NN_1)* <br> *(central_america_NN_1, c_am_canton_NN_1)* <br> *(malcolm_x_NN_1, environmentalist_NN_1)* <br> *(the_nazarene_NN_1, christian_JJ_1)* <br> *(thomas_aquinas_NN_1, church_father_NN_1)* <br> *(the_nazarene_NN_1, et_zino_NN_2)* |
| synset_domain_topic_of (C) | *(limitation_NN_4, trammel_VB_2)* <br> *(light_colonel_NN_1, colonel_NN_1)* <br> *(nurse_VB_1, nursing_NN_1)* <br> *(sermon_NN_1, prophecy_VB_2)* <br> *(slugger_NN_1, baseball_player_NN_1)* <br> *(rna_NN_1, chemistry_NN_1)* <br> *(metrify_VB_1, versify_VB_1)* <br> *(vivid_impression_NN_1, publish_NN_1)* <br> *(corpus_NN_1, war_of_am_ind_NN_1)* | *(roll-on_roll-off_NN_1, transport_NN_1)* <br> *(hizb_ut-tahrir_NN_1, asia_NN_1)* <br> *(slugger_NN_1, softball_game_NN_1)* <br> *(sermon_NN_1, sermonize_VB_1)* <br> *(draw_VB_1, drawing_NN_1)* <br> *(turn_VB_12, plow_NN_1)* <br> *(assist_NN_2, softball_game_NN_1)* <br> *(council_NN_2, assembly_NN_4)* <br> *(throughput_NN_1, turnout_NN_4)* <br> *(cream_VB_1, cream_NN_2)* | *(etymologizing_NN_1, explanation_NN_1)* <br> *(ferry_VB_3, travel_VB_1)* <br> *(public_prosecutor_NN_1, prosecute_VB_2)* <br> *(alchemize_VB_1, modify_VB_3)* <br> *(libel_NN_1, libel_VB_1)* <br> *(turn_VB_12, till_VB_1)* <br> *(hit_NN_1, hit_VB_1)* <br> *(fire_VB_3, flaming_NN_1)* <br> *(ring_NN_4, chemical_chain_NN_1)* <br> *(kilidinal_energy_NN_1, charge_NN_9)* | *(flat_JJ_5, matte_NN_2)* <br> *(etymologizing_NN_1, derive_VB_3)* <br> *(hide_out_VB_1, hide_NN_3)* <br> *(relationship_NN_4, relation_NN_1)* <br> *(drive_VB_4, driving_NN_1)* <br> *(etymologizing_NN_1, etymologize_VB_2)* <br> *(matrix_algebra_NN_1, diagonalization_NN_1)* <br> *(cabinetwork_NN_2, woodworking_NN_1)* <br> *(cabinetwork_NN_2, bottom_VB_1)* <br> *(cabinetwork_NN_2, upholster_VB_1)* |

Table 12: "Other" facts as predicted by TuckER.

| Relation (Type) | $\sigma(\phi(e_s, r, e_o)) \approx 0.4$ | $\sigma(\phi(e_s, r, e_o)) \approx 0.6$ | $\sigma(\phi(e_s, r, e_o)) \approx 0.8$ | $\sigma(\phi(e_s, r, e_o)) \approx 1$ |
|---|---|---|---|---|
| derivationally_related_form (R) | (tympanist_NN_1, gong_NN_2)<br>(indication_NN_1, signalize_VB_2)<br>(turn_over_VB_3, rotation_NN_3)<br>(date_VB_5, geological_dating_NN_1)<br>(set_VB_23, emblem_NN_2)<br>(tyro_NN_1, start_VB_5)<br>(identification_NN_1, name_VB_5)<br>(stabber_NN_1, thrust_VB_5)<br>(justification_NN_1, apology_NN_3)<br>(manufacture_VB_1, prevarication_NN_1) | (mash_NN_2, mill_VB_2)<br>(walk_VB_9, zimmer_frame_NN_1)<br>(use_VB_5, utility_NN_2)<br>(musical_instrument_NN_1, write_VB_6)<br>(lining_NN_3, wrap_up_VB_1)<br>(scrap_VB_2, struggle_NN_2)<br>(tape_VB_3, tape_recorder_NN_1)<br>(vindicate_VB_2, justification_NN_2)<br>(leaching_NN_1, percolate_VB_3)<br>(synchronize_VB_2, synchroscope_NN_1) | (take_chances_VB_1, venture_NN_1)<br>(shatter_NN_1, fill_up_VB_3)<br>(exit_NN_3, leave_VB_1)<br>(trembler_NN_1, vibrate_VB_1)<br>(motivator_NN_1, trip_VB_4)<br>(support_VB_6, indorsement_NN_1)<br>(federate_VB_2, confederation_NN_1)<br>(take_over_VB_6, return_NN_7)<br>(wait_on_VB_1, supporter_NN_3)<br>(denote_VB_3, promulgation_NN_1) | (venturer_NN_2, venture_NN_2)<br>(dynamitist_NN_1, dynamitist_NN_1)<br>(love_VB_3, lover_VB_1)<br>(snuggle_NN_1, squeeze_VB_8)<br>(departed_NN_1, die_VB_2)<br>(position_VB_1, placement_NN_1)<br>(repenian_JJ_1, repenian_JJ_1)<br>(tread_NN_1, step_VB_7)<br>(stockist_NN_1, stockist_NN_1)<br>(philanthropist_NN_1, philanthropist_NN_1) |
| instance_hypernym (S) | (deep_south_NN_1, south_NN_1)<br>(st._paul_NN_2, oregonist_NN_1)<br>(helsinki_NN_1, urban_center_NN_1)<br>(malcolm_x_NN_1, emancipationist_NN_1)<br>(st._gregory_of_m._NN_1, church_father_NN_1)<br>(robert_owen_NN_1, movie_maker_NN_1)<br>(st._paul_NN_2, philosopher_NN_1)<br>(theresa_NN_1, monk_NN_1)<br>(ibn-roshd_NN_1, pedagogue_NN_1) | (thomas_aquinas_NN_1, bishop_NN_1)<br>(irish_capital_NN_1, urban_center_NN_1)<br>(thomas_the_apostle_NN_1, apostle_NN_2)<br>(st._paul_NN_2, apostle_NN_3)<br>(mccormick_NN_1, painter_NN_1)<br>(thomas_the_apostle_NN_1, troglodyte_NN_1)<br>(mccormick_NN_1, electrical_engineer_NN_1)<br>(mississippi_river_NN_1, american_state_NN_1) | (cowpens_NN_1, siege_NN_1)<br>(mccormick_NN_1, arms_manufacturer_NN_1)<br>(thomas_the_apostle_NN_1, evangelist_NN_2)<br>(mccormick_NN_1, technologist_NN_1)<br>(st._gregory_i_NN_1, church_father_NN_1) | (t._e._ford_NN_1, siege_NN_1)<br>(shaw_NN_3, women's_rightist_NN_1)<br>(aegean_sea_NN_1, aegean_island_NN_1)<br>(thomas_aquinas_NN_1, church_father_NN_1) |
| synset_domain_topic_of (C) | (rollon_rolloff_NN_1, motorcar_NN_1)<br>(libel_NN_1, legislature_NN_1)<br>(rollon_rolloff_NN_1, passenger_vehicle_NN_1) | (drive_NN_12, badminton_NN_1) | | |

Table 13: "Other" facts as predicted by MuRE.

| Relation (Type) | $\sigma(\phi(e_s, r, e_o)) \approx 0.4$ | $\sigma(\phi(e_s, r, e_o)) \approx 0.6$ | $\sigma(\phi(e_s, r, e_o)) \approx 0.8$ | $\sigma(\phi(e_s, r, e_o)) \approx 1$ |
|---|---|---|---|---|
| derivationally_related_form (R) | (surround_VB_1, wall_NN_1)<br>(unpleasant_JJ_1, unpleasantness_NN_1)<br>(love_VB_3, enjoyment_NN_2)<br>(magnitude_NN_1, tall_JJ_1)<br>(testify_VB_2, information_NN_1)<br>(connect_VB_6, converging_NN_1)<br>(connect_VB_6, connexion_NN_4)<br>(operate_VB_4, psyop_NN_1)<br>(market_VB_1, trade_NN_4)<br>(operate_VB_4, mission_NN_2) | (word_picture_NN_1, sketch_VB_2)<br>(develop_VB_10, adjustment_NN_4)<br>(gloss_VB_3, commentary_NN_1)<br>(violate_VB_2, violation_NN_3)<br>(suffocate_VB_1, strangle_free_NN_1)<br>(number_VB_5, point_NN_12)<br>(develop_VB_10, organic_process_NN_1)<br>(plication_NN_1, twist_VB_4)<br>(split_up_VB_3, separation_NN_5)<br>(plication_NN_1, wrinkle_VB_2) | (smelling_NN_1, wind_VB_4)<br>(try_out_VB_1, somatic_cell_nuclear_transplantation_NN_1)<br>(lighting_NN_4, set_on_fire_VB_1)<br>(symphonique_NN_1, one-half_NN_1)<br>(just_JJ_3, validity_NN_1)<br>(reprove_VB_1, talking_to_NN_1)<br>(sustain_VB_5, beam_NN_2)<br>(spring_NN_6, hurdle_VB_1)<br>(spark_NN_1, scintillate_VB_1)<br>(utility_NN_2, functional_JJ_1) | (polution_NN_2, sack_VB_1)<br>(desire_NN_2, hope_VB_2)<br>(snuffle_VB_3, whine_NN_1)<br>(nasalization_NN_1, one-half_out_VB_1)<br>(tread_NN_1, step_VB_7)<br>(yarn_VB_1, pining_NN_1)<br>(unrelatedness_NN_1, unbinary_JJ_1)<br>(travesty_NN_2, travesty_NN_2)<br>(spark_NN_1, scintillate_VB_1)<br>(stockist_NN_1, stockist_NN_1) |
| instance_hypernym (S) | (malcolm_x_NN_1, hipster_NN_1)<br>(the_nazarene_NN_1, judaism_NN_2)<br>(told_line_state_NN_1, river_NN_1)<br>(t-e-_byrd_NN_1, peninsula_NN_1)<br>(south_korea_NN_1, peninsula_NN_1)<br>(st._gregory_of_n_NN_1, vicar_of_christ_NN_1)<br>(trippin_NN_1, indian_region_NN_1)<br>(robert_owen_NN_1, tycoon_NN_1)<br>(mandalay_NN_1, national_capital_NN_1)<br>(nan_ling_NN_1, urban_center_NN_1) | (central_america_NN_1, central_america_NN_1)<br>(st._gregory_i_NN_1, church_father_NN_1)<br>(south_korea_NN_1, african_nation_NN_1)<br>(malcolm_x_NN_1, passive_resister_NN_1)<br>(malcolm_x_NN_1, birth-control_reformer_NN_1)<br>(los_angeles_NN_1, port_NN_1)<br>(great_lakes_NN_1, canadian_province_NN_1)<br>(transylvanian_alps_NN_1, urban_center_NN_1)<br>(gettysburg_NN_2, siege_NN_1)<br>(wisconsin_NN_2, geographical_region_NN_1) | (theresa_NN_1, monk_NN_1)<br>(trippin_NN_1, campuno_nation_NN_1)<br>(great_lakes_NN_1, river_NN_1)<br>(t-e-_byrd_NN_1, noncommissioned_officer_NN_1)<br>(world_war_i_NN_1, pitched_battle_NN_1)<br>(irish_capital_NN_1, urban_center_NN_1)<br>(volcanic_islands_NN_1, urban_center_NN_1)<br>(trippin_NN_1, american_state_NN_1)<br>(helsinki_NN_1, urban_center_NN_1)<br>(capital_of_gambia_NN_1, urban_center_NN_1) | |
| synset_domain_topic_of (C) | (libel_NN_1, criminal_law_NN_1)<br>(bryohid_NN_1, mythology_NN_2)<br>(slugger_NN_1, sport_NN_1)<br>(tell_VB_1, law_NN_1)<br>(semitic_deity_NN_1, mythology_NN_1)<br>(nuclear_deterrence_NN_1, law_NN_1)<br>(reception_NN_5, baseball_game_NN_1)<br>(photosynthesis_NN_1, chemistry_NN_1)<br>(isolde_NN_1, parable_NN_1)<br>(assist_NN_2, court_game_NN_1) | (write_VB_8, transcription_NN_5)<br>(temple_NN_4, maximum_NN_2)<br>(assist_NN_2, hockey_NN_1)<br>(relationship_NN_4, biology_NN_1)<br>(apostle_NN_3, western_church_NN_1)<br>(trot_VB_2, equestrian_sport_NN_1)<br>(assist_NN_2, soccer_NN_1)<br>(assist_NN_2, football_game_NN_1) | (assist_NN_2, am_football_game_NN_1)<br>(drive_NN_12, court_game_NN_1)<br>(tell_VB_1, offense_NN_3)<br>(slugger_NN_1, softball_game_NN_1)<br>(drive_NN_12, badminton_NN_1) | |

## 6.3   Impact

According to Google Scholar, the paper has received 15 citations as of November 2021.

The presented paper has directly influenced the work of Guo and Kok (2021), who leverage the finding that both additive and multiplicative components are required for representing relations to achieve state-of-the-art results with their BiQUE model.

## 6.4   Discussion

By laying theoretical foundations for unifying word embeddings and knowledge graph representation, the presented paper opens up several avenues for future work.

Firstly, the insight about commonalities in the latent structure between PMI-based word embeddings and knowledge graph entity embeddings can be used to develop a theoretically principled joint model of words and knowledge graph entities and relations, which learns from statistics contained in multiple data sources and which may lead to unifying and improving both word and relation representations.

Further, the proposed theoretical model can serve as a basis for a method to impose logical rules between knowledge graph relations by ensuring that certain geometric relationships hold between the corresponding relation representations (as done by rule injection approaches of Rocktäschel et al. (2015) and Demeester et al. (2016)), avoiding the need for costly grounding of each rule. For example, to enforce the inversion rule $(e_s, r_1, e_o) \Leftrightarrow (e_o, r_2, e_s)$ (e.g. *(London, capital_of, UK)* $\Leftrightarrow$ *(UK, has_capital, London)*), both relations need to project entities onto the same subspace component through $\mathbf{P}_{r_1} = \mathbf{P}_{r_2}$ to test for relatedness and the constraint for relation specific vector offsets follows directly from $\mathbf{r}_1 = \mathbf{v}_r^o - \mathbf{v}_r^s$ and $\mathbf{r}_2 = \mathbf{v}_r^s - \mathbf{v}_r^o$, i.e. $\mathbf{r}_1 = -\mathbf{r}_2$. Rule injection will likely lead to improved relation representations and thus link prediction performance, particularly for relations with limited training data.

Lastly, the presented paper proposes a projection matrix $\mathbf{P}_r$ as a part of the score function which identifies relatedness between entities by projecting the corresponding entity embeddings onto a common subspace and comparing their images. However, it does not consider the specific type of projection necessary (e.g. MuRE (Balažević et al., 2019a) and DistMult (Yang et al., 2015) use a stretch, RotE (Chami et al., 2020b) uses rotation). Future work should investigate how different projection types relate to semantic aspects of relatedness.

# Chapter 7

# Conclusion

This thesis introduces three approaches to knowledge graph representation which improve current state-of-the-art in link prediction, as well as a theory on how semantics can be encoded in the geometry of relation representations. In Chapter 3, we introduce HypER, a convolutional knowledge graph representation model which is adjacent to the family of tensor factorisation models. Chapter 4 introduces TuckER, a straightforward linear model which learns entity and relation representations through the Tucker decomposition of the knowledge graph adjacency tensor. The focus of Chapter 5 is on modelling hierarchical relations by introducing MuRP, a knowledge graph representation model which embeds entities and relations in hyperbolic rather than Euclidean space. Lastly, in Chapter 6, we draw a link between the geometry of word embedding interactions and knowledge graph relation representations. More specifically, we build upon recent theoretical understanding on how different semantic relations, such as similarity, paraphrase and analogy, are encoded in the geometric interactions of PMI-based word embeddings and expand it to more general relations of knowledge graphs, with the aim of aiding and directing future model design and moving towards unified and potentially improved models which learn from multiple data sources.

All presented models are designed to be relatively simple in order to scale to large knowledge graphs and to foster theoretical analysis — HypER simplifies ConvE, TuckER can be seen as a simplified version of the tensor factorisation view of HypER and MuRE is in part inspired by the work of Allen et al. (2019) that connects the vector offset of analogies to the semantics of relations. Even though all models developed in this thesis are limited to binary relations, their extensions to temporal and n-ary relational knowledge graphs are relatively straightforward and have already been developed by follow-up work. However, unlike GNN-based models, the presented embedding-based models are transductive and are not easily extendable to an inductive setting without retraining the models from scratch on unseen training examples.

The theoretical part of the thesis which associates semantics with geometry of relation representations can be viewed as a starting point towards learning more interpretable and theoretically motivated knowledge graph representation models which opens up many interesting research questions for future work:

- How can we use the connection between word embeddings and knowledge graph representations to train a joint, potentially improved model?

- Can the developed theory which links semantics to geometric components of existing models (multiplicative components to identify relatedness, additive components to identify specialisation and context-shift) be used to inject logical rules between relations as geometric constraints between relation representations?

- Can we use the identified relation conditions for different semantic relation types to design geometric components of relation representations beyond simple additive and multiplicative components?

The presented theory can be seen as a starting point for developing models which explicitly integrate semantics of relations into the geometry of relation representations. However, a potential limitation of this theory is that it is based on models of distributional semantics, which (i) do not consider higher order statistics other than word co-occurrences; and (ii) disregard word order by averaging out positional information over the context window around each target word. Some of these drawbacks may be alleviated by instead considering the properties of word embeddings learned by large language models, which we leave to future work.

# Bibliography

Allen, C., Balažević, I., and Hospedales, T. (2019). What the Vec? Towards Probabilistically Grounded Embeddings. In *Advances in Neural Information Processing Systems*.

Allen, C. and Hospedales, T. (2019). Analogies Explained: Towards Understanding Word Embeddings. In *International Conference on Machine Learning*.

Amin, S., Varanasi, S., Dunfield, K. A., and Neumann, G. (2020). LowFER: Low-rank Bilinear Pooling for Link Prediction. In *International Conference on Machine Learning*.

Balažević, I., Allen, C., and Hospedales, T. M. (2019). Hypernetwork Knowledge Graph Embeddings. In *International Conference on Artificial Neural Networks*.

Balažević, I., Allen, C., and Hospedales, T. M. (2019a). Multi-relational Poincaré Graph Embeddings. In *Advances in Neural Information Processing Systems*.

Balažević, I., Allen, C., and Hospedales, T. M. (2019b). TuckER: Tensor Factorization for Knowledge Graph Completion. In *Empirical Methods in Natural Language Processing*.

Baroni, M., Dinu, G., and Kruszewski, G. (2014). Don't Count, Predict! A Systematic Comparison of Context-Counting vs. Context-Predicting Semantic Vectors. In *Association for Computational Linguistics*.

Battaglia, P. W., Hamrick, J. B., Bapst, V., Sanchez-Gonzalez, A., Zambaldi, V., Malinowski, M., Tacchetti, A., Raposo, D., Santoro, A., Faulkner, R., et al. (2018). Relational Inductive Biases, Deep Learning, and Graph Networks. *arXiv preprint arXiv:1806.01261*.

Battaglia, P. W., Pascanu, R., Lai, M., Rezende, D., and Kavukcuoglu, K. (2016). Interaction Networks for Learning about Objects, Relations and Physics. In *Advances in Neural Information Processing Systems*.

Bollacker, K., Evans, C., Paritosh, P., Sturge, T., and Taylor, J. (2008). Freebase: A Collaboratively Created Graph Database for Structuring Human Knowledge. In *ACM SIGMOD International Conference on Management of Data*.

Bordes, A., Usunier, N., Garcia-Duran, A., Weston, J., and Yakhnenko, O. (2013). Translating Embeddings for Modeling Multi-relational Data. In *Advances in Neural Information Processing Systems*.

Cannon, J. W., Floyd, W. J., Kenyon, R., Parry, W. R., et al. (1997). Hyperbolic Geometry. *Flavors of Geometry*, 31:59–115.

Chami, I., Abu-El-Haija, S., Perozzi, B., Ré, C., and Murphy, K. (2020a). Machine Learning on Graphs: A Model and Comprehensive Taxonomy. *arXiv preprint arXiv:2005.03675.*

Chami, I., Wolf, A., Juan, D.-C., Sala, F., Ravi, S., and Ré, C. (2020b). Low-Dimensional Hyperbolic Knowledge Graph Embeddings. In *Association for Computational Linguistics.*

Das, R., Dhuliawala, S., Zaheer, M., Vilnis, L., Durugkar, I., Krishnamurthy, A., Smola, A., and McCallum, A. (2018). Go for a Walk and Arrive at the Answer: Reasoning over Paths in Knowledge Bases Using Reinforcement Learning. In *International Conference on Learning Representations.*

De Sa, C., Gu, A., Ré, C., and Sala, F. (2018). Representation Tradeoffs for Hyperbolic Embeddings. In *International Conference on Machine Learning.*

Demeester, T., Rocktäschel, T., and Riedel, S. (2016). Lifted Rule Injection for Relation Embeddings. In *Empirical Methods in Natural Language Processing.*

Dettmers, T., Minervini, P., Stenetorp, P., and Riedel, S. (2018). Convolutional 2D Knowledge Graph Embeddings. In *Association for the Advancement of Artificial Intelligence.*

Devlin, J., Chang, M.-W., Lee, K., and Toutanova, K. (2018). BERT: Pre-training of Deep Bidirectional Transformers for Language Understanding. In *North American Chapter of the Association for Computational Linguistics.*

Engelcke, M., Kosiorek, A. R., Jones, O. P., and Posner, I. (2020). GENESIS: Generative Scene Inference and Sampling with Object-centric Latent Representations. In *International Conference on Learning Representations.*

Firth, J. R. (1957). A Synopsis of Linguistic Theory, 1930-1955. *Studies in Linguistic Analysis.*

Ganea, O., Bécigneul, G., and Hofmann, T. (2018a). Hyperbolic Neural Networks. In *Advances in Neural Information Processing Systems.*

Ganea, O.-E., Bécigneul, G., and Hofmann, T. (2018b). Hyperbolic Entailment Cones for Learning Hierarchical Embeddings. In *International Conference on Machine Learning.*

Greff, K., Kaufman, R. L., Kabra, R., Watters, N., Burgess, C., Zoran, D., Matthey, L., Botvinick, M., and Lerchner, A. (2019). Multi-object Representation Learning with Iterative Variational Inference. In *International Conference on Machine Learning.*

Guo, J. and Kok, S. (2021). BiQUE: Biquaternionic Embeddings of Knowledge Graphs. In *Empirical Methods in Natural Language Processing.*

Han, Z., Ma, Y., Chen, P., and Tresp, V. (2020). DyERNIE: Dynamic Evolution of Riemannian Manifold Embeddings for Temporal Knowledge Graph Completion. In *Empirical Methods in Natural Language Processing.*

Harris, Z. S. (1954). Distributional Structure. *Word*, 10:146–162.

Hogan, A., Blomqvist, E., Cochez, M., d'Amato, C., de Melo, G., Gutierrez, C., Gayo, J. E. L., Kirrane, S., Neumaier, S., Polleres, A., et al. (2020). Knowledge Graphs. *arXiv preprint arXiv:2003.02320.*

Ji, S., Pan, S., Cambria, E., Marttinen, P., and Yu, P. S. (2021). A Survey on Knowledge Graphs: Representation, Acquisition and Applications. In *IEEE Transactions on Neural Networks and Learning Systems*.

Kadlec, R., Bajgar, O., and Kleindienst, J. (2017). Knowledge Base Completion: Baselines Strike Back. In *ACL Workshop on Representation Learning for NLP*.

Kazemi, S. M., Goel, R., Jain, K., Kobyzev, I., Sethi, A., Forsyth, P., and Poupart, P. (2020). Representation Learning for Dynamic Graphs: A Survey. *Journal of Machine Learning Research*.

Kazemi, S. M. and Poole, D. (2018). SimplE Embedding for Link Prediction in Knowledge Graphs. In *Advances in Neural Information Processing Systems*.

Kipf, T., Fetaya, E., Wang, K.-C., Welling, M., and Zemel, R. (2018). Neural Relational Inference for Interacting Systems. In *International Conference on Machine Learning*.

Kipf, T., van der Pol, E., and Welling, M. (2020). Contrastive Learning of Structured World Models. In *International Conference on Learning Representations*.

Kipf, T. N. and Welling, M. (2017). Semi-supervised Classification with Graph Convolutional Networks. In *International Conference on Learning Representations*.

Lacroix, T., Usunier, N., and Obozinski, G. (2018). Canonical Tensor Decomposition for Knowledge Base Completion. In *International Conference on Machine Learning*.

Levy, O. and Goldberg, Y. (2014). Neural Word embedding as Implicit Matrix Factorization. In *Advances in Neural Information Processing Systems*.

Liu, Y., Yao, Q., and Li, Y. (2020). Generalizing Tensor Decomposition for n-ary Relational Knowledge Bases. In *The Web Conference*.

Luan, Y., He, L., Ostendorf, M., and Hajishirzi, H. (2018). Multi-task Identification of Entities, Relations, and Coreference for Scientific Knowledge Graph Construction. In *Empirical Methods in Natural Language Processing*.

Ma, Y., Tresp, V., and Daxberger, E. A. (2019). Embedding Models for Episodic Knowledge Graphs. *Journal of Web Semantics*, 59:100490.

Mikolov, T., Sutskever, I., Chen, K., Corrado, G. S., and Dean, J. (2013a). Distributed Representations of Words and Phrases and their Compositionality. In *Advances in Neural Information Processing Systems*.

Mikolov, T., Yih, W.-t., and Zweig, G. (2013b). Linguistic Regularities in Continuous Space Word Representations. In *North American Chapter of the Association for Computational Linguistics*.

Nguyen, D. Q. (2020). An Overview of Embedding Models of Entities and Relationships for Knowledge Base Completion. *arXiv preprint arXiv:1703.08098*.

Nguyen, D. Q., Sirts, K., Qu, L., and Johnson, M. (2016). STransE: A Novel Embedding Model of Entities and Relationships in Knowledge Bases. In *North American Chapter of the Association for Computational Linguistics: Human Language Technologies*.

Nickel, M. and Kiela, D. (2017). Poincaré Embeddings For Learning Hierarchical Representations. In *Advances in Neural Information Processing Systems*.

Nickel, M. and Kiela, D. (2018). Learning Continuous Hierarchies in the Lorentz Model of Hyperbolic Geometry. In *International Conference on Machine Learning*.

Nickel, M., Murphy, K., Tresp, V., and Gabrilovich, E. (2015). A Review of Relational Machine Learning for Knowledge Graphs. *Proceedings of the IEEE*, 104(1):11–33.

Nickel, M., Tresp, V., and Kriegel, H.-P. (2011). A Three-Way Model for Collective Learning on Multi-Relational Data. In *International Conference on Machine Learning*.

Pennington, J., Socher, R., and Manning, C. (2014). GloVe: Global Vectors for Word Representation. In *Empirical Methods in Natural Language Processing*.

Peters, M. E., Neumann, M., Logan IV, R. L., Schwartz, R., Joshi, V., Singh, S., and Smith, N. A. (2019). Knowledge Enhanced Contextual Word Representations. In *Empirical Methods in Natural Language Processing*.

Petroni, F., Rocktäschel, T., Lewis, P., Bakhtin, A., Wu, Y., Miller, A. H., and Riedel, S. (2019). Language Models as Knowledge Bases? In *Empirical Methods in Natural Language Processing*.

Pezeshkpour, P., Tian, Y., and Singh, S. (2020). Revisiting Evaluation of Knowledge Base Completion Models. In *Automated Knowledge Base Construction*.

Radford, A., Narasimhan, K., Salimans, T., and Sutskever, I. (2018). Improving Language Understanding by Generative Pre-training.

Radford, A., Wu, J., Child, R., Luan, D., Amodei, D., and Sutskever, I. (2019). Language Models are Unsupervised Multitask Learners.

Riedel, S., Yao, L., McCallum, A., and Marlin, B. M. (2013). Relation Extraction with Matrix Factorization and Universal Schemas. In *North American Chapter of the Association for Computational Linguistics: Human Language Technologies*.

Rocktäschel, T., Singh, S., and Riedel, S. (2015). Injecting Logical Background Knowledge into Embeddings for Relation Extraction. In *North American Chapter of the Association for Computational Linguistics: Human Language Technologies*.

Ruffinelli, D., Broscheit, S., and Gemulla, R. (2019). You Can Teach an Old Dog New Tricks! On Training Knowledge Graph Embeddings. In *International Conference on Learning Representations*.

Sarkar, R. (2011). Low Distortion Delaunay Embedding of Trees in Hyperbolic Plane. In *International Symposium on Graph Drawing*.

Sarkar, R., Goswami, K., Arcan, M., and McCrae, J. P. (2020). Suggest me a Movie for tonight: Leveraging Knowledge Graphs for Conversational Recommendation. In *International Conference on Computational Linguistics*.

Saxena, C., Liu, T., and King, I. (2020). A Survey of Graph Curvature and Embedding in Non-Euclidean Spaces. In *International Conference on Neural Information Processing*, pages 127–139. Springer.

Schlichtkrull, M., Kipf, T. N., Bloem, P., van den Berg, R., Titov, I., and Welling, M. (2018). Modeling Relational Data with Graph Convolutional Networks. In *European Semantic Web Conference*.

Shimizu, R., Mukuta, Y., and Harada, T. (2021). Hyperbolic Neural Networks++. In *International Conference on Learning Representations*.

Spelke, E. S. and Kinzler, K. D. (2007). Core Knowledge. *Developmental Science*, 10(1):89–96.

Speranskaya, M., Schmitt, M., and Roth, B. (2020). Ranking vs. Classifying: Measuring Knowledge Base Completion Quality. In *Automated Knowledge Base Construction*.

Sun, Z., Vashishth, S., Sanyal, S., Talukdar, P., and Yang, Y. (2020). A Re-evaluation of Knowledge Graph Completion Methods. In *Association for Computational Linguistics*.

Tifrea, A., Bécigneul, G., and Ganea, O.-E. (2019). Poincaré GloVe: Hyperbolic Word Embeddings. In *International Conference on Learning Representations*.

Toutanova, K., Chen, D., Pantel, P., Poon, H., Choudhury, P., and Gamon, M. (2015). Representing Text for Joint Embedding of Text and Knowledge Bases. In *Empirical Methods in Natural Language Processing*.

Trouillon, T., Welbl, J., Riedel, S., Gaussier, É., and Bouchard, G. (2016). Complex Embeddings for Simple Link Prediction. In *International Conference on Machine Learning*.

Tucker, L. R. (1964). The Extension of Factor Analysis to Three-Dimensional Matrices. *Contributions to Mathematical Psychology*, 110119.

Ungar, A. A. (2001). Hyperbolic Trigonometry and its Application in the Poincaré Ball Model of Hyperbolic Geometry. *Computers & Mathematics with Applications*, 41(1-2):135–147.

Vashishth, S., Sanyal, S., Nitin, V., and Talukdar, P. (2020). Composition-based Multi-relational Graph Convolutional Networks. In *International Conference on Learning Representations*.

Veličković, P., Cucurull, G., Casanova, A., Romero, A., Lio, P., and Bengio, Y. (2018). Graph Attention Networks. In *International Conference on Learning Representations*.

Wadden, D., Wennberg, U., Luan, Y., and Hajishirzi, H. (2019). Entity, Relation, and Event Extraction with Contextualized Span Representations. In *Empirical Methods in Natural Language Processing*.

Wang, S., Wei, X., Nogueira dos Santos, C., Wang, Z., Nallapati, R., Arnold, A., Xiang, B., and Yu, P. S. (2020). H2KGAT: Hierarchical Hyperbolic Knowledge Graph Attention Network. In *Empirical Methods in Natural Language Processing*.

Wang, S., Wei, X., Nogueira dos Santos, C. N., Wang, Z., Nallapati, R., Arnold, A., Xiang, B., Yu, P. S., and Cruz, I. F. (2021). Mixed-Curvature Multi-Relational Graph Neural Network for Knowledge Graph Completion. In *The Web Conference 2021*.

West, R., Gabrilovich, E., Murphy, K., Sun, S., Gupta, R., and Lin, D. (2014). Knowledge Base Completion via Search-Based Question Answering. In *International Conference on World Wide Web*.

Xiong, W., Hoang, T., and Wang, W. Y. (2017). DeepPath: A Reinforcement Learning Method for Knowledge Graph Reasoning. In *Empirical Methods in Natural Language Processing*.

Yan, S., Zhang, Z., Sun, X., Xu, G., Jin, L., and Li, S. (2021). HYPER$^2$: Hyperbolic Poincare Embedding for Hyper-Relational Link Prediction. *arXiv preprint arXiv:2104.09871*.

Yang, B., Yih, W.-t., He, X., Gao, J., and Deng, L. (2015). Embedding Entities and Relations for Learning and Inference in Knowledge Bases. In *International Conference on Learning Representations*.

Zhang, S., Tay, Y., Jiang, W., Juan, D.-c., and Zhang, C. (2021a). Switch Spaces: Learning Product Spaces with Sparse Gating. *arXiv preprint arXiv:2102.08688*.

Zhang, Y., Wang, X., Shi, C., Liu, N., and Song, G. (2021b). Lorentzian Graph Convolutional Networks. In *The Web Conference*.